%% file: main.tex
\documentclass{article}
\usepackage{fix-cm}
\usepackage{arxiv}

\usepackage[utf8]{inputenc}
\usepackage[T1]{fontenc}

\newcommand{\eg}{e.g.\xspace}
\newcommand{\ie}{i.e.\xspace}

\newcommand{\wrt}{w.r.t.\xspace}
\newcommand{\vs}{vs.\xspace}

\usepackage{graphicx}
\usepackage{booktabs}
\usepackage{xcolor}
\usepackage{pifont}
\usepackage{url}
\usepackage{amsmath}
\usepackage{textgreek}
\usepackage{amsfonts}
\usepackage{nicefrac}
\usepackage{microtype}
\usepackage[numbers,sort&compress]{natbib}
\usepackage{xspace}
\newcommand{\vmark}{\ding{51}}
\newcommand{\xmark}{\ding{55}}

\usepackage{multirow}
\usepackage[normalem]{ulem} 
\usepackage{siunitx}

\providecommand{\bestval}[1]{\textbf{#1}}
\providecommand{\secondval}[1]{\uline{#1}}

\usepackage{hyperref}
\usepackage{cleveref}

\newcommand{\blfootnote}[1]{%
  \begingroup
    \renewcommand\thefootnote{}
    \footnotetext{#1}%
    \addtocounter{footnote}{-1}%
  \endgroup
}

\input{sections/figures}

\title{
    \texorpdfstring{$\lambda$Split}{lambdaSplit}: Self-Supervised Content-Aware Spectral Unmixing for Fluorescence Microscopy
}

\renewcommand{\shorttitle}{$\lambda$Split}
\author{
Federico Carrara \\
Fondazione Human Technopole, Milan, Italy \\
Università Campus Bio-Medico, Rome, Italy \\
\texttt{federico.carrara@fht.org} \\
\And
Talley Lambert \\
Harvard Medical School, Boston, MA, USA \\
\texttt{talley\_lambert@hms.harvard.edu} \\
\And
Mehdi Seifi \\
Fondazione Human Technopole, Milan, Italy \\
\texttt{mehdi.seifi@fht.org} \\
\And
Florian Jug \\
Fondazione Human Technopole, Milan, Italy \\
\texttt{florian.jug@fht.org}
}

\hypersetup{
pdftitle={λSplit: Self-Supervised Content-Aware Spectral Unmixing for Fluorescence Microscopy},
pdfauthor={Federico Carrara, Talley Lambert, Mehdi Seifi, Florian Jug},
pdfkeywords={Spectral Unmixing, Hierarchical Variational Inference, Fluorescence Microscopy}
}

\date{}

\begin{document}

\maketitle

\blfootnote{Accepted at the 19th European Conference on Computer Vision
(ECCV 2026), Malmö, Sweden, September 8--12, 2026.}

\begin{abstract}
In fluorescence microscopy, spectral unmixing aims to recover individual fluorophore concentrations from spectral images that capture mixed fluorophore emissions. 
Since classical methods operate pixel-wise and rely on least-squares fitting, their performance degrades with increasingly overlapping emission spectra and higher levels of noise, suggesting that a data-driven approach that can learn and utilize a structural prior might lead to improved results. 
Learning-based approaches for spectral imaging do exist, but they are either not optimized for microscopy data or are developed for very specific cases that are not applicable to fluorescence microscopy settings. 
To address this, we propose $\lambda$Split, a physics-informed deep generative model that learns a conditional distribution over concentration maps using a hierarchical Variational Autoencoder. 
A fully differentiable Spectral Mixer enforces consistency with the image formation process, while the learned structural priors enable state-of-the-art unmixing and implicit noise removal. 
We demonstrate $\lambda$Split on $3$ real-world datasets that we synthetically cast into a total of $58$ challenging spectral unmixing benchmarks. 
We compare our results against a total of $10$ baseline methods, including classical methods and a range of learning-based methods. 
Our results consistently show competitive performance and improved robustness in high noise regimes, when spectra overlap considerably, or when the spectral dimensionality is lowered, making $\lambda$Split a new state-of-the-art for spectral unmixing of fluorescence microscopy data. 
Importantly, $\lambda$Split is compatible with spectral data produced by standard confocal microscopes, enabling immediate adoption without specialized hardware modifications.
\end{abstract}

\setcounter{footnote}{0}

\vspace{10mm}
\input{sections/introduction}

\input{sections/related_works}

\input{sections/methods}

\input{sections/data}

\input{sections/results}

\input{sections/discussion}

\paragraph{\bf Disclosure.}
A patent application related to the methods described in this work has been filed.

\section*{Acknowledgements}

We thank all the members of the Jug Group and the National Facility for Bioimage Analysis at Fondazione Human Technopole for their excellent feedback and insightful discussions.
This work was supported by generous core funding from Fondazione Human Technopole.


\input{references}
\appendix
\input{supplements}

\end{document}

%% file: sections/figures.tex
\newcommand{\figOverview}{
\begin{figure}[b]
    \centering
    \includegraphics[width=\linewidth]{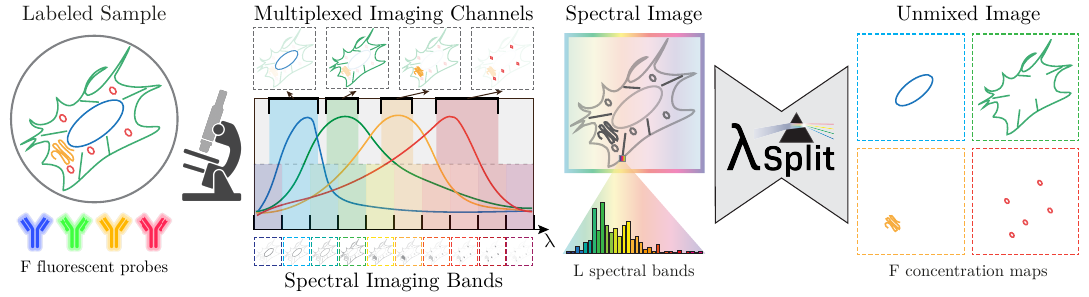}
    \caption{{\bf Spectral Imaging and Unmixing in a nutshell.} 
    A biological sample is labeled with $F$ fluorescent probes (FPs) whose emission spectra typically overlap. 
    While for multiplexed imaging each FP is imaged in isolation using $F$ appropriate band-pass filters (see Multiplexed Imaging Channels), spectral imaging, instead, acquires $L$ spectral bands in which photons of all FPs are unfiltered and therefore mixed (see Spectral Imaging Bands).
    Note that multiplexed imaging can suffer from bleed-through artifacts due to overlapping emission spectra.
    To retrieve all $F$ unmixed concentration maps (right), we propose $\lambda$Split, a self-supervised method that leverages learned structural priors to achieve state-of-the-art results even when spectra overlap considerably, when imaging noise is high, and when the spectral dimensionality $L$ is lowered.}
    \vspace{-3mm}
    \label{fig:spectral_imaging_overview}
\end{figure}
}

\newcommand{\figArchitecture}{
\begin{figure}[bt]
    \centering
    \includegraphics[width=\linewidth]{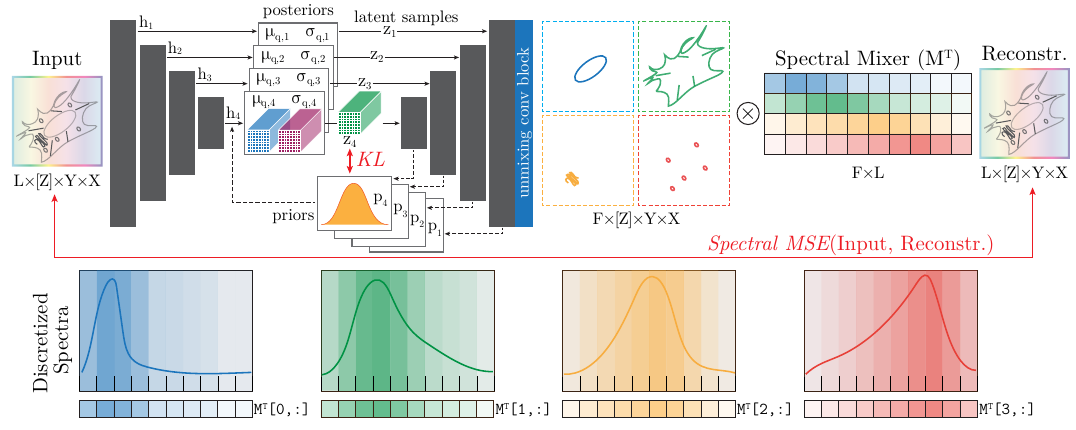}
    \caption{{\bf Proposed architecture of $\lambda$Split.}
    The model builds on an LVAE backbone~\cite{Sonderby2016-zg}, where a bottom-up encoder produces features $h_i$ at multiple hierarchy levels.
    At the highest hierarchy level we employ a default multivariate Gaussian prior, followed by learnable top-down priors $p_i$ (conditioned from above) that, combined with the respective $h_i$, form the final posterior distributions $(\mu_{q,i}, \sigma_{q,i})$. 
    Latent samples are generated by drawing from the topmost Gaussian and successively propagating and resampling down the hierarchy through a deterministic decoder to, finally, generate FP concentration maps. 
    The last decoder layer comprises an unmixing convolutional block (blue) that maps the final activation into the desired unmixed concentration map tensor. 
    To render our method fully self-supervised, a differentiable Spectral Mixer multiplies the predicted concentration maps by the transposed mixing matrix $M^T \in \mathbb{R}^{F \times L}$, obtainable by discretizing the emission spectra of the FPs contained in the imaged sample (bottom).
    This models the physical image formation process and, given that the predicted concentration maps are correct, reconstructs the spectral image we received as input, rendering this setup an end-to-end trainable variational autoencoder. 
    The model is trained by minimizing the sum of the hierarchical KL divergence (computed from latent samples) and a spectral MSE reconstruction loss.}
    \label{fig:architecture}
    \vspace{-4mm}
\end{figure}
}

\newcommand{\figNoiseBioSR}{
\begin{figure}[t]
    \centering
    \includegraphics[width=\linewidth]{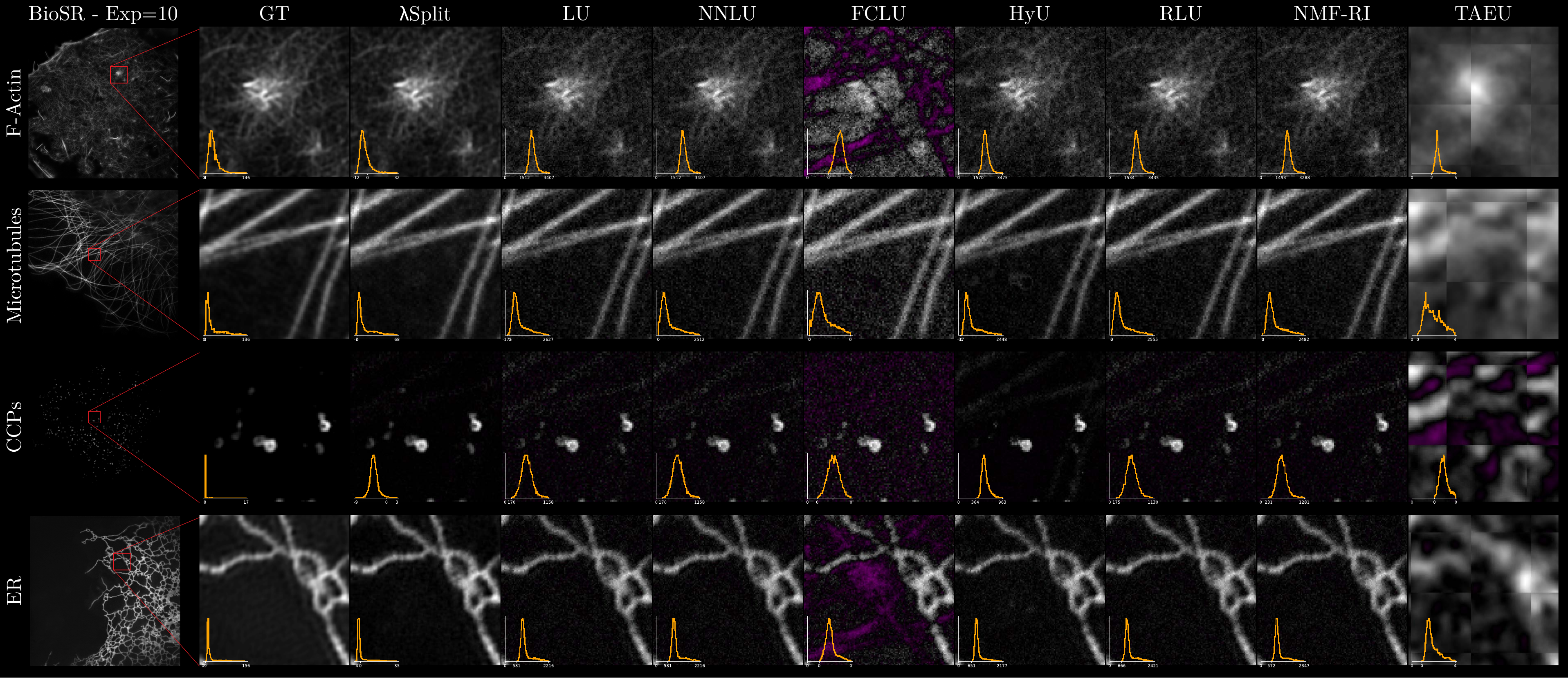}
    \caption{
    {\bf Qualitative results on high-noise BioSR data} (exposure 10\,ms). 
    Compared to $7$ baselines, $\lambda$Split produces the cleanest results, also benefiting from its implicit noise removal capabilities. 
    Objects and details (\eg, puncta in the CCPs channel or F-Actin structures) are well recovered, enabling best possible downstream analyses.
    }
    \label{fig:SNR_exp_BioSR}
\end{figure}
}

\newcommand{\figNoiseCellAtlas}{
\begin{figure}[t]
    \centering
    \includegraphics[width=\linewidth]{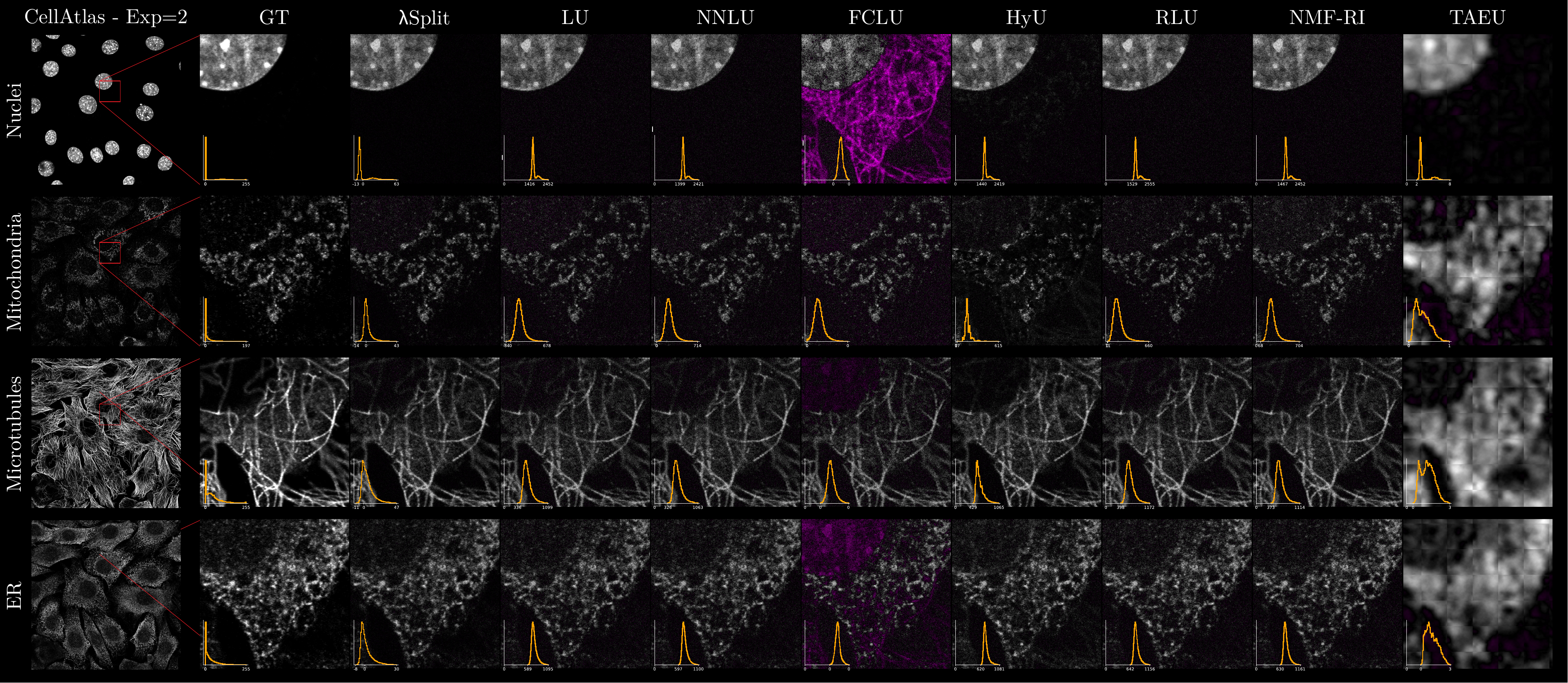}
    \caption{{\bf Qualitative results on high-noise CellAtlas data} (exposure 2\,ms). 
    Compared to $7$ baselines, $\lambda$Split produces the cleanest results, also benefiting from its implicit noise removal capabilities. 
    Still, in this example, we see that very high levels of noise cannot be fully denoised solely by relying on $\lambda$Split's implicit capabilities, suggesting that future work should also focus on incorporating explicit denoising modules~\cite{Krull2020-ap,Prakash2021-zj,Prakash2021-qx}.
    Objects and details (\eg, puncta and roundish structures in the mitochondria and ER channels or elongated microtubule structures) are well recovered, facilitating possible downstream analyses.}
    \label{fig:SNR_exp_CellAtlas}
\end{figure}
}

\newcommand{\figNoiseHHMI}{
\begin{figure}[t]
    \centering
    \includegraphics[width=\linewidth]{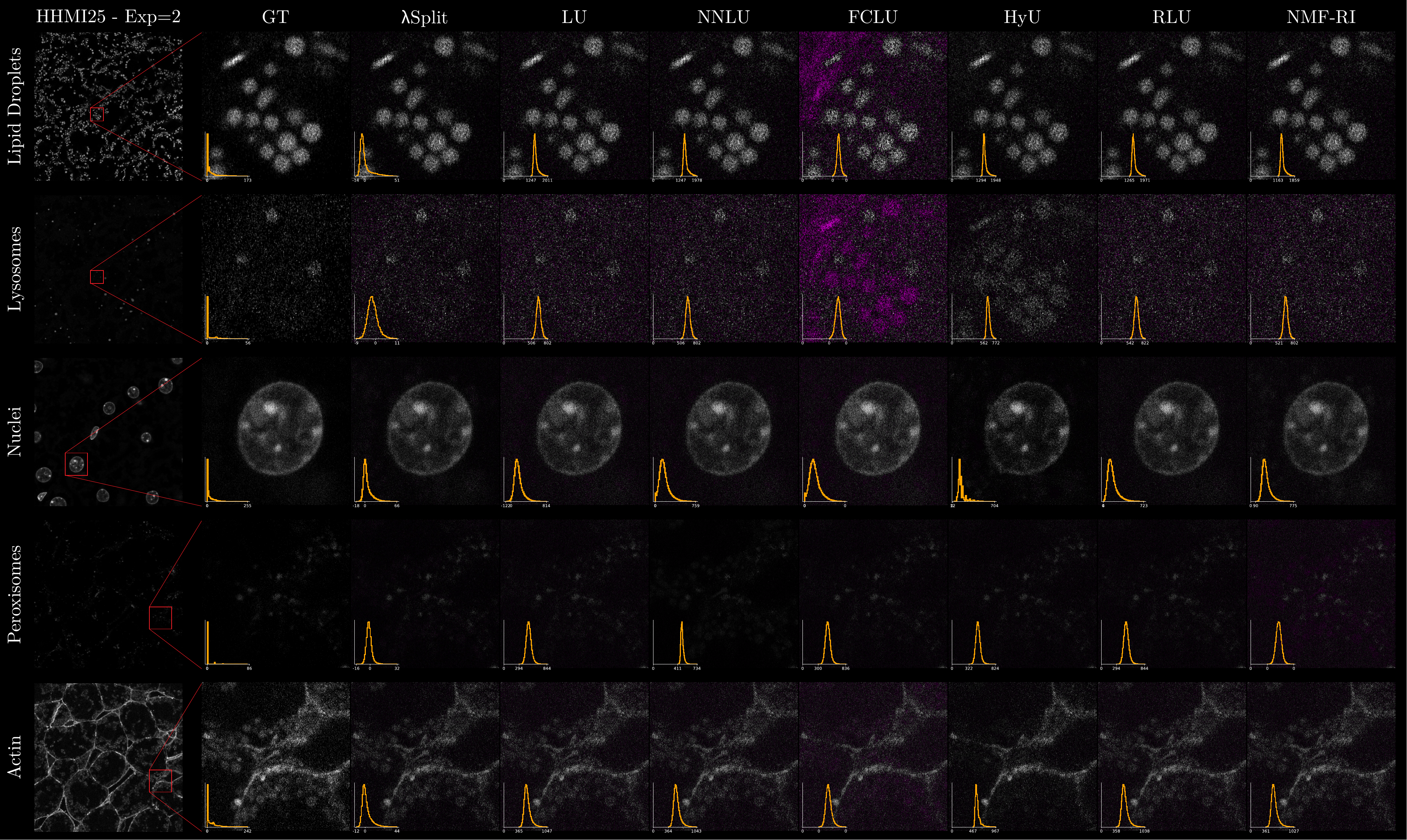}
    \caption{
    {\bf Qualitative results on high-noise HHMI25 data} (exposure 2\,ms). 
    Compared to $6$ baselines, $\lambda$Split produces the cleanest results, also benefiting from its implicit noise removal capabilities. 
    Still, in this example, we see that very high levels of noise cannot be fully denoised solely by relying on $\lambda$Split's implicit capabilities, suggesting that future work should also focus on incorporating explicit denoising modules~\cite{Krull2020-ap,Prakash2021-zj,Prakash2021-qx}
    }.
    \label{fig:SNR_exp_HHMI25}
\end{figure}
}

\newcommand{\figOverlapECthreetwobands}{
\begin{figure}[t]
    \centering
    \includegraphics[width=\linewidth]{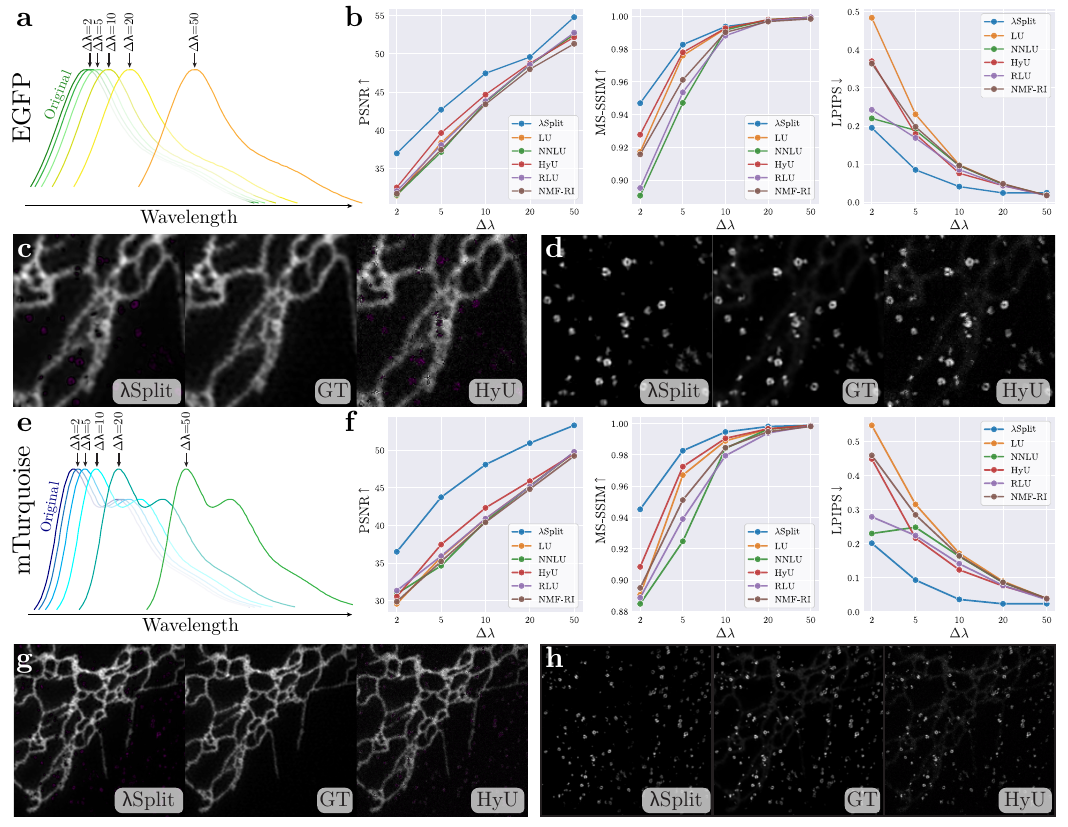}
    \caption{
    {\bf Spectral unmixing with controlled spectral overlap (32 bands).} 
    For this experiment, we use the ER and CCPs structures from the BioSR dataset, with 32 spectral bands imaged. 
    {\bf a,e}~Synthetically shifted EGFP and mTurquoise emission spectra, respectively.
    While we fix the spectrum associated with ER, the emission spectrum of CCPs is rigidly shifted with increasing $\Delta\lambda$. 
    {\bf b,f}~Quantitative metrics over $\Delta\lambda$, showing that $\lambda$Split (blue) consistently outperforms all other baselines and that the performance gap widens when spectral overlap is larger (hence, when $\Delta\lambda$ is smaller).
    {\bf c,d,g,h}~Representative crops from unmixing results for $\Delta\lambda=2$\,nm by $\lambda$Split and the best-performing baseline (\wrt PSNR), plus the corresponding ground truth (GT) in the center. 
    In c,d, we show ER, while in g,h, we show CCPs.
    }
    \label{fig:overlap_EC_32bands}
\end{figure}
}

\newcommand{\figOverlapMCthreetwobands}{
\begin{figure}[t!!!]
    \centering
    \includegraphics[width=\linewidth]{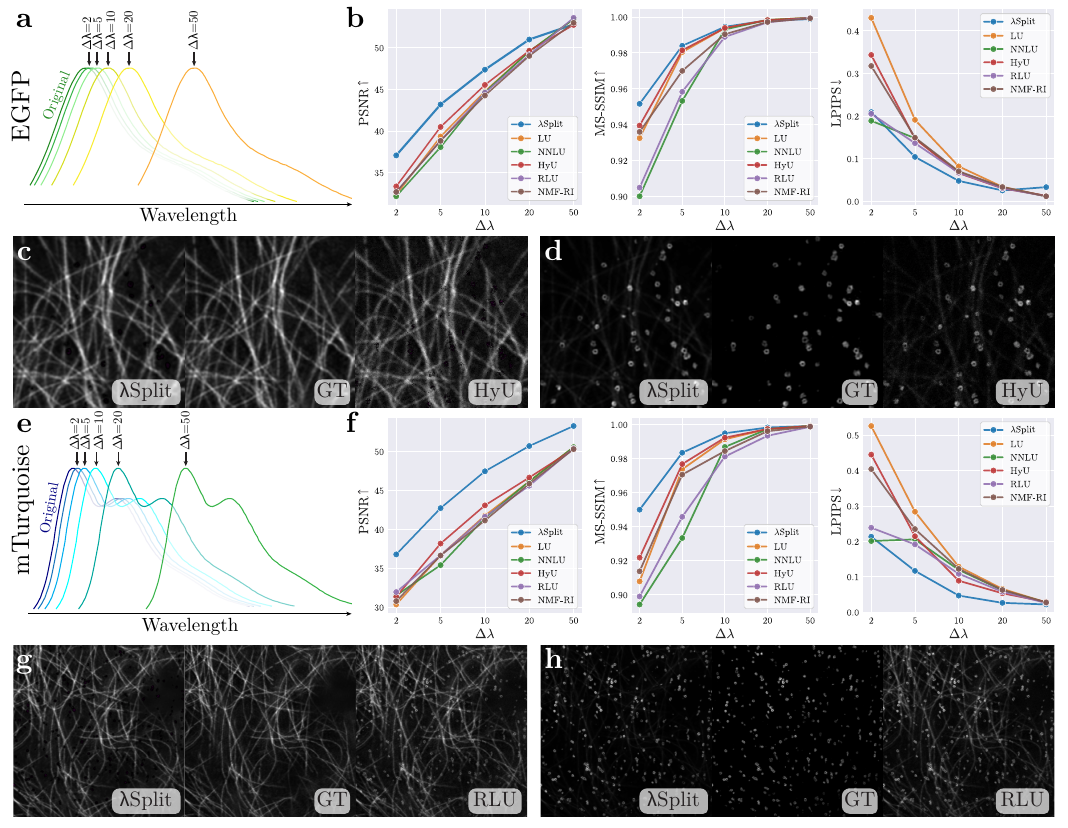}
    \caption{
    {\bf Spectral unmixing with controlled spectral overlap (32 bands).} 
    For this experiment, we use the Microtubules and CCPs structures from the BioSR dataset, with 32 spectral bands imaged. 
    {\bf a,e}~Synthetically shifted EGFP and mTurquoise emission spectra, respectively. 
    While we fix the spectrum associated with Microtubules, the emission spectrum of CCPs is rigidly shifted with increasing $\Delta\lambda$. 
    {\bf b,f}~Quantitative metrics over $\Delta\lambda$, showing that $\lambda$Split (blue) consistently outperforms all other baselines and that the performance gap widens when spectral overlap is larger (hence, when $\Delta\lambda$ is smaller).
    {\bf c,d,g,h}~Representative crops from unmixing results for $\Delta\lambda=2$\,nm by $\lambda$Split and the best-performing baseline (\wrt PSNR), plus the corresponding ground truth (GT) in the center.
    In c,d, we show Microtubules, while in g,h, we show CCPs.
    }
    \vspace{-4mm}
    \label{fig:overlap_MC_32bands}
\end{figure}
}

\newcommand{\figOverlapECfivebands}{
\begin{figure}[t]
    \centering
    \includegraphics[width=\linewidth]{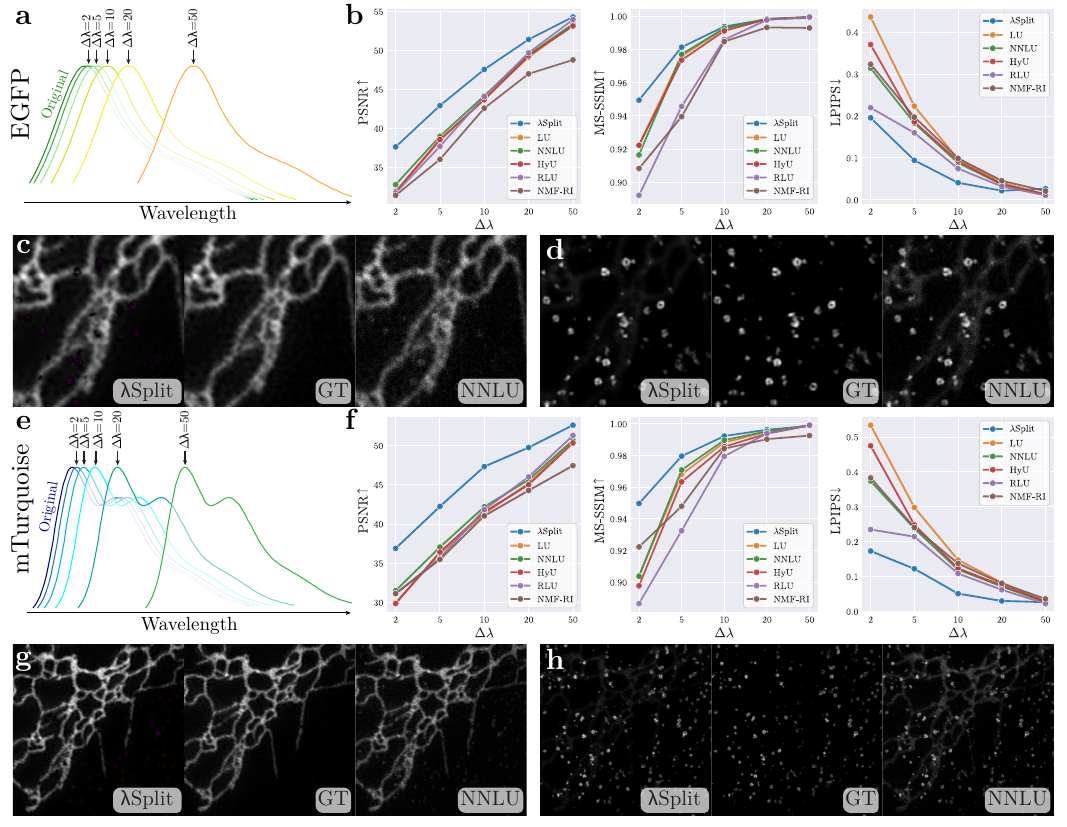}
    \caption{
    {\bf Spectral unmixing with controlled spectral overlap (5 bands).} 
    For this experiment, we use the ER and CCPs structures from the BioSR dataset, with 5 spectral bands imaged. 
    {\bf a,e}~Synthetically shifted EGFP and mTurquoise emission spectra, respectively.
    While we fix the spectrum associated with ER, the emission spectrum of CCPs is rigidly shifted with increasing $\Delta\lambda$. 
    {\bf b,f}~Quantitative metrics over $\Delta\lambda$, showing that $\lambda$Split (blue) consistently outperforms all other baselines and that the performance gap widens when spectral overlap is larger (hence, when $\Delta\lambda$ is smaller). 
    {\bf c,d,g,h}~Representative crops from unmixing results for $\Delta\lambda=2$\,nm, by $\lambda$Split and the best-performing baseline (\wrt PSNR), plus the corresponding ground truth (GT) in the center.
    In c,d, we show ER, while in g,h, we show CCPs.
    }
    \vspace{-8mm}
    \label{fig:overlap_EC_5bands}
\end{figure}
}

\newcommand{\figOverlapMCfivebands}{
\begin{figure}[t]
    \centering
    \includegraphics[width=\linewidth]{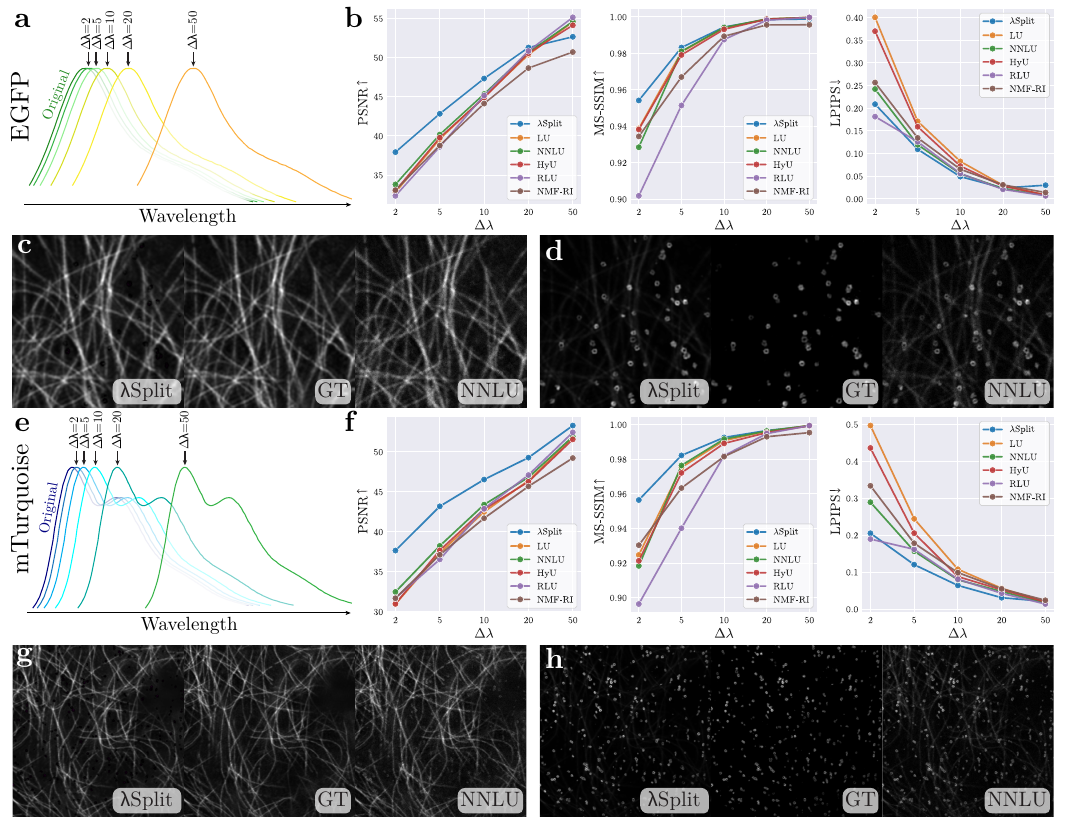}
    \caption{
    {\bf Spectral unmixing with controlled spectral overlap (5 bands).} 
    For this experiment, we use the Microtubules and CCPs structures from the BioSR dataset, with 5 spectral bands imaged. 
    {\bf a,e}~Synthetically shifted EGFP and mTurquoise emission spectra, respectively.
    While we fix the spectrum associated with Microtubules, the emission spectrum of CCPs is rigidly shifted with increasing $\Delta\lambda$. 
    {\bf b,f}~Quantitative metrics over $\Delta\lambda$, showing that $\lambda$Split (blue) consistently outperforms all other baselines and that the performance gap widens when spectral overlap is larger (hence, when $\Delta\lambda$ is smaller). 
    {\bf c,d,g,h}~Representative crops from unmixing results for $\Delta\lambda=2$\,nm, by $\lambda$Split and the best-performing baseline (\wrt PSNR), plus the corresponding ground truth (GT) in the center.
    In c,d, we show Microtubules, while in g,h, we show CCPs.
    }
    \vspace{-8mm}
    \label{fig:overlap_MC_5bands}
\end{figure}
}

\newcommand{\figFourFluorophores}{
\begin{figure}[t]
    \centering
    \includegraphics[width=\linewidth]{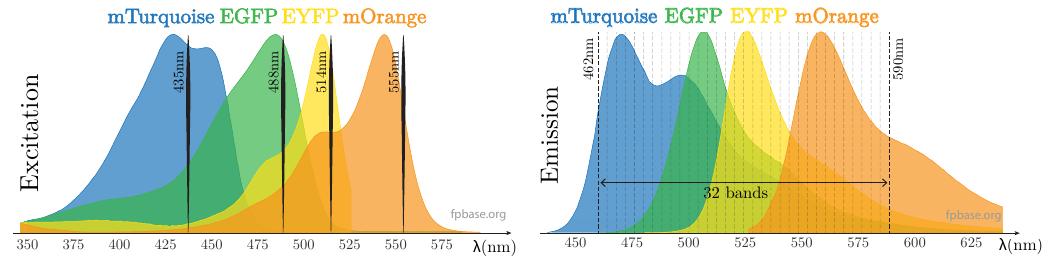}
    \caption{
    {\bf Four FP setup for CellAtlas and BioSR datasets for the \textit{Spectral Unmixing and Noisy Data} experiment.} 
    On the left, we report the excitation spectra of the four fluorophores together with the positions of the four excitation lasers. 
    The number of lasers, their placement, and their relative power are selected by solving an optimization problem to achieve balanced excitation across all fluorophores, accounting for their excitation and emission efficiency (excitation cross-sections and quantum yields). 
    On the right, we display the corresponding emission spectra alongside the 
    placement of the 32 spectral detection bands used for imaging.
    }
    \label{fig:4FPs_setup}
\end{figure}
}

\newcommand{\figFiveFluorophores}{
\begin{figure}[t]
    \centering
    \includegraphics[width=\linewidth]{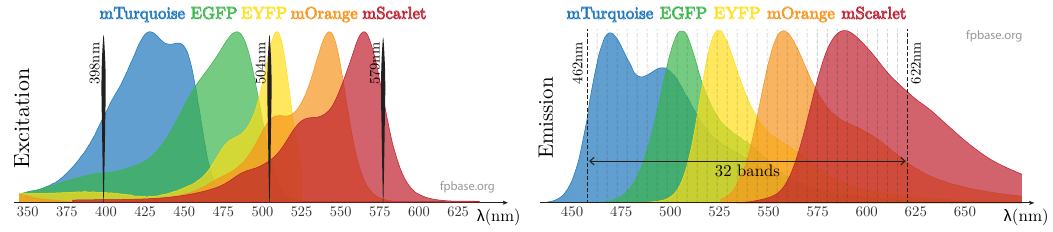}
    \caption{
    {\bf Five FP setup for HHMI25 dataset for the \textit{Spectral Unmixing and Noisy Data} experiment.} 
    On the left, we report the excitation spectra of the five fluorophores together with the positions of the three excitation lasers. 
    The number of lasers, their placement, and their relative power are selected by solving an optimization problem to achieve balanced excitation across all fluorophores, accounting for their excitation and emission efficiency (excitation cross-sections and quantum yields).
    On the right, we display the corresponding emission spectra alongside the placement of the 32 spectral detection bands used for imaging.
    }
    \label{fig:5FPs_setup}
\end{figure}
}

\newcommand{\figMicrosimPipeline}{
\begin{figure}[t]
    \centering
    \includegraphics[width=\linewidth]{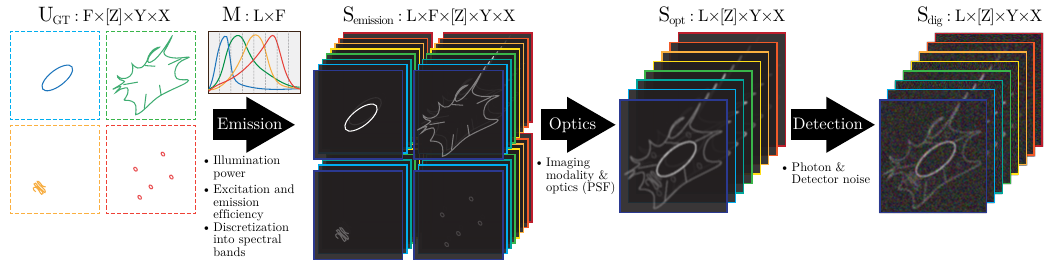}
    \caption{
        {\bf Overview of the spectral fluorescence microscopy simulation pipeline based on \texttt{microsim~\cite{Lambert2026-em}}.} 
        Multi-channel inputs are interpreted as fluorophore concentration maps $U_{\mathrm{GT}}$, where each channel corresponds to a fluorescent probe (FP) with a known emission spectrum. 
        First, for each FP, a noise-free spectral emission volume $S_{\mathrm{emission}}$ is generated by combining concentration maps with their discretized emission spectra $M$, while modeling tunable illumination power and fluorophore-specific emission and excitation efficiency. 
        Next, microscope-specific optical effects, such as point-spread functions (PSFs), are applied, and per-fluorophore emission volumes are combined to obtain an optical spectral volume $S_{\mathrm{opt}}$ consistent with the chosen imaging modality. 
        Finally, the detection process introduces realistic noise sources, including photon and detector noise, producing the final digital spectral image $S_{\mathrm{dig}}$. 
    }
    \label{fig:simulation_pipeline}
\end{figure}
}

\newcommand{\figAutounmixOverlap}{
\begin{figure}[t]
    \centering
    \includegraphics[width=0.9\linewidth]{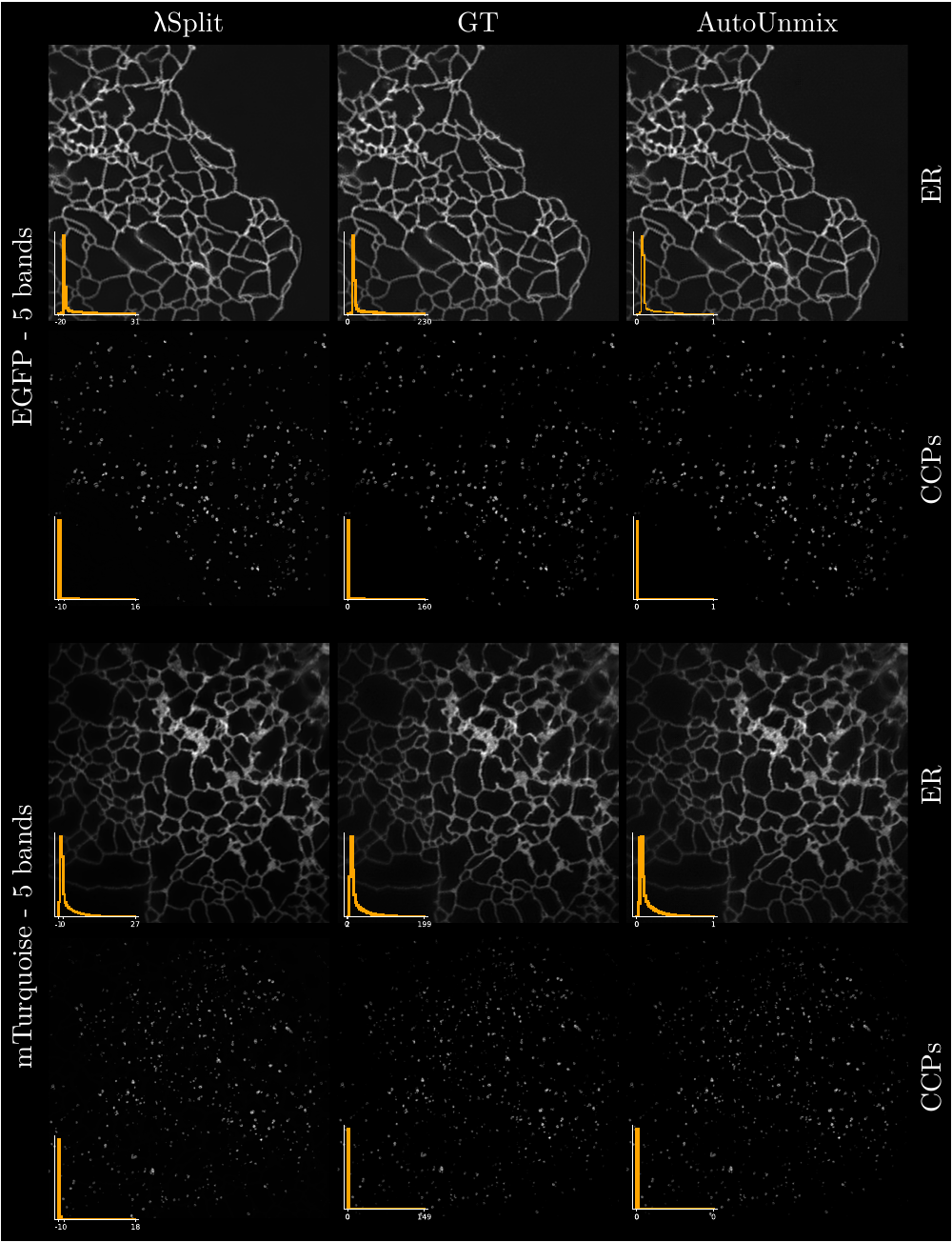}
    \caption{
    \textbf{Qualitative comparison with AutoUnmix for increasing spectral overlap experiments.}
    Representative unmixed concentration maps of ER and CCP structures on the BioSR dataset with 5 spectral bands are shown for $\lambda$Split and AutoUnmix at $\Delta\lambda = 50$.
    Because of the absence of "negative" structures in this case, we display the unmixed images using the standard grayscale colormap.
    At this level of spectral separation, $\lambda$Split produces results comparable to AutoUnmix both visually and according to the corresponding intensity histograms, despite operating without supervision or explicit denoising.
    }
    \label{fig:autounmix_overlap}
\end{figure}
}

\newcommand{\figAutounmixUNetLowBands}{
\begin{figure}[t]
    \centering
    \includegraphics[width=\linewidth]{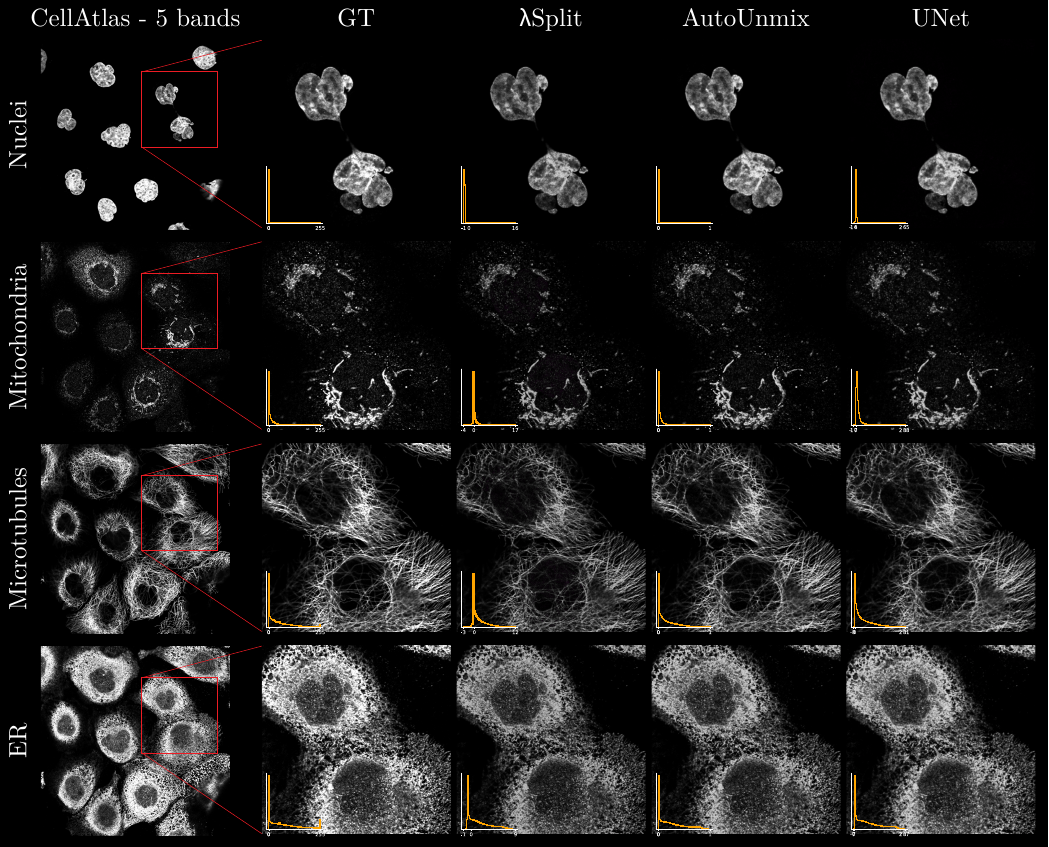}
    \caption{
    {\textbf{Qualitative results for supervised baselines for the reduced spectral dimensionality experiments}} (CellAtlas, 5-band acquisition).
    We compare $\lambda$Split against two supervised baselines, AutoUnmix and UNet.
    Both baselines benefit from strong supervision through paired noise-free GT unmixed targets, and both produce high-quality unmixing results that are visually very close to each other.
    $\lambda$Split predictions remain structurally consistent and faithfully recover the underlying biological structures without any GT supervision, although they may appear perceptually less contrasted than those of the supervised baselines.
    }
    \label{fig:autounmix_unet_num_bands}
\end{figure}
}

\newcommand{\figLowBandsQuali}{
\begin{figure}[t]
    \centering
    \includegraphics[width=\linewidth]{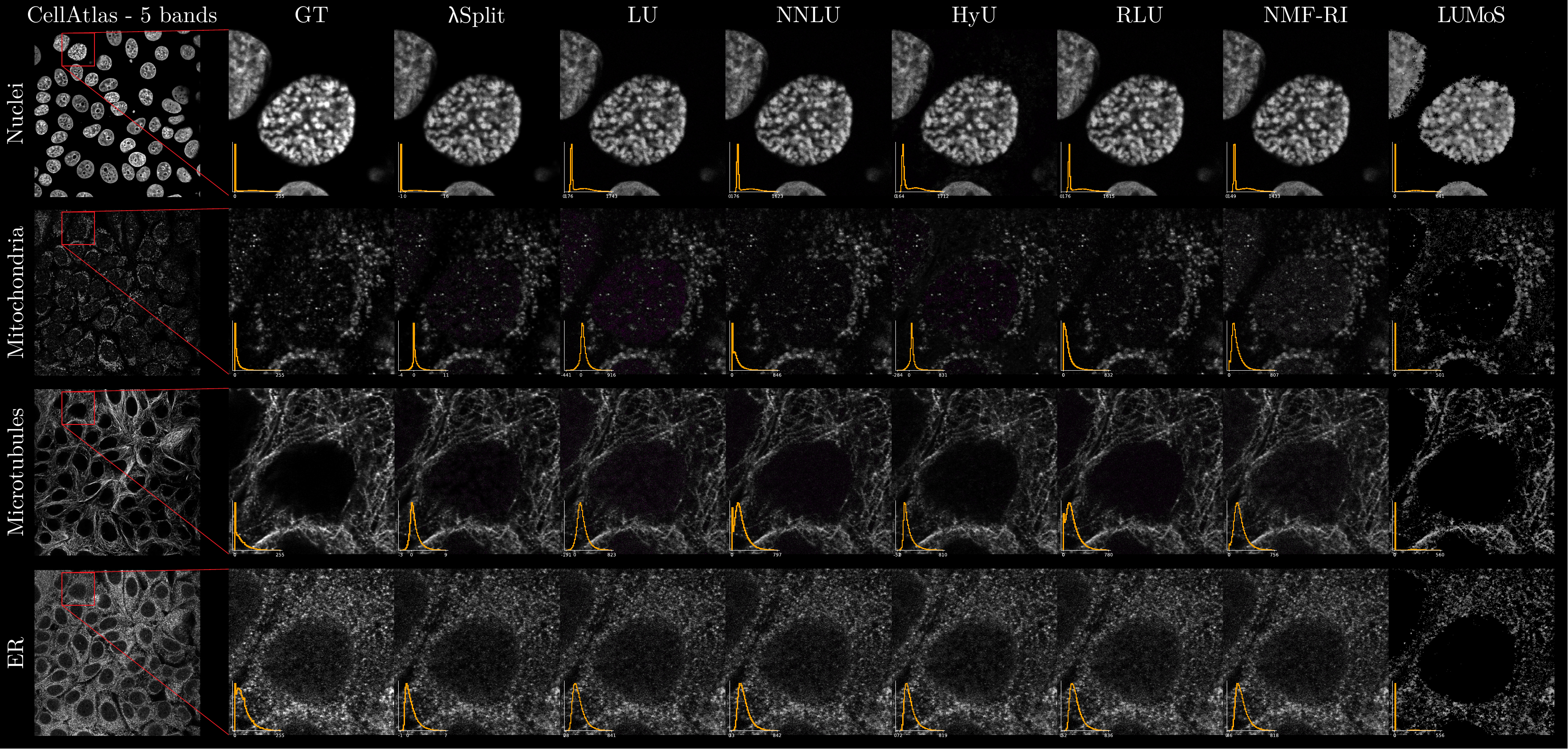}
    \caption{
    \textbf{Qualitative results for the reduced spectral dimensionality experiments} (CellAtlas, 5-band acquisition).
    We compare $\lambda$Split against six baselines on representative structures.
    Although $\lambda$Split achieves the best quantitative fidelity, its predictions do not always appear as high-contrast as those of some baselines.
    This behavior can be attributed to two design choices.
    First, the network is trained solely with an MSE objective and does not include perceptual losses, which could be explored in future work to improve visual appearance.
    Second, $\lambda$Split does not impose constraints on the unmixed images (e.g., non-negativity or sparsity), allowing the model to fully exploit learned structural priors.
    While this flexibility improves reconstruction fidelity, spectral overlap can induce intensity redistribution across channels.
    For instance, bright structures in one channel may appear as faint negative counterparts in another, reducing perceived contrast in our figure while still yielding the best MSE.
    }
    \label{fig:exp_num_bands}
\end{figure}
}

\newcommand{\figSNRLinehist}{
\begin{figure}[tb]
    \centering
    \vspace{-8mm}
    \includegraphics[width=\linewidth]{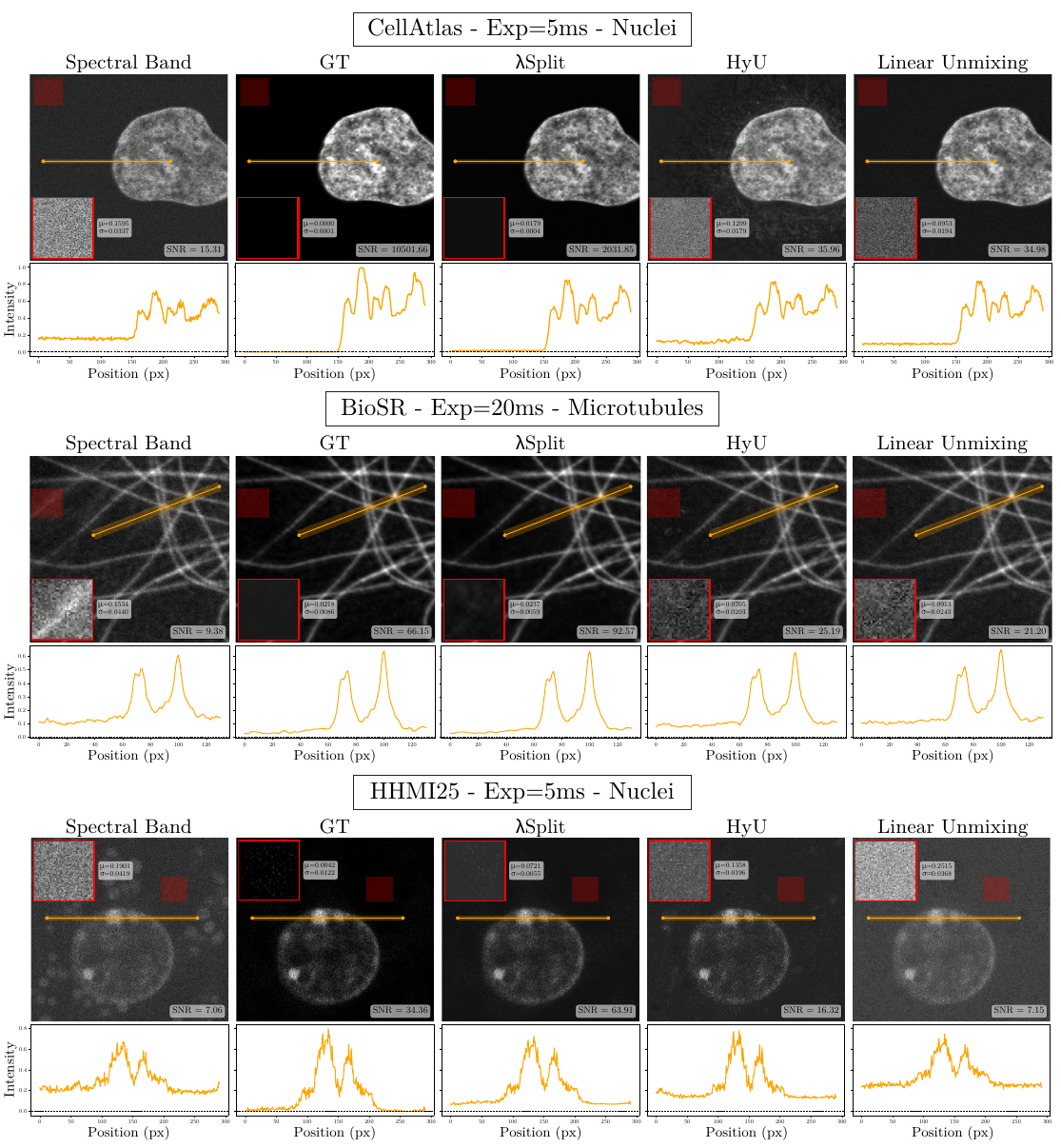}
    \vspace{-5mm}
    \caption{
    {\bf Analysis of the SNR computation and the implicit denoising capability of $\lambda$Split. }
    For each dataset used in the increasing noise set of experiments, we show a representative structure and exposure level. 
    In each top row, from left to right, we report example crops from: $(i)$~the spectral band corresponding to the emission peak of the FP associated with the structure; $(ii)$~the ground truth (GT) concentration map; $(iii)$~the concentration map predicted by $\lambda$Split; $(iv)$~the baseline methods achieving the highest SNR in that setting; $(v)$~the concentration map obtained from linear unmixing.
    All crops are min-max normalized to the same intensity range  and displayed using a standard grayscale colormap.
    For each image, a background patch used for SNR estimation is highlighted (red square); the content of the extracted patch is displayed in a corner, together with its mean ($\mu$) and standard deviation ($\sigma$); the resulting SNR value is also reported. 
    Notice that the SNR values differ from those reported in the result tables, as they are computed only on a small sample crop rather than the entire dataset.
    The orange profiles show intensity along the annotated 10-pixel thick line and highlight the denoising effect of $\lambda$Split.
    Indeed, we observe a reduction in noise, as evidenced by fewer oscillations in the line histogram and a larger gap separating the signal peaks from the background level.
    }
    \label{fig:SNR_linehist}
\end{figure}
}

\newcommand{\figAblationBeta}{
\begin{figure}[htb!]
    \centering
    \includegraphics[width=\linewidth]{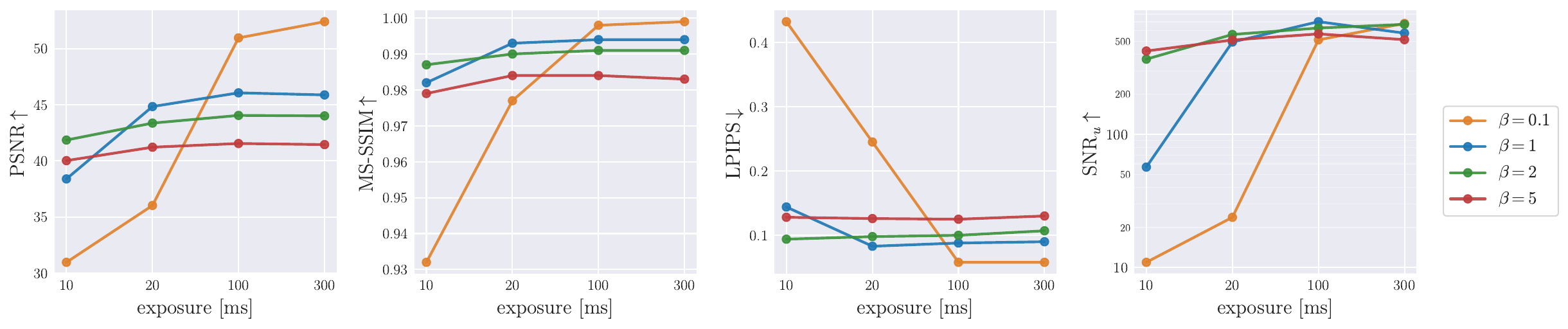}
    \caption{
    {\bf Ablation on the KL loss weight $\beta$.} 
    Reconstruction quality as a function of exposure time (shorter exposure $\rightarrow$ lower SNR) on BioSR, for $\beta\in\{0.1,1,2,5\}$; our default $\beta=1$ is shown in blue.
    We report PSNR, MS-SSIM, LPIPS, and the SNR of the unmixed images ($\mathrm{SNR_u}$, log scale); higher is better except for LPIPS. 
    Weak regularization ($\beta=0.1$) preserves detail at high SNR but fits noise at low SNR, whereas stronger regularization denoises more aggressively at the cost of high-SNR fidelity.
    }
    \label{fig:beta_ablation}
\end{figure}
}

\newcommand{\figAblationLevels}{
\begin{figure}[htb!]
    \centering
    \includegraphics[width=\linewidth]{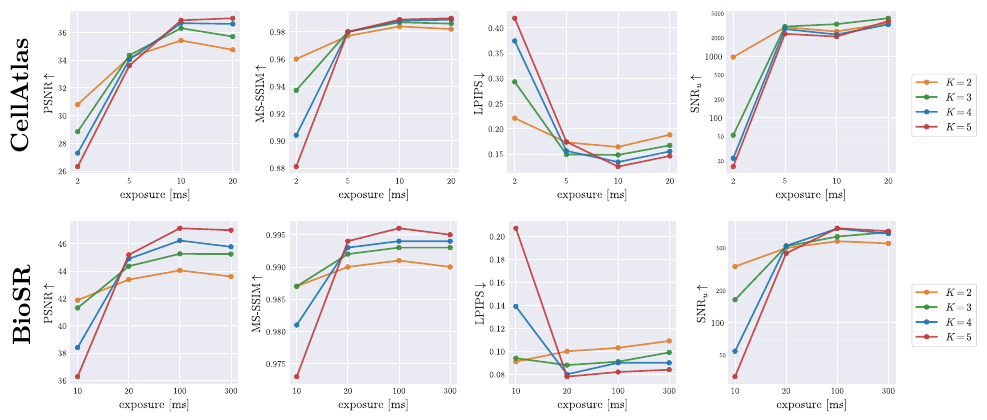}
    \caption{
    {\bf Ablation on the number of hierarchical levels $K$. }
    Reconstruction quality as a function of exposure time (shorter exposure $\rightarrow$ lower SNR) on CellAtlas (top) and BioSR (bottom), for $K\in\{2,3,4,5\}$ stochastic LVAE levels; our default $K=4$ is shown in blue. 
    We report PSNR, MS-SSIM, LPIPS, and the SNR of the unmixed images ($\mathrm{SNR_u}$, log scale); higher is better except for LPIPS. 
    On both datasets, deeper hierarchies perform best at high SNR while shallower ones are more robust under noise, confirming the fidelity--denoising tradeoff.
    }
    \label{fig:levels_ablation}
\end{figure}
}

%% file: sections/introduction.tex
\section{Introduction}
\label{sec:intro}
Fluorescence microscopy is a fundamental tool for biomedical discovery in the life sciences, enabling us to visualize selected biological structures labeled with fluorescent markers called fluorescent probes (FPs). 
The most common way to image biological samples is arguably multiplexed imaging, where multiple FPs labeling different structures in a sample are imaged one at a time using appropriate bandpass filtering (see~\cref{fig:spectral_imaging_overview}). 
Due to the emission properties of FPs, the limited spectral range of visible light, and the available microscopy hardware, multiplexed imaging is, in practice, used to image 2 to 4 structures in their respective image channels. 
In addition, sequential acquisitions increase the required imaging time and, for live samples, result in channels that do not correspond to the same time point.
Last but not least, repeated rounds of imaging can lead to increased light exposure and, therefore, a higher degree of photobleaching and phototoxic effects, which are naturally undesired~\cite{Lichtman2005-ht, Shaner2005-kz, Icha2017-nc, Waters2014-hp}.
These limitations motivate other approaches to image multiple fluorescently labeled structures in a biological sample, \ie, spectral imaging approaches.

\figOverview

Spectral fluorescence microscopy does not image FPs one at a time but instead collects all emitted light at once~\cite{Hiraoka2002-kj, Zimmermann2003-ja}. 
To later recover the individual FP emissions (fluorophore concentrations), the emitted light is collected in a fixed number of spectral bands.
Here, the signal at each spatial location is not a single intensity value but a spectral measurement that quantifies how much light was collected over a set of wavelength intervals (bands), as shown in \cref{fig:spectral_imaging_overview}. 

Spectral imaging shifts the burden from hardware-based channel separation to computational separation of fluorophore emissions. 
This separation is commonly addressed through spectral unmixing, where classical approaches estimate fluorophore contributions independently at each pixel using least-squares fitting to reference spectra~\cite{Zimmermann2014-nl}. 
While conceptually simple, such pixel-wise formulations degrade when spectra overlap considerably, when noise levels are high, or when the spectral dimensionality is lowered (\ie, when fewer bands are imaged).
Learning-based approaches offer an alternative by incorporating spatial context and data-driven structural priors. 
However, existing methods either do not account for the optics and noise characteristics of fluorescence microscopy or are tailored to narrowly defined acquisition settings.

To address these limitations, we introduce $\lambda$Split, a physics-informed deep generative framework for spectral unmixing in fluorescence microscopy that leverages a fully differentiable spectral mixing module (\ie, the Spectral Mixer), which enables self-supervised training. 
Our approach achieves robust unmixing and implicit denoising across challenging imaging conditions, as we demonstrate experimentally on a total of $58$ benchmarks we defined for this work.

Importantly, $\lambda$Split operates directly on spectral acquisitions produced by standard point-scanning confocal microscopes, enabling immediate integration into existing experimental workflows without hardware modifications.

\section{Problem Setting}
\label{sec:problem_definition}
A spectral image can be represented as $S \in \mathbb{R}^{L \times Z \times Y \times X}$, where $(Z, Y, X)$ denotes the spatial dimensions and $L$ the spectral dimension, \ie, the number of spectral bands. 
Depending on the microscope configuration, $L$ can vary substantially: some systems provide relatively dense spectral sampling with a number of contiguous, equispaced, narrow bands of up to $32$, while others rely on coarser spectral measurements, leading to only $4$ or $5$ bands that can even be flexibly tuned to cover arbitrary wavelength intervals.

The measured spectral signal at each spatial location arises from the union of all photons emitted by FPs that are present at that location. 
Because FPs typically exhibit relatively broad and overlapping emission spectra (see, \eg, \cref{fig:4FPs_setup}), the signal in each spectral band contains contributions from multiple fluorescent sources.

The goal of spectral unmixing is therefore to recover, from an observed spectral acquisition $S$, a set of images representing the spatial concentration maps of the individual FPs, also referred to as the unmixed images $U$~\cite{Zimmermann2014-nl}. 
The spectral image formation process can be described by a linear mixing model:
\begin{equation} \label{eq:problem_definition}
S = M U + \varepsilon,
\end{equation}
where $U \in \mathbb{R}^{F \times Z \times Y \times X}$ denotes the unknown fluorophore concentration maps for $F$ FPs, $M \in \mathbb{R}^{L \times F}$ is the mixing matrix whose columns correspond to the discretized emission spectra of the fluorophores, and $\varepsilon$ represents pixel noise. 
In~\cref{eq:problem_definition}, tensors are multiplied following the Einstein summation convention.

The unmixing problem at hand is inherently an ill-posed inverse problem. 
Overlap between emission spectra makes the unmixing harder, and together with additional noise introduces ambiguities for this inverse problem.
Additionally, a limited number of spectral bands (lower spectral dimensionality) can render the system underdetermined, particularly in regimes where $L < F$. 
These challenges are further amplified in realistic biological samples, where fluorophores may be spatially co-localized, and experimental conditions can induce spatial and spectral variability.

%% file: sections/related_works.tex
\section{Related Work}
\label{sec:relworks}

\paragraph{\textbf{Classical spectral unmixing.}}
A widely used approach is linear unmixing~(LU), which estimates fluorophore concentration maps by solving a least-squares problem given reference spectra~\cite{Zimmermann2014-nl}. 
Common variants, non-negative LU (NNLU) and fully constrained LU (FCLU), incorporate additional constraints such as non-negativity and sum-to-one priors~\cite{Bioucas-Dias2012-kl,Heinz2001-fx,Slawski2011-nt}. 
While efficient and widely adopted in fluorescence microscopy, these pixel-wise formulations remain sensitive to noise, spectral overlap, and reduced spectral dimensionality. 
Alternative approaches, such as Richardson--Lucy unmixing (RLU)~\cite{Kumar2025-iy} and hybrid phasor-based methods (HyU)~\cite{Chiang2023-vm} improve robustness in noisy regimes while still operating pixel by pixel.

\paragraph{\textbf{Blind spectral unmixing.}}
Blind unmixing methods estimate both emission spectra and fluorophore concentrations directly from spectral data. 
A common approach is non-negative matrix factorization (NMF)~\cite{Lee1999-zk}, including variants developed for microscopy such as NMF-RI~\cite{Ortiz-de-Solorzano2013-oc,Jimenez-Sanchez2020-mh}. 
These approaches still rely on pixel-wise LU once spectra are estimated.
Other approaches formulate unmixing as a mutual information minimization problem between target structures, such as PICASSO and Mosaic-PICASSO~\cite{Seo2022-td,Cang2024-pq}. 
Still, their practical applicability remains limited to cases where the spectral dimensionality matches the number of target structures. 


\paragraph{\textbf{Learning-based approaches.}}
Learning-based spectral unmixing methods have recently emerged in both microscopy and remote sensing. 
In fluorescence microscopy, clustering-based approaches such as LUMoS~\cite{McRae2019-cn} and supervised convolutional models such as AutoUnmix~\cite{Jiang2023-fm} and UNMIX-ME~\cite{Smith2020-hr} have been proposed. 
More broadly, a large body of work in remote sensing explores deep architectures for hyperspectral unmixing, including autoencoder-based models, convolutional networks exploiting spatial context, and model-inspired networks that incorporate the mixing process into the decoder or unroll iterative optimization schemes~\cite{Palsson2018-iw,Su2019-ov,Palsson2021-lv,Hong2022-fr,Rasti2022-ec,Rasti2022-ky,Gao2022-nj,Ghosh2022-fm}. 
Despite conceptual similarities, the microscopy setting differs substantially due to its high spatial resolution, limited spectral dimensionality, and distinct noise characteristics.

\paragraph{\textbf{Content-aware models in microscopy.}}
Recent work has shown that content-aware priors can substantially improve inverse problems in fluorescence microscopy~\cite{Weigert2018-jr,Krull2018-id,Krull2020-ap,Prakash2021-qx,Prakash2021-zj}. 
MicroSplit~\cite{Ashesh2026-da}, for example, uses hierarchical variational architectures to separate multiple structures from a single fluorescence image. 
While this approach does not model spectral mixing, it highlights the potential of learned structural priors for decomposing microscopy images.

\medskip
\noindent
A more extensive discussion of related work and methodological differences can be found in the supplementary material (\cref{sec:suppl_relworks}).

%% file: sections/methods.tex
\section{Methods}
\label{sec:methods}

\figArchitecture

Our approach builds on three key components. 
First, we introduce a convolutional hierarchical variational framework based on a Ladder Variational Autoencoder (LVAE) backbone~\cite{Sonderby2016-zg}, replacing pixel-wise estimation with a spatially aware model that learns multi-scale structural priors from data. 
The variational formulation implicitly regularizes the inverse problem, models a conditional distribution over fluorophore concentration maps, and improves robustness under noise, spectral overlap, and varying acquisition configurations. 
Second, we incorporate a fixed, fully differentiable spectral mixing module that implements the physics-based linear image formation model by mapping predicted concentration maps into spectral space using known emission profiles (see~\cref{fig:architecture}). 
This enables self-supervised training directly from mixed spectral observations while maintaining interpretability and general applicability across acquisition setups. 
Third, we perform extensive validation using a high-fidelity spectral simulation pipeline and three real multiplexed microscopy datasets, constructing $58$ controlled unmixing benchmarks that systematically vary photon noise, spectral overlap, and spectral dimensionality, including underdetermined regimes. 
Across these settings, we demonstrate improved robustness and stability compared to classical and learning-based baselines.

As introduced in \cref{sec:problem_definition}, given a spectral image $S \in \mathbb{R}^{L \times Z \times Y \times X}$, our goal is to recover FP concentration maps $U \in \mathbb{R}^{F \times Z \times Y \times X}$.
Here, our objective is to learn a conditional generative model $p_\theta(U|S)$ that captures a distribution over plausible unmixed solutions while remaining consistent with the physically grounded linear mixing process introduced in \cref{eq:problem_definition}. To enforce this consistency, we introduce a differentiable Spectral Mixer that reconstructs a spectral image $\hat S$ directly from the generated unmixed output $U$ via multiplication with the spectral mixing matrix $M \in \mathbb{R}^{L \times F}$. The column $M_{:,j} = [m_{1j},\,m_{2j},\,\dots,\,m_{Lj}]^\top$ of the mixing matrix encodes the discretized spectrum of the fluorophore $j$, with entries $m_{\ell j}$ obtained by averaging the continuous emission profile over the $\ell$\nobreakdash-th spectral band. In practice, these spectra can either be obtained from public databases such as FPBase~\cite{Lambert2019-jk} or measured directly from dataset-specific single-fluorophore control samples. We further $\ell_1$-normalize each spectrum such that $\sum_{\ell=1}^{L} m_{\ell j}=1$, ensuring that the Spectral Mixer preserves the total intensity between spectral and unmixed images, \ie,
\begin{equation}
    \label{eq:sum-to-one}
    \sum_{\ell=1}^{L} \hat{S}_{\ell,p} 
    =\sum_{\ell=1}^{L} \sum_{j=1}^{F} m_{\ell j} U_{j,p}
    =\sum_{j=1}^{F} U_{j,p},
\end{equation}
for all spatial locations $p$.

\paragraph{\textbf{Modified ELBO and LVAE Loss for Spectral Unmixing.}} To model the generative conditional distribution $p_\theta(U|S)$, we train a hierarchical VAE, specifically an LVAE~\cite{Sonderby2016-zg}.
Our objective is therefore to maximize the conditional log-likelihood
\begin{equation}
\label{eq:likelihood}
    \arg\max_\theta \sum_{i=1}^{N} \log p_\theta(U_i \mid S_i),   
\end{equation}
where $\theta$ denotes the decoder parameters defining the generative distribution.
By introducing hierarchical latent variables $\mathbf{z}=\{z_1,\dots,z_K\}$, where $K$ is the number of hierarchy levels, and the variational posterior parameterized by the encoder $q_\phi(\mathbf{z}\mid S)$ with parameters $\phi$, we obtain the standard variational evidence lower bound (ELBO), which, according to our notation, reads as
\begin{equation}
\label{eq:ELBO}
    \mathbb{E}_{q_\phi(\mathbf{z}|S)}\big[\log p_\theta(U|\mathbf{z})\big]
    -
    \mathrm{KL}\big(q_\phi(\mathbf{z}|S)\,\|\,p_\theta(\mathbf{z})\big).
\end{equation}
A detailed derivation of the ELBO is presented in~\cref{subsec:elbo_derivation}.
As the unmixed FP concentration channels $U$ are not directly observable, and to recover a self-supervised autoencoder setup in which the reconstructed output can be compared to the spectral measurement given as the original input, we map $U$ back to the spectral domain as a reconstructed image $\hat S$, using the previously introduced Spectral Mixer.
This fully differentiable spectral mixing layer also ensures consistency with the physical image formation process. 

Hence, the spectral observation is now obtained through a fixed, deterministic mixing process,
$S = g(U) = MU$.
Given latent variables $\mathbf{z}$ sampled from the variational posterior $q_\phi(\mathbf{z}\mid S)$, the decoder predicts fluorophore concentration maps $U(\mathbf{z}) \sim p_\theta(U\mid \mathbf{z})$, which are subsequently mapped to the spectral domain to generate a reconstructed spectral image $\hat S(\mathbf{z}) = MU(\mathbf{z})$.

Since the spectral formation process is deterministic and we do not assume a probabilistic noise model, the conditional likelihood $p_\theta(S\mid \mathbf z)$ cannot be directly specified in closed form. 
In practice, we approximate the corresponding reconstruction term using the spectral mean squared error (MSE) between the observed spectral image and its reconstruction, 

\begin{equation}
    \mathcal{L}_{sp\mathrm{MSE}}\big(S, \hat{S} \big)
    =
    \frac{1}{LP}\sum_{\ell=1}^{L}\sum_{p=1}^{P}\big(S_{\ell,p} - \hat{S}_{\ell,p} \big)^2,
\end{equation}
where $P = Z\cdot Y\cdot X$ is the total number of voxels in the image.
The resulting training objective is therefore defined as
\begin{equation}
\label{eq:lambdasplit_loss}
    \mathcal{L}
    =
    \mathbb{E}_{q_\phi(\mathbf{z}\mid S)}
    \Big[
    \mathcal{L}_{sp\mathrm{MSE}}\big(S, \hat S(\mathbf{z})\big)
    \Big]
    +
    \mathrm{KL}\big(q_\phi(\mathbf{z}\mid S)\,\|\,p_\theta(\mathbf{z})\big).
\end{equation}

\paragraph{\textbf{Implementation Details.}} 
As mentioned before, we employ an LVAE architecture~\cite{Sonderby2016-zg}. 
Since it allows sampling from the posterior distribution over unmixed images, we infer predictions by averaging $50$ posterior samples, giving us an approximate MMSE estimate~\cite{Prakash2021-qx}. 
We employ a latent hierarchy with $4$ stochastic levels, for which the spatial resolution is progressively reduced by factors of two, while the feature channel dimensionality is kept constant at $128$.
The KL loss is computed per latent level as a Monte~Carlo
estimate over the batch by averaging the KL across all entries of the latent tensors, and is then averaged across levels.
The final KL term is weighted by $\beta = 1$.
The choice of the number of stochastic levels and the KL weight $\beta$ is supported by ablation studies reported in the supplementary material~\cref{sec:ablations}.

Our implementation can operate on 2D and 3D spectral data by employing 2D and 3D convolutional layers, respectively.
While training is performed on randomly extracted patches of size $64\times 64$ for 2D inputs and $8\times 64\times 64$ for 3D data, at inference time we employ inner tiling~\cite{Ashesh2022-ah}, with one-quarter overlap between neighboring tiles.

Models are trained using the Adamax~\cite{Kingma2014-hy} optimizer with default hyperparameters. 
We use a batch size of $32$ and an initial learning rate of $10^{-3}$. 
Training is performed for $150$ epochs using mixed precision and early stopping with patience set to 30 epochs. 
The learning rate is reduced on plateau with a patience of $15$ epochs.
To preserve the relative intensity between spectral bands, input spectral images are normalized using global mean–standard deviation normalization.
All models were trained on an NVIDIA DGX H200 system using either an 18GB or 35GB MIG partition for 2D and 3D models, respectively. 
Training time depends on the dataset size and dimensionality, typically ranging from $2$ hours to approximately one day.

\paragraph{\textbf{Baselines.}} 
We compare $\lambda$Split against a set of classical and learning-based spectral unmixing baselines. 
As direct linear methods based on least-squares, we include linear unmixing (LU), non-negative LU (NNLU)~\cite{Slawski2011-nt}, and fully-constrained LU (FCLU)~\cite{Heinz2001-fx}. We further evaluate HyU~\cite{Chiang2023-vm}, Richardson--Lucy Unmixing (RLU)~\cite{Kumar2025-iy}, and NMF-RI~\cite{Jimenez-Sanchez2020-mh}, which are specifically developed for fluorescence microscopy and low-photon settings. 
We additionally consider learning-based methods. 
For low-band settings ($\leq5$ bands), we compare to LUMoS clustering-based unmixing~\cite{McRae2019-cn}. 
Additionally, we include the Transformer Unmixing Autoencoder (TAEU)~\cite{Ghosh2022-fm}, developed for spectral unmixing in remote sensing. 
The motivation for choosing this algorithm is its popularity and success in the field, as well as the availability of an open-source implementation. 
We also compare $\lambda$Split against two supervised DL-based methods: a standard supervised UNet~\cite{Ronneberger2015-ei} and AutoUnmix~\cite{Jiang2023-fm}. 

Please find additional information about baselines and their implementation in \cref{subsec:suppl_baseline_details}.

\paragraph{\textbf{Evaluation Metrics.}} We evaluate unmixing performance using a combination of intensity-based, structural, perceptual, and signal-quality metrics. We report range-invariant \textit{\textbf{PSNR}} and \textit{\textbf{MS-SSIM}} (abbreviated as MS3IM), as introduced in~\cite{Weigert2018-jr}, which compensate for global intensity scaling differences between prediction and ground truth (\eg, due to different output normalization). 
Moreover, we include the \textit{\textbf{Pearson}} correlation coefficient, \textit{\textbf{LPIPS}}~\cite{Zhang2018-xx}, and \textit{\textbf{SNR}} of the predicted unmixed images. 
Finally, we also report \textit{\textbf{MicroMS-SSIM}}~\cite{Ashesh2024-ev} (abbreviated as $\mu$MS3IM), which is a structural similarity metric tailored to microscopy data. More details about the implementation of the employed metrics are reported in \cref{subseq:suppl_eval_metrics}.

%% file: sections/data.tex
\section{Data and Data Handling}
\label{sec:data}

\paragraph{\textbf{Data Simulation Pipeline.}}
Since ground-truth~(GT) fluorophore concentration maps are typically unavailable (and even unobtainable) in microscopy settings, we rely on synthetic data to enable controlled benchmarking across acquisition conditions with variable noise, spectral overlap, and spectral dimensionality. 
To generate realistic spectral mixtures, we use a powerful microscopy simulation pipeline based on the open-source \texttt{microsim} Python package~\cite{Lambert2026-em}. 
Multi-channel inputs, either real multiplexed fluorescence images or multiclass segmentation maps, are interpreted as spatial fluorophore concentration fields and serve as GT for evaluation. 
Each channel is associated with a fluorophore emission spectrum, from which a noise-free spectral volume is generated. 
The simulator then models optical effects (\eg, point spread functions) and the image acquisition process, including photon statistics, readout noise, and configurable spectral band placements that emulate realistic spectral microscopy setups. 
This pipeline enables the controlled generation of physically consistent spectral mixtures across diverse experimental settings. 
Additional details and a schematic overview are provided in the supplementary material (\cref{fig:simulation_pipeline}).

\paragraph{\textbf{Real-world Datasets.}}
We use three publicly available fluorescence microscopy datasets as inputs for our simulation pipeline. 
The \textbf{\textit{BioSR}} dataset~\cite{Qiao2024-ih} consists of 2D high-SNR fluorescence images (1004$\times$1004 pixels) of four organelles (microtubules, F-actin, endoplasmic reticulum (ER), and clathrin-coated pits (CCPs)). 
These images provide structurally well-resolved organelle distributions suitable for controlled spectral simulations. The \textbf{\textit{CellAtlas}} dataset contains four-channel multiplexed images from the Human Protein Atlas project~\cite{Uhlen2015-mw, Ouyang2019-ax}, comprising tens of thousands of 2D images across diverse cell types. 
In this work, we select a subset of $300$, 2048$\times$2048 images where the four channels correspond to nuclei, microtubules, endoplasmic reticulum, and mitochondria. 
Finally, the HHMI-D25-8bit dataset introduced in~\cite{Ashesh2026-da}, here referred to as \textbf{\textit{HHMI25}}, includes 3D multiplexed fluorescence volumes (17$\times$2048$\times$2048) acquired from mouse liver tissue samples. Each volume contains five labeled structures: lipid droplets, nuclei, peroxisomes, cortical actin, and lysosomes, enabling us to conduct volumetric spectral simulations.

%% file: sections/results.tex
\section{Experiments and Results}
\label{sec:results}
We evaluate $\lambda$Split across a series of controlled spectral unmixing benchmarks designed to isolate the main sources of difficulty in microscopy data. 
Specifically, we analyze robustness to noise, increasing spectral overlap between FP emissions, and reduced spectral dimensionality (fewer spectral bands). 
In all cases, we report quantitative and qualitative~\footnote{
In $\lambda$Split, as in all other spectral unmixing baselines, some of the unmixed channels could exhibit negative structures to compensate for extra-brightness in the other channels. 
For cleaner visual comparison, in all figures we use a custom colormap centered on the background value, with positive pixels shown in grayscale and negative ones in magenta tones, all with the same scaling.
} 
results against the selected baselines.

\subsection{Spectral Unmixing and Noisy Data}
\label{sec:results_more_noise}
\figNoiseBioSR
\input{tables/table_1_SNR_CellAtlas}

To assess robustness to pixel noise, we evaluate all applicable baselines on simulated 32-band spectral data derived from the BioSR, CellAtlas, and HHMI25 datasets. 
To simulate a realistic setting, we assigned four partially overlapping fluorophores (mTurquoise, EGFP, EYFP, and mOrange) to the four structures (fluorescent channels) in the BioSR and CellAtlas data. 
For the HHMI25 data, which has 5 channels, we added mScarlet. 
The emission spectra of all used FPs are shown in \cref{fig:4FPs_setup,fig:5FPs_setup}.

For each dataset, we simulate four exposure levels with increasing Poisson noise while keeping Gaussian readout noise fixed at realistic magnitudes. 
The highest signal-to-noise ratio (SNR) configuration is selected such that baseline methods perform well, representing a well-conditioned regime in which classical methods perform comparably to $\lambda$Split.
As shown in \cref{tab:results_cellatlas,tab:results_biosr,tab:results_hhmi25}, $\lambda$Split continues to perform well as SNR decreases, consistently outperforming all baselines. 
Note that $\lambda$Split's reported SNR$_u$ values are much higher than those of other baseline methods. This is a result of the implicit denoising capabilities of our proposed method.
An analysis supporting this observation is reported in the supplement (see \cref{subsec:implicit_denoising}).

Qualitative comparisons also reveal that pixel-wise methods naturally propagate the noise in the input into the unmixed channels.
Results obtained with $\lambda$Split, in comparison, can discriminate the imaged structures from purely stochastic pixel fluctuations, preserving boundary sharpness and leading to denoised predictions (see \cref{fig:SNR_exp_BioSR}). 

Finally, we observe that the data-driven TAEU~\cite{Ghosh2022-fm} and classical FCLU~\cite{Heinz2001-fx} baselines, both created for spectral unmixing in remote sensing applications, seem unsuitable for application to fluorescence microscopy data, leading to sub-par results in all our experiments. 
For this reason, and to save space, we do not report results for these two baselines in subsequent sections.

\subsection{Spectral Unmixing and Overlapping Spectra}
\label{sec:results_more_overlap}
\figOverlapMCthreetwobands

Next, we ask how unmixing performance changes as the overlap between the used FPs increases. 
Everything else remains unchanged, allowing us to isolate the effect of spectral overlap.
To this end, we conducted a total of $40$ experiments where we simulated spectral data, just as before, but moved the emission spectra closer together without changing their shape.

More concretely, we have used two pairs of the BioSR structures (Microtubules\,+\,CCPs and ER\,+\,CCPs), and assigned to both the same emission spectrum (either EGFP or mTurquoise).
For the second structure (CCPs), we have then rigidly shifted its spectrum by $\Delta\lambda \in \{2\text{nm},5\text{nm},10\text{nm},20\text{nm},50\text{nm}\}$ (see \cref{fig:overlap_MC_32bands}\textbf{a,c} for an illustration). 

Further, we conducted these experiments using 32- and 5-band acquisition regimes, in line with existing, commonly used commercial microscopy setups.
To keep the total photon budget fixed, we left the laser power and exposure times (20\textit{ms}) unchanged in both cases, which leads to higher per-band SNR when fewer are being imaged. 
This, again, leads to an initial regime (for $\Delta\lambda=50\text{nm}$) in which classical baselines lead to results comparable to those of $\lambda$Split.

For smaller values of $\Delta\lambda$, the quality gap of predictions consistently widens, showing that $\lambda$Split can effectively utilize its learned structural prior in these more difficult settings (see \cref{fig:overlap_MC_32bands} and \cref{fig:overlap_EC_32bands,fig:overlap_EC_5bands,fig:overlap_MC_5bands} and \cref{tab:overlap_EGFP_EC_merged,tab:overlap_egfp_mc_merged,tab:ss_bins_mturq_ec_merged,tab:ss_bins_mturq_mc_merged}), where it consistently outperforms all other baselines.

\subsection{Spectral Unmixing and Spectral Dimensionality} 
\label{sec:results_fewer_bands}
\input{tables/table_5_low_num_bands}

Next, we studied the effect of reducing the number of spectral bands.
While classical methods desire the number of bands to be at least equal to the number of unknowns (\ie, the number of used FPs), having a higher number of bands offers numerical advantages.

We conducted two sets of experiments on the CellAtlas data:
$(i)$~we take the highest SNR setup from \cref{sec:results_more_noise} for the 32 band setup (note that the respective column in \cref{tab:results_low_num_bands} reports the same values as \cref{tab:results_cellatlas}).
We then reduced the number of bands to 5, 4, and 3 while keeping the per-band SNR constant.
This removes SNR as a confounder, isolating the effect of spectral dimensionality, and
$(ii)$~we take the total photon budget of the 5-band setup from above and simulate a 3-, 4-, and 32-band setup. 
Note that the 5-band results in \cref{tab:results_low_num_bands} are, for this reason, also identical.
The question we are asking here is, therefore, how the potential downside of having fewer bands balances with the advantage of having them at a higher SNR.

Overall, we observe that, as spectral dimensionality is reduced, $\lambda$Split consistently outperforms the baselines in PSNR and MS-SSIM while remaining competitive in perceptual quality (evaluated via LPIPS). 
It is important to note that when the number of bands drops below the number of unknowns (\ie, where the number of structures to be unmixed is larger than the spectral dimension), the inverse problem becomes underdetermined, and the performance of classical least-squares methods (\ie, LU or HyU) drops significantly (see the columns showing results for 3 bands in \cref{tab:results_low_num_bands}). 
In this regime, $\lambda$Split leverages its learned structural prior and consistently leads to state-of-the-art results.

\subsection{λSplit \vs Supervised Spectral Unmixing}
\label{sec:results_us_vs_supervised}
Finally, we compare $\lambda$Split with a supervised UNet and AutoUnmix, a supervised convolutional neural network for spectral unmixing in fluorescence microscopy. 
For both baselines, we use noise-free ground-truth (GT) concentration maps as targets, which is possible because of our data simulations. 
Note that this enables supervised baselines to also learn to perform supervised denoising, giving them an additional advantage. 

In the low spectral dimensionality experiments (\cref{tab:autounmix_bands,fig:autounmix_unet_num_bands}), we immediately observe that the performance discrepancy between $\lambda$Split and AutoUnmix progressively decreases as the number of bands increases: richer spectral sampling provides $\lambda$Split with more information about the emission profiles, while AutoUnmix benefits less from additional bands since the supervision signal and overall signal content remain essentially unchanged. 

Conversely, in the increasing spectral overlap experiments (\cref{tab:autounmix_overlap_EGFP-EC,tab:autounmix_overlap_EGFP-MC,tab:autounmix_overlap_mTurquoise-EC,tab:autounmix_overlap_mTurquoise-MC,fig:autounmix_overlap}), we observe that, as the overlap reduces, $\lambda$Split matches or surpasses AutoUnmix.

Still, AutoUnmix is not a practical alternative for $\lambda$Split. 
On one hand, supervised training data is not available in practice; we could only train it due to our synthetic data generation setup.
On the other hand, AutoUnmix (84M parameters, compared to $\lambda$Split's ~3M) processes every pair of spectral bands with a shared UNet, so its compute and memory grow quadratically with the number of bands (10 UNet passes per image at 5 bands, 496 at 32 bands).

Finally, the supervised UNet, trained on the same noise-free targets, performs on par with AutoUnmix in both experiments (see~\cref{fig:autounmix_unet_num_bands}).
It outperforms $\lambda$Split for every reduced number of bands and for strongly overlapping spectra ($\Delta\lambda \leq 10$\,nm), while $\lambda$Split matches or surpasses it as the overlap decreases.
As for AutoUnmix, however, these results rely on paired GT concentration maps, which are not available in practice.

In summary, high-fidelity predictions can be obtained with a significantly more compact, physics-consistent, and self-supervised framework that can be instantly applied to real-world imaging setups. 

%% file: tables/table_1_SNR_CellAtlas.tex
\begin{table}[t!!!] 
\centering
\caption{
\textbf{Quantitative results on the \textit{CellAtlas} data for various noise levels.}
Columns, from left to right, report:
$(i)$~exposure times (less causing more noise);
$(ii)$~the resulting SNR of the spectral input images (aggregated over all spectral bands); 
$(iii)$~baselines considered for the set of experiments;
$(iv)$~quality metrics, \ie, PSNR, MS-SSIM (MS3IM), MicroMS-SSIM ($\mu$MS3IM), LPIPS, Pearson correlation coefficient, and average SNR of unmixed images (see main text for details). 
Higher values are better, except for LPIPS.
In each metric column, the best result is shown in \textbf{bold}, while the second best is \uline{underlined}.
}
\label{tab:results_cellatlas}
\renewcommand{\arraystretch}{1.0}
\footnotesize

\begin{tabular*}{\linewidth}{lll@{\extracolsep{\fill}}cccccc}
\toprule
\textbf{Exp} & \textbf{SNR$_{sp}$} & \textbf{Method}
& PSNR $\uparrow$
& MS3IM $\uparrow$
& $\mu$MS3IM $\uparrow$
& LPIPS $\downarrow$
& Pearson $\uparrow$
& SNR$_{u}$ $\uparrow$ \\
\midrule

\multirow{8}{*}{2\textit{ms}}
& \multirow{8}{*}{5.33}
& LU~\cite{Zimmermann2014-nl}    & 23.96 & 0.799 & \secondval{0.647} & 0.485 & 0.779 & 8.44 \\
& & NNLU~\cite{Slawski2011-nt}  & 24.06 & \secondval{0.803} & \secondval{0.647} & \secondval{0.439} & 0.785 & 8.59 \\
& & FCLU~\cite{Heinz2001-fx}  & 20.70 & 0.520 & 0.619 & 0.511 & 0.540 & 5.04 \\
& & HyU~\cite{Chiang2023-vm}   & 24.28 & \secondval{0.803} & 0.500 & 0.442 & \secondval{0.795} & \secondval{11.35} \\
& & RLU~\cite{Kumar2025-iy}   & 23.98 & 0.799 & \secondval{0.647} & 0.442 & 0.781 & 8.47 \\
& & NMF-RI~\cite{Jimenez-Sanchez2020-mh} & 24.09 & \secondval{0.803} & 0.474 & 0.440 & 0.790 & 8.69 \\
& & TAEU~\cite{Ghosh2022-fm}  & \secondval{24.52} & 0.761 & \bestval{0.652} & 0.520 & 0.800 & \bestval{60.79} \\
& & $\lambda$Split (ours) & \bestval{27.14} & \bestval{0.904} & 0.625 & \bestval{0.373} & \bestval{0.885} & 21.35 \\
\midrule

\multirow{8}{*}{5\textit{ms}}
& \multirow{8}{*}{11.04}
& LU       & 28.52 & 0.921 & 0.792 & 0.378 & 0.920 & 20.33 \\
& & NNLU   & 28.55 & 0.921 & 0.721 & 0.319 & 0.921 & 20.34 \\
& & FCLU   & 22.03 & 0.595 & 0.607 & 0.435 & 0.685 & 9.57 \\
& & HyU    & 27.52 & 0.892 & 0.790 & 0.369 & 0.897 & 29.28 \\
& & RLU    & \secondval{28.91} & \secondval{0.927} & 0.659 & 0.316 & \secondval{0.929} & 20.50 \\
& & NMF-RI & 28.86 & \secondval{0.927} & \secondval{0.795} & \secondval{0.315} & 0.926 & 20.91 \\
& & TAEU   & 23.97 & 0.735 & 0.655 & 0.520 & 0.786 & \secondval{66.67} \\
& & $\lambda$Split (ours) & \bestval{34.03} & \bestval{0.980} & \bestval{0.979} & \bestval{0.155} & \bestval{0.975} & \bestval{2047.10} \\
\midrule

\multirow{8}{*}{10\textit{ms}}
& \multirow{8}{*}{22.57}
& LU       & 32.56 & 0.966 & 0.834 & 0.307 & 0.961 & 41.38 \\
& & NNLU   & 32.64 & 0.967 & 0.834 & 0.228 & 0.962 & 41.35 \\
& & FCLU   & 22.52 & 0.621 & 0.638 & 0.404 & 0.716 & 14.76 \\
& & HyU    & 29.84 & 0.929 & 0.650 & 0.274 & 0.926 & \secondval{62.78} \\
& & RLU    & \secondval{33.14} & \secondval{0.971} & 0.850 & \secondval{0.221} & \secondval{0.968} & 41.48 \\
& & NMF-RI & 32.99 & \secondval{0.971} & \secondval{0.839} & 0.224 & 0.967 & 42.93 \\
& & TAEU   & 24.26 & 0.767 & 0.694 & 0.452 & 0.801 & 101.82 \\
& & $\lambda$Split (ours) & \bestval{36.66} & \bestval{0.989} & \bestval{0.988} & \bestval{0.137} & \bestval{0.986} & \bestval{2041.18} \\
\midrule

\multirow{8}{*}{20\textit{ms}}
& \multirow{8}{*}{43.05}
& LU       & 35.91 & 0.985 & 0.696 & 0.194 & 0.987 & 75.87 \\
& & NNLU   & 36.02 & 0.986 & 0.696 & 0.128 & 0.988 & 75.78 \\
& & FCLU   & 22.38 & 0.614 & 0.626 & 0.361 & 0.720 & 21.75 \\
& & HyU    & 30.37 & 0.936 & 0.873 & 0.184 & 0.953 & \secondval{109.65} \\
& & RLU    & \bestval{36.68} & \secondval{0.988} & 0.907 & \secondval{0.125} & \secondval{0.990} & 75.83 \\
& & NMF-RI & \secondval{36.49} & \bestval{0.989} & \secondval{0.703} & \bestval{0.124} & 0.989 & 77.96 \\
& & TAEU   & 23.66 & 0.739 & 0.675 & 0.479 & 0.797 & 781.03 \\
& & $\lambda$Split (ours) & 36.46 & \secondval{0.988} & \bestval{0.988} & 0.155 & \bestval{0.988} & \bestval{2697.59} \\
\bottomrule
\end{tabular*}
\vspace{-4mm}
\end{table}

%% file: tables/table_5_low_num_bands.tex
\begin{table*}[t!!!]
\centering
\caption{
{\bf Quantitative results on \textit{CellAtlas} for varying spectral dimensionality.}
We evaluate spectral unmixing performance when reducing the number of acquisition bands.
Two experimental settings are considered:
(\textit{left}) the number of bands is reduced while keeping the per-band SNR constant, isolating the effect of spectral dimensionality; 
(\textit{right}) the total photon budget is fixed, such that fewer bands lead to higher per-band SNR.
Find more details in \cref{sec:results_fewer_bands}.
Best results are shown in \textbf{bold}, second best \uline{underlined}.
}
\label{tab:results_low_num_bands}
\vspace{2mm}
\renewcommand{\arraystretch}{1.0}
\setlength{\tabcolsep}{5.0pt}
\footnotesize

\begin{tabular*}{\linewidth}{@{\extracolsep{\fill}}llcccccccc}
\toprule
& & \multicolumn{4}{c}{\shortstack{\textbf{\# bands} \\ {Same per-band SNR}}} & \multicolumn{4}{c}{\shortstack{\textbf{\# bands} \\ {Same total
 photon budget}}} \\
\cmidrule(lr){3-6} \cmidrule(lr){7-10}
\textbf{Metric} & \textbf{Method} & 3 & 4 & 5 & 32 & 3 & 4 & 5 & 32 \\
\midrule

\multirow{7}{*}{PSNR $\uparrow$}
& LU~\cite{Zimmermann2014-nl}    & 23.69 & 26.76 & 27.77 & 35.91 & 23.72 & 27.77 & 27.77 & 25.81 \\
& NNLU~\cite{Slawski2011-nt}  & 26.21 & 27.30 & 28.56 & 36.02 & 26.79 & 28.36 & 28.56 & 25.83 \\
& HyU~\cite{Chiang2023-vm}    & 23.69 & 26.87 & 27.19 & 30.37 & 23.72 & 27.61 & 27.19 & 25.66 \\
& RLU~\cite{Kumar2025-iy}    & \secondval{26.39} & 27.56 & \secondval{28.91} & \bestval{36.68} & \secondval{27.13} & 28.63 & \secondval{28.91} & 26.15 \\
& NMF-RI~\cite{Jimenez-Sanchez2020-mh} & 26.31 & \secondval{27.76} & 28.38 & \secondval{36.49} & 26.77 & \secondval{28.70} & 28.38 & 26.13 \\
& LUMoS~\cite{McRae2019-cn}  & 21.97 & 21.98 & 21.37 & 23.21 & 21.40 & 22.14 & 21.37 & 21.47 \\
& $\lambda$Split (ours)
         & \bestval{27.25} & \bestval{29.95} & \bestval{31.12} & 36.46 & \bestval{27.45} & \bestval{31.54} & \bestval{31.12} & \bestval{31.32} \\
\midrule

\multirow{7}{*}{MS3IM $\uparrow$}
& LU     & 0.760 & 0.904 & 0.925 & 0.985 & 0.773 & 0.924 & 0.925 & 0.864 \\
& NNLU  & 0.863 & 0.916 & 0.938 & 0.986 & 0.874 & 0.934 & 0.938 & 0.865 \\
& HyU    & 0.759 & 0.895 & 0.900 & 0.936 & 0.773 & 0.906 & 0.900 & 0.855 \\
& RLU    & \secondval{0.866} & \secondval{0.921} & \secondval{0.943} & \secondval{0.988} & \bestval{0.882} & \secondval{0.938} & \secondval{0.943} & \secondval{0.872} \\
& NMF-RI & 0.857 & 0.918 & 0.934 & \bestval{0.989} & \secondval{0.866} & 0.935 & 0.934 & \secondval{0.872} \\
& LUMoS  & 0.723 & 0.764 & 0.743 & 0.796 & 0.704 & 0.766 & 0.743 & 0.748 \\
& $\lambda$Split (ours)
         & \bestval{0.878} & \bestval{0.960} & \bestval{0.971} & \secondval{0.988} & \secondval{0.868} & \bestval{0.967} & \bestval{0.971} & \bestval{0.971} \\
\midrule

\multirow{7}{*}{LPIPS $\downarrow$}
& LU     & 0.431 & 0.436 & 0.427 & 0.194 & 0.397 & 0.413 & 0.427 & 0.441 \\
& NNLU  & 0.239 & 0.267 & 0.232 & 0.128 & 0.192 & 0.238 & 0.232 & 0.392 \\
& HyU    & 0.434 & 0.364 & 0.398 & 0.184 & 0.398 & 0.341 & 0.398 & 0.412 \\
& RLU    & \secondval{0.238} & \secondval{0.258} & \bestval{0.220} & \secondval{0.125} & \secondval{0.187} & \secondval{0.229} & \bestval{0.220} & \bestval{0.387} \\
& NMF-RI & \bestval{0.237} & \bestval{0.247} & \secondval{0.227} & \bestval{0.124} & \bestval{0.185} & \bestval{0.217} & \secondval{0.227} & \secondval{0.389} \\
& LUMoS  & 0.324 & 0.327 & 0.310 & 0.287 & 0.293 & 0.314 & 0.310 & 0.381 \\
& $\lambda$Split (ours)
         & 0.283 & 0.267 & 0.313 & 0.155 & 0.273 & 0.238 & 0.313 & 0.218 \\
\bottomrule
\end{tabular*}
\end{table*}

%% file: sections/discussion.tex
\section{Discussion}
\label{sec:discussion}

The results presented in this work indicate that spectral unmixing in fluorescence microscopy benefits substantially from incorporating learned structural priors in addition to explicit spectral modeling. 
Across $58$ controlled experiments varying noise, spectral overlap, and spectral dimensionality, $\lambda$Split consistently remains stable in regimes where purely pixel-wise, least-squares-based baselines, as well as existing data-driven methods, lead to sub-par results. 
This suggests that resolving ambiguities in spectral data is not solely determined by spectral separability, but also by constraining solutions to structurally plausible patterns (by enforcing consistency with a learned structural prior).

A key consequence of our formulation is that such structural priors can be learned in a fully self-supervised manner by embedding the spectral mixing process into the model. 
This removes the dependence on paired unmixed ground truth while maintaining physical consistency and interpretability. 
In practice, this makes data-driven spectral unmixing accessible in realistic microscopy settings where labeled training data are unavailable.

More broadly, our findings support a shift away from purely analytical, pixel-wise inversion schemes toward hybrid approaches that combine explicit forward models with learned, content-aware priors. 
While our current formulation assumes linear mixing with known spectra, future extensions may relax these assumptions and incorporate spectrum estimation or uncertainty-aware downstream analysis. 

%% file: supplements.tex
\newpage
\begin{center}
{\LARGE \textbf{\textit{Supplementary Material}}} \\
\vspace{4mm}
{\huge \textbf{$\lambda$Split: Self-Supervised Content-Aware\\
Spectral Unmixing for Fluorescence Microscopy}}\\
\vspace{6mm}
{\Large Federico~Carrara, Talley~Lambert, Mehdi~Seifi, Florian~Jug}
\vspace{8mm}
\end{center}

\setcounter{page}{1}
\setcounter{section}{0}
\setcounter{figure}{0}
\setcounter{table}{0}
\setcounter{equation}{0}
\renewcommand{\thepage}{S.\arabic{page}} 
\renewcommand*{\thesection}{S.\arabic{section}}
\renewcommand{\thefigure}{S.\arabic{figure}}
\renewcommand{\thetable}{S.\arabic{table}}
\renewcommand{\theequation}{S.\arabic{equation}}

\input{supplementary/intro_suppl}

\input{supplementary/related_works_suppl}

\section{Extended Methods}
\label{sec:suppl_methods}
\input{supplementary/elbo_derivation_suppl}
\input{supplementary/baseline_suppl}
\input{supplementary/metrics_suppl}

\input{supplementary/data_suppl}
\input{supplementary/ablations_suppl}

\input{supplementary/results_suppl}

%% file: supplementary/intro_suppl.tex
\section{Limitations of Pixel-wise Least-Squares Spectral Unmixing}

Classical spectral unmixing methods typically estimate fluorophore concentration maps by solving a linear least-squares (LS) problem independently at each spatial location (\ie, pixel or voxel). 

Let $s \in \mathbb{R}^L$ denote the spectral measurement at some pixel and $u \in \mathbb{R}^F$ the corresponding vector of fluorophore concentrations. 
The mixing model presented in~\cref{eq:problem_definition} then reduces to the pixel-wise linear system
\begin{equation} \label{eq:pixelwise_mixing_model}
s = M u + \varepsilon .
\end{equation}
A common estimator is obtained by solving the LS problem, that is, minimizing the squared reconstruction error
\begin{equation}
\hat{u} = \arg\min_{u} \| s - Mu \|_2^2 .
\end{equation}
When the mixing matrix $M$ has full column rank, and the number of spectral bands satisfies $L \geq F$, the optimization problem above admits the closed-form solution
\begin{equation} \label{eq:ls_solution}
\hat{u} = (M^\top M)^{-1} M^\top s = M^\dagger s ,
\end{equation}
where $M^\dagger$ denotes the Moore--Penrose pseudo--inverse of $M$.
While computationally efficient and widely adopted in fluorescence microscopy, this pixel-wise LS formulation exhibits several limitations in realistic imaging scenarios. 

\paragraph{\textbf{Spectral overlap causes numerical instability of LS solution.}}

The main limitation is numerical stability and arises when FP emission spectra overlap significantly. 
As introduced above, the columns of the mixing matrix $M \in \mathbb{R}^{L\times F}$ correspond to the discretized FP emission spectra. 
When spectral overlap is substantial, these columns become increasingly correlated and may approach linear dependence, which affects the stability of the inversion of the mixing process.

The stability of the LS solution can be quantified through the condition number of the mixing matrix,
\begin{equation}
\kappa(M) = \frac{\sigma_{\max}(M)}{\sigma_{\min}(M)},
\end{equation}
where $\sigma_{\max}(M)$ and $\sigma_{\min}(M)$ denote the largest and smallest singular values of $M$. 
Increasing spectral overlap causes $\sigma_{\min}(M)$ to become tiny, leading to a large condition number, and, in practice, to an ill-conditioned inverse problem.

This can be better understood by focusing on the singular value decomposition (SVD) of the mixing matrix. 
Let $M = U \Sigma V^\top$
denote the SVD of $M$, where
$\Sigma = \mathrm{diag}(\sigma_1,\dots,\sigma_F)$ contains the singular values with
$\sigma_1 \ge \dots \ge \sigma_F > 0$, \ie, $\sigma_{\min}(M) = \sigma_F$. 
The Moore--Penrose pseudo--inverse used in LS inversion can then be written as
\begin{equation}
M^\dagger = V \Sigma^{-1} U^\top,
\end{equation}
where $\Sigma^{-1} = \mathrm{diag}(1/\sigma_1,\dots,1/\sigma_F)$. 
When the smallest singular value $\sigma_F$ becomes very small, \ie, $\sigma_F \approx 0 $, the corresponding entry $1/\sigma_F$ becomes large, making the pseudo-inverse numerically unstable.

This instability directly affects the propagation of measurement noise. 
Substituting the forward model from~\cref{eq:pixelwise_mixing_model} into the LS solution in~\cref{eq:ls_solution} yields
\begin{equation}
\hat{u} = M^\dagger (Mu + \varepsilon) = u + M^\dagger \varepsilon .
\end{equation}
Hence, the estimation error is given by 
\begin{equation}
\hat{u} - u = M^\dagger \varepsilon .
\end{equation}
Therefore, the magnitude of this error is bounded by
\begin{equation}
\|\hat{u} - u\|_2 = \|M^\dagger \varepsilon\|_2 \leq \|M^\dagger\|_2 \, \|\varepsilon\|_2 \leq \sigma_F^{-1} \|\varepsilon\|_2 ,
\end{equation}
since the spectral norm of a matrix is equal to its largest singular value.
Therefore, it is evident that when the smallest singular value of $M$ approaches zero (\ie, when the system is ill-conditioned), measurement noise is strongly amplified in the LS estimate. 
Consequently, even small perturbations in spectral measurements can lead to large errors in the estimated fluorophore concentrations. 
This effect becomes particularly severe in high-noise regimes, such as those reported in the experiments in~\cref{sec:results_more_noise}.

%% file: supplementary/related_works_suppl.tex
\section{Extended Related Works}
\label{sec:suppl_relworks}

A widely used approach for spectral unmixing is linear unmixing (LU), where concentration maps $U$ are estimated by solving a least-squares problem given reference spectra $M$~\cite{Zimmermann2014-nl}. 
While the simplicity and interpretability of this approach are arguably reasons for its success, linear unmixing is sensitive to noise and spectral overlap, requires accurate reference spectra, and operates independently at each spatial location, thereby not leveraging spatial information or structural priors.

In the hyperspectral imaging and remote sensing literature, the concentration maps in $U$ are also commonly referred to as abundances, while the columns of $M$ are known as endmembers~\cite{Bhatt2020-nz}.

Common variants of linear unmixing are non-negative LU (NNLU) and fully constrained LU (FCLU)~\cite{Bioucas-Dias2012-kl, Heinz2001-fx, Slawski2011-nt}, which solve the least-squares problem while also enforcing physically meaningful priors such as non-negativity and sum-to-one constraints. While these approaches remain widely used due to their simple formulation and efficiency, the resulting linear system may become ill-conditioned under strong spectral overlap and low spectral dimensionality, compromising numerical stability and, hence, unmixing accuracy. Additionally, the sum-to-one constraint naturally induces sparse solutions that may not be well suited to microscopy data, where signals from different emitters often show substantial colocalization.

Alternative methods have been introduced to improve robustness in noisy settings. Richardson–Lucy Unmixing (RLU)~\cite{Kumar2025-iy} adapts the classical Richardson–Lucy deconvolution algorithm~\cite{Richardson1972-tc, Lucy1974-ye} by replacing the convolution operator with the spectral mixing matrix, yielding an iterative maximum-likelihood procedure tailored to Poisson photon noise. HyU~\cite{Chiang2023-vm} leverages hyperspectral phasor analysis to aggregate similar spectra into histogram bins before applying linear unmixing on the averaged signals, enhancing unmixing robustness in low-photon conditions. However, both methods are pixel-wise and, therefore, do not exploit spatial correlations or structural priors in the unmixed images.

Blind unmixing approaches aim to recover FP concentration maps without relying on reference spectra. Instead, they jointly estimate the FPs' emission spectra and the corresponding concentrations directly from the observed mixed spectral image. A widely used framework is Non-negative Matrix Factorization (NMF)~\cite{Lee1999-zk}, which iteratively decomposes spectral data into non-negative spectra and concentration maps. NMF is generally sensitive to initialization and may converge to suboptimal local minima. To alleviate the issue, several variants have been proposed. NMF-ML and NMF-RI~\cite{Ortiz-de-Solorzano2013-oc, Jimenez-Sanchez2020-mh}, two of the most notable examples in fluorescence microscopy, adopt multi-layer factorization schemes and informed initialization strategies using theoretical emission spectra to improve convergence. Nevertheless, once the emission spectra are estimated, these approaches typically resort to pixel-wise LU to recover concentration maps. PICASSO and its variant Mosaic-PICASSO~\cite{Seo2022-td, Cang2024-pq} formulate spectral unmixing in fluorescence microscopy as a mutual information minimization problem between pairs of target structures. Their formulation assumes that the number of spectral bands matches the number of structures to be unmixed, which makes it difficult to apply directly to common spectral imaging setups that provide many contiguous bands.

Learning-based methods tailored for spectral unmixing in fluorescence microscopy have begun to emerge, but they remain relatively limited. LUMoS~\cite{McRae2019-cn} performs hard pixel-wise assignments through k-Means spectral clustering, which may lead to spatial inconsistencies and fail to exploit the inherent spatial smoothness of microscopy images. Supervised convolutional models such as AutoUnmix~\cite{Jiang2023-fm} and UNMIX-ME~\cite{Smith2020-hr} learn a direct mapping from spectral images to concentration maps but require paired ground-truth unmixed data, which are rarely available in practical spectral microscopy settings. Moreover, these models do not explicitly incorporate the underlying spectral-mixing physics, and in the case of AutoUnmix, the computational cost grows quadratically with the number of spectral bands, rendering it impractical for spectral images with more than a handful of bands.

Recent works such as MicroSplit~\cite{Ashesh2026-da} explore content-aware image decomposition in fluorescence microscopy using hierarchical variational architectures. Unlike spectral unmixing, MicroSplit addresses a conceptually different inverse problem: given a single superimposed fluorescence image containing multiple biological structures, it learns, in a supervised manner, to decompose it into separate semantic channels. Although it does not model spectral mixing, the shared microscopy domain and use of content-aware priors make several of its methodological ideas transferable to spectral unmixing scenarios.

Finally, a substantial body of work on deep learning–based spectral unmixing has been developed in the remote sensing community. Recent approaches include autoencoder-based architectures that embed abundance constraints in the latent space, convolutional networks exploiting spatial context, and model-inspired networks that incorporate the mixing process into the decoder or unroll iterative optimization schemes~\cite{Palsson2018-iw, Su2019-ov, Palsson2021-lv, Hong2022-fr, Rasti2022-ec, Rasti2022-ky, Gao2022-nj, Ghosh2022-fm}. However, despite the many conceptual similarities, spectral unmixing in fluorescence microscopy differs fundamentally from remote sensing. As a matter of fact, the two domains exhibit an inverse relationship between spatial and spectral dimensionality: microscopy images provide extremely high spatial resolution with fine, crisp structures, but only a limited number of spectral bands, whereas hyperspectral remote sensing data offer hundreds of spectral bands but orders-of-magnitude coarser spatial resolution. Additionally, the two settings differ substantially in their noise characteristics and magnitude, as well as in the nature and sources of spectral variability within the observed scene. These factors greatly hinder the direct transferability of remote sensing unmixing architectures to microscopy settings, motivating domain-specific models that exploit biological spatial priors.

%% file: supplementary/elbo_derivation_suppl.tex
\subsection{Derivation of the LVAE's ELBO}
\label{subsec:elbo_derivation}

We provide the derivation of the evidence lower bound (ELBO) in
Eq.~\eqref{eq:ELBO}, starting from the conditional log-likelihood objective of Eq.~\eqref{eq:likelihood}. 
Throughout, $\mathbf{z}$ denotes the full set of latent variables of the model.

\paragraph{The Objective.}
Recall from Eq.~\eqref{eq:likelihood} that we seek the decoder parameters
$\theta$ that maximize the conditional log-likelihood of the unmixed
concentration maps given the spectral observation, \ie, for a single training
example we are interested in computing $\log p_\theta(U \mid S)$.
This quantity is not directly accessible: evaluating it requires accounting for
the latent variables $\mathbf{z}$ that the model uses to generate $U$, that is,
marginalizing them out,
\begin{equation}
  \log p_\theta(U \mid S) = \log \int p_\theta(U, \mathbf{z} \mid S)\,
  \mathrm{d}\mathbf{z}.
  \label{eq:marginalization}
\end{equation}

\paragraph{Importance Sampling and Jensen's Inequality.}
The integral in Eq.~\eqref{eq:marginalization} is itself intractable. We render
it estimable by importance sampling, using the variational posterior (the
encoder) $q_\phi(\mathbf{z} \mid S)$ as the proposal distribution, and then
apply Jensen's inequality to the concave logarithm:
\begin{align}
  \log p_\theta(U \mid S)
  &= \log \int q_\phi(\mathbf{z} \mid S)\,
     \frac{p_\theta(U, \mathbf{z} \mid S)}{q_\phi(\mathbf{z} \mid S)}\,
     \mathrm{d}\mathbf{z}
   = \log \mathbb{E}_{q_\phi(\mathbf{z} \mid S)}\!\left[
     \frac{p_\theta(U, \mathbf{z} \mid S)}{q_\phi(\mathbf{z} \mid S)}\right]
   \nonumber\\[2pt]
  &\ge \mathbb{E}_{q_\phi(\mathbf{z} \mid S)}\!\left[
     \log \frac{p_\theta(U, \mathbf{z} \mid S)}{q_\phi(\mathbf{z} \mid S)}\right].
  \label{eq:jensen}
\end{align}

\paragraph{Recovering the ELBO.}
To derive the ELBO formula, we factorize the joint distribution inside
Eq.~\eqref{eq:jensen} as
$p_\theta(U, \mathbf{z} \mid S) = p_\theta(U \mid \mathbf{z}, S)\,
p_\theta(\mathbf{z} \mid S)$ and invoke two assumptions, both consistent with
our generative model:
(i)~the latent code captures all the information the decoder requires, so that
$U$ depends on the observation only through $\mathbf{z}$, i.e.
$p_\theta(U \mid \mathbf{z}, S) = p_\theta(U \mid \mathbf{z})$;
and (ii)~the generative prior over the latent variables is independent of the
observation, $p_\theta(\mathbf{z} \mid S) = p_\theta(\mathbf{z})$. The latter
holds by construction in our architecture: the topmost prior is a fixed
multivariate Gaussian and the lower priors are conditioned only on the latents
above them, so that all dependence on $S$ resides in the encoder
$q_\phi(\mathbf{z} \mid S)$.

Substituting $p_\theta(U, \mathbf{z} \mid S) = p_\theta(U \mid \mathbf{z})\,
p_\theta(\mathbf{z})$ into Eq.~\eqref{eq:jensen} and separating the logarithm
yields
\begin{align}
  \log p_\theta(U \mid S)
  &\ge \mathbb{E}_{q_\phi(\mathbf{z} \mid S)}\!\left[
     \log \frac{p_\theta(U \mid \mathbf{z})\, p_\theta(\mathbf{z})}
     {q_\phi(\mathbf{z} \mid S)}\right]
   \nonumber\\[2pt]
  &= \mathbb{E}_{q_\phi(\mathbf{z} \mid S)}\!\big[\log p_\theta(U \mid \mathbf{z})\big]
   - \mathbb{E}_{q_\phi(\mathbf{z} \mid S)}\!\left[
     \log \frac{q_\phi(\mathbf{z} \mid S)}{p_\theta(\mathbf{z})}\right]
   \nonumber\\[2pt]
  &= \mathbb{E}_{q_\phi(\mathbf{z} \mid S)}\!\big[\log p_\theta(U \mid \mathbf{z})\big]
   - \mathrm{KL}\big(q_\phi(\mathbf{z} \mid S)\,\|\,p_\theta(\mathbf{z})\big),
  \label{eq:elbo-final}
\end{align}
which is precisely the ELBO of Eq.~\eqref{eq:ELBO}, consisting of a
reconstruction term and a KL regularization term. The remaining steps that turn
this bound into the trainable objective of Eq.~\eqref{eq:lambdasplit_loss} are described in
the main text.

%% file: supplementary/baseline_suppl.tex
\subsection{Implementation Details of Baselines}
\label{subsec:suppl_baseline_details}

In this section, we provide additional implementation details for the baselines included in our benchmarks. 
All methods use the fixed ground-truth mixing matrix of emission spectra when required by the algorithm. 
Unless stated otherwise, hyperparameters are set to the values recommended in the original publications or reference implementations. 
When these default settings resulted in suboptimal performance, we performed a grid search over the most relevant hyperparameters to identify configurations better suited to our data.
Finally, all baselines are executed using our unified evaluation pipeline to ensure consistent data processing and metric computation.

\paragraph{Linear Unmixing.} 
Our implementation solves the linear unmixing problem via pixel-wise least-squares using its closed-form solution.
The algorithm operates directly on the measured spectral intensities without additional normalization.

\paragraph{Non-negative Linear Unmixing.}
Our implementation solves the linear unmixing problem under a non-negativity constraint on concentrations using the Alternating Direction Method of Multipliers (ADMM) approach. 
Specifically, we use an ADMM penalty parameter $\rho=1$, a convergence tolerance of $10^{-6}$, and a maximum of 200 iterations.
The algorithm operates directly on the measured spectral intensities without additional normalization.

\paragraph{Fully Constrained Linear Unmixing.}
As in the previous baseline, we solve the unmixing problem under non-negativity and sum-to-one constraints using ADMM. 
Specifically, we use an ADMM penalty parameter $\rho=1$, a convergence tolerance of $10^{-6}$, and a maximum of 200 iterations. 
To enforce the constraints, at each iteration, the abundance estimates are projected onto the probability simplex. 
In addition, we apply pixel-wise $\ell_1$ normalization to the input spectral data to ensure consistency with the imposed sum-to-one constraint on the abundances. 
Indeed, since the mixing spectra are $\ell_1$-normalized and the abundances satisfy $\sum_{i=1}^{F} u_i = 1$, the mixing model $s = Mu$ implies $\|s\|_1 = 1$ (see \cref{eq:sum-to-one}). 
We therefore normalize the input spectra to ensure the constraint remains mathematically consistent.

\paragraph{HyU.} 
Due to the lack of an open-source code repository, we provide our own Python implementation of the HyU algorithm, faithfully following the description reported in the original paper. 
Some hyperparameters are selected via grid search. In particular, we use the $1^{st}$ harmonic for phasor representation and construct a 2D histogram with 128$\times$128 equispaced bins over the $[-1, 1] \times [-1, 1]$ domain in the phasor space, which achieved the best performance on the datasets used in our experiments.
Prior to computing the phasor transform, each pixel spectrum is normalized by its $\ell_1$ norm so that phasor coordinates depend only on spectral shape.
Pixel spectra falling in the same phasor histogram bin are averaged to obtain a representative sample for that bin. 
These representative spectra are then unmixed using the linear unmixing implementation described above. 
The resulting abundance estimates are subsequently mapped back to the corresponding pixels and rescaled by the integrated intensity of the original spectra to restore the original signal magnitude.

\paragraph{Richardson--Lucy Unmixing.}
The algorithm's implementation is taken from the official code repository available at \url{https://github.com/jdmanton/rlsu}.
The abundance estimates are initialized uniformly and updated using the multiplicative Richardson–Lucy rule, which inherently enforces non-negativity. 
The algorithm operates directly on the measured spectral intensities without additional normalization. The number of iterations is empirically set to 100.

\paragraph{NMF-RI.} 
We provide our own Python implementation of the NMF-RI algorithm, faithfully reproducing the official MATLAB code publicly available at \url{https://github.com/djimenezsanchez/NMF-RI}. 
In our experiments, we initialize the mixing matrix with the ground-truth spectra used in the simulation. This provides the algorithm with a favorable initialization and ensures a fair comparison with the other non-blind baselines, which also assume known mixing spectra.
The abundance matrix is initialized with random non-negative values and optimized for 500 iterations.

\paragraph{Transformer Unmixing Autoencoder (TAEU).} 
We rely on the official open-source implementation available at \url{https://github.com/preetam22n/DeepTrans-HSU}. 
We adapt the architecture to fluorescence microscopy data as follows: $(i)$~we replace the softmax activation applied to the FP concentration estimate with a leaky-ReLU to remove the sum-to-one constraint, which we found to further degrade unmixing performance on spectral fluorescence microscopy data; $(ii)$~we train the model on 96$\times$96 patches, to approximately match the lateral resolution of hyperspectral satellite images used in the original paper; $(iii)$~we initialize the decoder with the GT spectra and we keep it frozen during training. 
The model is trained in a self-supervised fashion by minimizing the MSE between the input spectra and the spectra reconstructed from the predicted abundances. Optimization is performed using Adam with a learning rate of $10^{-3}$ for 100 epochs.
The TAEU architecture only works on 2D spectral data. 
Additionally, we observed unstable training when the number of spectral bands is below 32. 
This behavior is consistent with the model's original design, which is intended for hyperspectral remote sensing data with hundreds of spectral bands.

\paragraph{LUMoS.} 
We use our own Python implementation of the LUMoS algorithm based on the existing open-source code for the ImageJ LUMoS plug-in written in Java and available at \url{https://github.com/tristan-mcrae-rochester/Multiphoton-Image-Analysis/blob/master/Spectral%20Unmixing/Code/ImageJ-FIJI/LUMoS_Spectral_Unmixing.java}.
Prior to clustering, each pixel spectrum is normalized by its $\ell_1$ norm so that clustering is driven by spectral shape rather than absolute intensity.
We run LUMoS with 10 different K-means initializations, each optimized for up to 200 iterations, and report the run with the lowest reconstruction error.
This baseline is evaluated only for low-band settings, since clustering high-dimensional spectra (\eg, 32-band) is substantially harder due to the curse of dimensionality.

\paragraph{AutoUnmix.}
We use the official code available at~\url{https://github.com/AlphaYuan/AutoUnmix}. 
We train AutoUnmix with input size (256$\times$256) for 100 epochs, utilizing Adam as optimizer with betas (0.5, 0.999) and a learning rate of 0.0002 for consistency with the official implementation.
AutoUnmix processes each pair of input channels with a shared UNet and uses a second UNet during training, for 84M trainable parameters.
Its compute and memory grow quadratically with the number of spectral bands (120 UNet passes per image at 16 bands, 496 at 32). 
For this reason, we train AutoUnmix and report results only for spectral data with 3, 4, and 5 bands.

\paragraph{UNet.}
We rely on the popular open-source UNet implementation available at ~\url{https://github.com/milesial/Pytorch-UNet}.
The UNet architecture employed in our experiments has a depth of four (\ie, 4-layer encoder and decoder). 
The model is trained in a supervised fashion using the MSE loss and the AdamW optimizer, with a weight decay of 0.0001 and a learning rate of 0.0002, for 70 epochs on patches of $64\times64$ (BioSR) or $256\times256$ (CellAtlas) pixels with a batch size of 32.
The learning rate is reduced by a factor of 10 whenever the validation loss does not improve for 3 epochs, and the checkpoint with the lowest validation loss is used for evaluation.

%% file: supplementary/metrics_suppl.tex
\subsection{Evaluation Metrics}
\label{subseq:suppl_eval_metrics}

Here, we report the implementation details of the evaluation metrics used throughout the experiments. 
Unless otherwise specified, metrics are computed using their standard implementations available in common Python libraries. 
All evaluation metrics computed on unmixed images are first evaluated independently for each fluorophore channel and then averaged across channels. 

\paragraph{Range-Invariant PSNR and MS-SSIM.}
Following Weigert et al.~\cite{Weigert2018-jr}, predictions $\hat U$ are rescaled by a single global scalar $\tilde\alpha$ applied to the entire image,
\[
\tilde\alpha = \arg\min_\alpha \| U - \alpha \hat U \|_2^2.
\]
Therefore, PSNR and MS-SSIM are computed between the GT unmixed image $U$ and the rescaled prediction $\tilde\alpha \hat U$. 
This formulation makes the metrics invariant to linear intensity scaling, ensuring robustness to range mismatches introduced by different normalization schemes.

\paragraph{Signal-to-Noise Ratio (SNR).}
SNR is defined as
\begin{equation}
    \mathrm{SNR} =
    \frac{P_{99}(X) - \mu_{\mathrm{bg}}}
    {\sigma_{\mathrm{bg}}},
    \label{eq:SNR_formula}
\end{equation}
where $X$ represents one predicted unmixed channel and $P_{99}(X)$ denotes its 99th percentile intensity, chosen instead of the maximum to mitigate potential outliers such as saturated pixels. 
A set of background patches is selected to estimate the background mean $\mu_{\mathrm{bg}}$—an estimate of the acquisition offset—and the background standard deviation $\sigma_{\mathrm{bg}}$—an estimate of the pixel noise. 
Specifically, background patches are automatically identified as those with the lowest 2\% of intensity standard deviation, providing a more robust estimate of these background statistics.
Finally, for a spectral image, we compute the spectral SNR ($SNR_{sp}$) as the median across spectral bands, which improves robustness to potential SNR outliers in bands with very low signal.

\paragraph{MicroMS-SSIM.}
MicroMS-SSIM is a variant of MS-SSIM designed for benchmarking image regression tasks such as denoising and unmixing, specifically tailored to capture the distinctive characteristics of microscopy images~\cite{Ashesh2024-ev}. 
We use the open-source implementation available at \url{https://github.com/juglab/MicroSSIM}. 
Note that this implementation currently only supports 2D images.

\paragraph{LPIPS.}
Before evaluation, both GT and predicted images are normalized to the unit interval using min--max normalization. 
Since the neural network used to compute LPIPS (\textit{SqueezeNet} in our case) is pre-trained on RGB data and thus expects three-channel inputs, grayscale predictions are replicated across three channels before computing the perceptual distance.

%% file: supplementary/data_suppl.tex
\section{Extended Data Simulation Pipeline}
\label{sec:suppl_data}

As reported in the main manuscript, we generate synthetic fluorescence spectral images under controlled acquisition conditions using a microscopy simulation pipeline built on the open-source Python package \texttt{microsim}~\cite{Lambert2026-em}.
An illustrated overview of our data simulation pipeline is shown in~\cref{fig:simulation_pipeline}.

Multi-channel inputs, such as real multiplexed fluorescence microscopy images or multiclass segmentation maps, are interpreted as spatial fluorophore concentration fields and serve as GT for quantitative evaluation of spectral unmixing.
We denote these GT concentration maps as $U_{\mathrm{GT}} \in \mathbb{R}^{F \times Z \times Y \times X}$, where $F$ represents the number of fluorophores or, equivalently, of channels.

As mentioned above, GT channels are associated with specific FPs characterized by their excitation and emission spectra.
Hence, a per-fluorophore noise-free spectral emission volume $S_{\mathrm{emission}} \in \mathbb{R}^{L \times F \times Z \times Y \times X}$ is generated by combining the GT concentration maps with the corresponding emission spectra, which are previously discretized over the spectral dimension $L$.
At this stage, tunable illumination power and the excitation and emission efficiencies of each fluorophore are modeled to generate realistic emission profiles.

The simulator then models microscope-specific optical effects, including configurable point spread functions (PSFs) to reproduce different imaging modalities, such as widefield or confocal microscopy, as well as optical aberrations introduced by the imaging system.
Notice that when the inputs consist of experimentally acquired microscopy images, these optical effects are not applied, as already present in the data source.
After optical effects are applied to the input volume $S_{emission}$, the obtained per-fluorophore optical array is integrated over the FP dimension ($F$), resulting in the optical spectral volume $S_{\mathrm{opt}} \in \mathbb{R}^{L \times Z \times Y \times X}$.

Finally, \texttt{microsim} simulates the image acquisition process with high fidelity.
First, it models realistic photon and detector noise (\eg, read-out noise, thermal noise, and other sensor-specific sources).
Additionally, the simulator allows the explicit specification of the bandwidth and placement of the $L$ spectral detection bands defined above, enabling emulation of a variety of spectral imaging configurations.
The resulting digitalized spectral volume is denoted as $S_{\mathrm{dig}} \in \mathbb{R}^{L \times Z \times Y \times X}$.

Overall, this framework enables the controlled generation of physically consistent spectral mixtures across diverse experimental conditions.

\figMicrosimPipeline

%% file: supplementary/ablations_suppl.tex
\section{Ablation Study}
\label{sec:ablations}

All main results are obtained using a single fixed configuration applied uniformly across all benchmarks, with key parameters being: $(i)$~the number of hierarchical levels in the LVAE architecture ($4$); $(ii)$~the KL loss weight ($\beta=1$) for the KL loss.
Here, we study the sensitivity of $\lambda$Split to each of these parameters to motivate our choices.

We demonstrate that these parameters shape key properties of the model, most notably its implicit denoising capability.
In particular, we show that: $(i)$~our default configuration offers a balanced compromise that performs reliably throughout, a desirable property for a fair comparison against the baselines; $(ii)$~however, tuning individual parameters to a known imaging regime can yield additional gains over our reported results.

\subsection{Number of Hierarchical Levels in the LVAE architecture}
\label{subsec:num_levels_ablation}

The number of stochastic levels $K$ in the LVAE determines the model's capacity and expressivity. 
In our setting, it specifically determines the ability to capture and faithfully reproduce high-frequency patterns in the input.
Crucially, noise is itself high-frequency content: a more expressive network reproduces crisp detail in clean data, but the same sensitivity causes it to fit noise as the input degrades, eroding the implicit denoising we rely on.

We probe this by ablating $K\in\{2,3,4,5\}$ on the BioSR and CellAtlas datasets across exposure times spanning the low- to high-SNR regimes. 
The results reported in~\cref{fig:levels_ablation} confirm this hypothesis: deeper hierarchies resolve finer detail at high SNR, while at low SNR the same high-frequency sensitivity instead captures noise, making shallower hierarchies more robust.
Our choice, $K=4$, retains most of the high-frequency detail of deeper models while avoiding the low-SNR degradation of $K=5$.

\subsection{KL Loss Weight}
\label{subsec:KL_weight_ablation}

The KL term in the variational objective regularizes the latent space, smoothing the learned posterior toward the prior. 
A larger weight $\beta$ enforces a smoother latent representation and thereby suppresses high-frequency content in the reconstruction. 
For image-restoration tasks, this is generally undesirable, as fine detail is lost. 
However, for the exact same reason, KL regularization enables removing noise, which is itself a high-frequency content.
For this reason, an LVAE architecture with a non-negligible $\beta$ is beneficial in our noisy setting.
Thus, similarly to network depth, $\beta$ trades reconstruction fidelity for denoising strength. 

We probe this behavior by ablating $\beta\in\{0.1,1,2,5\}$ on
the BioSR dataset across exposure times spanning the low- to high-SNR regimes. 
The results reported in~\cref{fig:beta_ablation} confirm our hypothesis: weak regularization preserves detail at high SNR but fits noise as the input degrades, whereas strong regularization is markedly more robust under noise at the cost of high-SNR fidelity. 
We adopt $\beta=1$ as a balanced choice across regimes.

\figAblationLevels
\figAblationBeta

%% file: supplementary/results_suppl.tex
\section{Additional Results}
\label{sec:suppl_results}

\subsection{Spectral Unmixing and Noisy Data}

\figFourFluorophores
\figFiveFluorophores
\input{tables/table_2_SNR_BioSR}
\input{tables/table_3_SNR_HHMI25}
\figNoiseCellAtlas
\figNoiseHHMI

This section provides additional experimental details and results supporting the evaluation presented in~\cref{sec:results_more_noise}. 
We first describe the spectral simulation setup used in these experiments by reporting the excitation and emission configurations of the employed fluorophores (\cref{fig:4FPs_setup,fig:5FPs_setup}). 
We then report the full quantitative results for the BioSR and HHMI25 datasets under varying noise levels (\cref{tab:results_biosr,tab:results_hhmi25}), together with additional qualitative comparisons illustrating the behavior of the evaluated methods in low-SNR regimes (\cref{fig:SNR_exp_CellAtlas,fig:SNR_exp_HHMI25}). 
These figures complement the examples shown in the main paper and further illustrate the robustness of $\lambda$Split under increasing noise conditions.

\subsection{SNR Computation and Implicit Denoising in λSplit}
\label{subsec:implicit_denoising}

To better understand the large SNR$_u$ improvements observed for $\lambda$Split in the experiments using increasingly noisy data, we analyze representative examples from the three evaluated datasets. 
\cref{fig:SNR_linehist} illustrates an example of SNR computation and highlights the differences between $\lambda$Split and classical baselines. 
Across all examples, pixel-wise methods tend to propagate measurement noise from the spectral input into the unmixed channels, leading to larger signal fluctuations. 
In contrast, $\lambda$Split produces considerably smoother background regions while preserving structural foreground details, effectively increasing the separation between background noise and signal peaks. 
This alone accounts for significantly higher SNR$_u$ values reported for $\lambda$Split in the quantitative results.

\figSNRLinehist

\subsection{Spectral Unmixing and Overlapping Spectra}

\input{tables/table_4a_EGFP-EC}
\input{tables/table_4b_EGFP-MC}
\input{tables/table_4d_mTurquoise-EC}
\input{tables/table_4e_mTurquoise-MC}
\figOverlapECthreetwobands
\figOverlapECfivebands
\figOverlapMCfivebands

This section reports additional results for the spectral overlap experiments introduced in~\cref{sec:results_more_overlap}. 
We provide the full quantitative results for all evaluated fluorophore pairs (\cref{tab:overlap_EGFP_EC_merged,tab:overlap_egfp_mc_merged,tab:ss_bins_mturq_ec_merged,tab:ss_bins_mturq_mc_merged}) together with additional qualitative examples for both 5- and 32-band acquisition setups (\cref{fig:overlap_EC_32bands,fig:overlap_EC_5bands,fig:overlap_MC_5bands}). 
These figures and tables complement the discussion in the main paper and further illustrate the behavior of the evaluated methods as spectral overlap increases.

\subsection{Spectral Unmixing and Spectral Dimensionality}

\input{tables/table_5_low_num_bands_extra_metrics}
\figLowBandsQuali

In this section, we present additional results from experiments investigating the effect of reducing the number of spectral bands. 
We report the quantitative results for MicroMS-SSIM, Pearson, and SNR over unmixed images for all evaluated configurations in \cref{tab:results_low_num_bands_extra}. 
These tables complement the discussion in~\cref{sec:results_fewer_bands} and further illustrate the behavior of the evaluated methods as spectral dimensionality decreases, including regimes where the number of bands falls below the number of fluorophores. 
Additional qualitative examples are also included to visually compare the reconstructions produced by the different approaches (see~\cref{fig:exp_num_bands}).

\subsection{λSplit \vs Supervised Spectral Unmixing}

\input{tables/table_6_autounmix_bands}
\input{tables/table_7a_autounmix_overlap_EGFP-EC}
\input{tables/table_7b_autounmix_overlap_EGFP-MC}
\input{tables/table_7c_autounmix_overlap_mTurquoise-EC}
\input{tables/table_7d_autounmix_overlap_mTurquoise-MC}
\figAutounmixUNetLowBands
\figAutounmixOverlap

This section provides additional quantitative and qualitative comparisons between $\lambda$Split and the supervised baselines AutoUnmix and UNet. 
We report result tables for the reduced spectral dimensionality experiment (\cref{tab:autounmix_bands}) as well as for scenarios with increasing spectral overlap between fluorophores (\cref{tab:autounmix_overlap_EGFP-EC,tab:autounmix_overlap_EGFP-MC,tab:autounmix_overlap_mTurquoise-EC,tab:autounmix_overlap_mTurquoise-MC}). 
These tables complement the discussion in~\cref{sec:results_us_vs_supervised} and further illustrate the behavior of the considered approaches under increasingly ill-conditioned unmixing settings. 
Additional qualitative comparisons (see \cref{fig:autounmix_unet_num_bands,fig:autounmix_overlap}) are included to compare unmixing quality and fidelity between $\lambda$Split and the supervised baselines.

%% file: tables/table_2_SNR_BioSR.tex
\begin{table}[t]
\centering
\caption{
\textbf{Quantitative results on the \textit{BioSR} dataset for various noise levels. }
Columns, from left to right, report:
$(i)$~exposure times (less causing more noise);
$(ii)$~the resulting SNR of the spectral input images (aggregated over all spectral bands); 
$(iii)$~baselines considered for the set of experiments;
$(iv)$~quality metrics, \ie, PSNR, MS-SSIM (MS3IM), MicroMS-SSIM ($\mu$MS3IM), LPIPS, Pearson correlation coefficient, and average SNR of unmixed images (see main text for details). 
Higher values are better, except for LPIPS.
In each metric column, the best result is shown in \textbf{bold}, while the second best is \uline{underlined}.
}
\label{tab:results_biosr}
\renewcommand{\arraystretch}{1.0}

\footnotesize
\begin{tabular*}{\linewidth}{lll@{\extracolsep{\fill}}cccccc}
\toprule
 \textbf{Exp} & \textbf{SNR$_{sp}$} & \textbf{Method}
& PSNR $\uparrow$
& MS3IM $\uparrow$
& $\mu$MS3IM $\uparrow$
& LPIPS $\downarrow$
& Pearson $\uparrow$
& SNR$_{u}$ $\uparrow$ \\
\midrule

\multirow{8}{*}{10\textit{ms}}
& \multirow{8}{*}{5.60}
& LU~\cite{Zimmermann2014-nl}    & 30.33 & 0.922 & 0.837 & 0.548 & 0.849 & 9.86 \\
& & NNLU~\cite{Slawski2011-nt}  & 30.42 & 0.923 & 0.842 & 0.503 & 0.852 & 10.69 \\
& & FCLU~\cite{Heinz2001-fx}  & 26.63 & 0.752 & 0.736 & 0.638 & 0.607 & 5.51 \\
& & HyU~\cite{Chiang2023-vm}   & 30.34 & 0.900 & 0.845 & \secondval{0.476} & 0.852 & \secondval{15.17} \\
& & RLU~\cite{Kumar2025-iy}   & 30.54 & \secondval{0.925} & 0.846 & 0.502 & 0.854 & 11.45 \\
& & NMF-RI~\cite{Jimenez-Sanchez2020-mh} & \secondval{30.56} & \secondval{0.925} & 0.842 & 0.497 & 0.854 & 10.59 \\
& & TAEU~\cite{Ghosh2022-fm}  & 28.06 & 0.801 & 0.785 & 0.505 & 0.682 & 35.12 \\
& & $\lambda$Split (ours)  & \bestval{38.55} & \bestval{0.981} & \bestval{0.945} & \bestval{0.138} & \bestval{0.977} & \bestval{56.71} \\
\midrule

\multirow{8}{*}{20\textit{ms}}
& \multirow{8}{*}{9.77}
& LU       & 34.42 & 0.965 & 0.907 & 0.368 & 0.938 & 18.21 \\
& & NNLU   & 34.51 & \secondval{0.966} & 0.910 & 0.325 & 0.939 & 19.77 \\
& & FCLU   & 27.53 & 0.773 & 0.764 & 0.588 & 0.681 & 8.48 \\
& & HyU    & 33.29 & 0.940 & 0.673 & \secondval{0.318} & 0.919 & \secondval{28.83} \\
& & RLU    & \secondval{34.64} & \secondval{0.966} & 0.914 & 0.321 & 0.939 & 21.64 \\
& & NMF-RI & 34.63 & \secondval{0.966} & 0.911 & 0.319 & 0.939 & 19.86 \\
& & TAEU   & 27.44 & 0.784 & 0.766 & 0.448 & 0.644 & 34.59 \\
& & $\lambda$Split (ours) & \bestval{44.83} & \bestval{0.993} & \bestval{0.988} & \bestval{0.080} & \bestval{0.995} & \bestval{509.24} \\
\midrule

\multirow{8}{*}{100\textit{ms}}
& \multirow{8}{*}{33.45}
& LU       & 42.68 & \secondval{0.989} & 0.958 & 0.107 & 0.987 & 67.26 \\
& & NNLU   & 42.76 & \secondval{0.989} & 0.711 & 0.072 & \secondval{0.987} & 71.88 \\
& & FCLU   & 27.91 & 0.789 & 0.790 & 0.459 & 0.731 & 16.77 \\
& & HyU    & 37.78 & 0.971 & 0.461 & 0.124 & 0.962 & \secondval{93.38} \\
& & RLU    & \secondval{43.00} & \secondval{0.989} & 0.960 & \bestval{0.068} & 0.986 & 74.90 \\
& & NMF-RI & 42.98 & \secondval{0.989} & 0.959 & \secondval{0.071} & \secondval{0.987} & 73.25 \\
& & TAEU   & 28.00 & 0.803 & 0.791 & 0.398 & 0.686 & 161.53 \\
& & $\lambda$Split (ours) & \bestval{46.15} & \bestval{0.994} & \bestval{0.990} & 0.087 & \bestval{0.996} & \bestval{660.22} \\
\midrule

\multirow{8}{*}{300\textit{ms}}
& \multirow{8}{*}{57.43}
& LU       & 46.08 & \secondval{0.991} & 0.498 & 0.068 & 0.991 & 124.25 \\
& & NNLU   & 46.10 & \secondval{0.991} & 0.748 & \secondval{0.039} & \secondval{0.991} & 129.21 \\
& & FCLU   & 27.85 & 0.790 & 0.792 & 0.391 & 0.725 & 25.00 \\
& & HyU    & 39.12 & 0.976 & 0.953 & 0.092 & 0.969 & \secondval{162.10} \\
& & RLU    & \secondval{46.66} & 0.990 & 0.749 & \bestval{0.037} & 0.989 & 135.53 \\
& & NMF-RI & \bestval{46.77} & \secondval{0.991} & 0.749 & \secondval{0.039} & \secondval{0.991} & 125.27 \\
& & TAEU~  & 27.79 & 0.791 & 0.779 & 0.409 & 0.676 & 87.34 \\
& & $\lambda$Split (ours) & 46.03 & \bestval{0.993} & \bestval{0.990} & 0.091 & \bestval{0.996} & \bestval{666.50} \\
\bottomrule
\end{tabular*}
\end{table}

%% file: tables/table_3_SNR_HHMI25.tex
\begin{table}[t]
\centering
\caption{
\textbf{Quantitative results on the \textit{HHMI25} dataset for various noise levels. }
Columns, from left to right, report:
$(i)$~exposure times (less causing more noise);
$(ii)$~the resulting SNR of the spectral input images (aggregated over all spectral bands); 
$(iii)$~baselines considered for the set of experiments;
$(iv)$~quality metrics, \ie, PSNR, MS-SSIM (MS3IM), LPIPS, Pearson correlation coefficient, and average SNR of unmixed images (see main text for details). 
MicroMS-SSIM is not reported as not implemented for 3D images.
Higher values are better, except for LPIPS.
In each metric column, the best result is shown in \textbf{bold}, while the second best is \uline{underlined}.
}
\label{tab:results_hhmi25}
\renewcommand{\arraystretch}{1.0}

\footnotesize
\begin{tabular*}{\linewidth}{lll@{\extracolsep{\fill}}cccccc}
\toprule
\textbf{Exp} & \textbf{SNR$_{sp}$} & \textbf{Method}
& PSNR $\uparrow$
& MS3IM $\uparrow$
& LPIPS $\downarrow$
& Pearson $\uparrow$
& SNR$_{u}$ $\uparrow$ \\
\midrule

\multirow{7}{*}{2\textit{ms}}
& \multirow{7}{*}{4.49}
& LU~\cite{Zimmermann2014-nl}    & 27.60 & 0.878 & 0.419 & 0.665 & 4.88 \\
& & NNLU~\cite{Slawski2011-nt}   & 27.62 & 0.878 & \secondval{0.333} & 0.666 & 4.91 \\
& & FCLU~\cite{Heinz2001-fx}    & 26.01 & 0.772 & 0.398 & 0.524 & 4.39 \\
& & HyU~\cite{Chiang2023-vm}    & \secondval{27.94} & 0.873 & 0.402 & \secondval{0.687} & \secondval{9.49} \\
& & RLU~\cite{Kumar2025-iy}     & 27.77 & 0.882 & \secondval{0.332} & 0.679 & 5.57 \\
& & NMF-RI~\cite{Jimenez-Sanchez2020-mh} & 27.83 & \secondval{0.885} & 0.347 & 0.684 & 5.84 \\
& & $\lambda$Split (ours) & \bestval{29.02} & \bestval{0.916} & \bestval{0.273} & \bestval{0.753} & \bestval{9.87} \\
\midrule

\multirow{7}{*}{5\textit{ms}}
& \multirow{7}{*}{8.64}
& LU     & 31.65 & 0.949 & 0.336 & 0.864 & 9.63 \\
& & NNLU   & 31.76 & 0.951 & 0.219 & 0.868 & 9.63 \\
& & FCLU & 27.73 & 0.818 & 0.324 & 0.690 & 7.10 \\
& & HyU  & 30.54 & 0.925 & 0.302 & 0.815 & \secondval{16.77} \\
& & RLU    & \secondval{32.05} & 0.956 & \secondval{0.216} & \secondval{0.875} & 9.58 \\
& & NMF-RI & 31.96 & \secondval{0.957} & 0.223 & 0.874 & 9.72 \\
& & $\lambda$Split (ours) & \bestval{34.69} & \bestval{0.978} & \bestval{0.132} & \bestval{0.934} & \bestval{64.58} \\
\midrule

\multirow{7}{*}{10\textit{ms}}
& \multirow{7}{*}{12.25}
& LU     & 35.07 & 0.972 & 0.253 & 0.935 & 13.20 \\
& & NNLU   & 35.38 & 0.976 & 0.121 & 0.940 & 13.14 \\
& & FCLU   & 28.65 & 0.830 & 0.263 & 0.748 & 7.69 \\
& & HyU    & 31.84 & 0.940 & 0.233 & 0.856 & \secondval{21.90} \\
& & RLU    & \secondval{35.76} & \secondval{0.980} & \secondval{0.120} & \secondval{0.944} & 13.30 \\
& & NMF-RI & 35.57 & \secondval{0.980} & 0.122 & 0.943 & 13.32 \\
& & $\lambda$Split (ours) & \bestval{36.91} & \bestval{0.986} & \bestval{0.090} & \bestval{0.960} & \bestval{75.01} \\
\midrule

\multirow{7}{*}{20\textit{ms}}
& \multirow{7}{*}{16.62}
& LU     & 38.13 & 0.981 & 0.202 & 0.965 & 17.66 \\
& & NNLU   & 38.73 & 0.987 & \secondval{0.065} & 0.970 & 17.61 \\
& & FCLU   & 28.87 & 0.832 & 0.216 & 0.767 & 7.38 \\
& & HyU    & 32.65 & 0.947 & 0.170 & 0.877 & \secondval{23.35} \\
& & RLU    & \bestval{39.22} & \secondval{0.989} & \secondval{0.065} & \bestval{0.973} & 17.69 \\
& & NMF-RI & \secondval{39.00} & \bestval{0.990} & \bestval{0.064} & \secondval{0.971} & 17.74 \\
& & $\lambda$Split (ours) & 38.03 & \secondval{0.989} & 0.079 & 0.969 & \bestval{72.02} \\
\bottomrule
\end{tabular*}
\end{table}

%% file: tables/table_4a_EGFP-EC.tex

\begin{table*}[t]
\centering
\caption{
\textbf{Quantitative results for different spectral overlaps. }
For these experiments, we use the ER and CCP structures from the \textit{BioSR} dataset and associate them with the EGFP fluorophore. 
We report the results for both 5- and 32-band input spectral data.
For each metric, the best result is shown in \textbf{bold}, while the second best is \uline{underlined}.
}
\label{tab:overlap_EGFP_EC_merged}
\vspace{-16mm}
\rotatebox{90}{%
\begin{minipage}{\textheight}
\centering
\footnotesize
\renewcommand{\arraystretch}{0.95}
\setlength{\tabcolsep}{2.2pt}

\resizebox{0.80\textheight}{!}{%
\begin{tabular}{llccccc@{\hskip 6pt}ccccc}
\toprule
\multicolumn{12}{c}{\textbf{EGFP -- ER/CCPs}} \\
\midrule
\textbf{Metric} & \textbf{Method}
& \multicolumn{5}{c}{\textbf{5 bands}} & \multicolumn{5}{c}{\textbf{32 bands}} \\
\cmidrule(lr){3-7} \cmidrule(lr){8-12}
&
& $\Delta\lambda=2$ & $\Delta\lambda=5$ & $\Delta\lambda=10$ & $\Delta\lambda=20$ & $\Delta\lambda=50$
& $\Delta\lambda=2$ & $\Delta\lambda=5$ & $\Delta\lambda=10$ & $\Delta\lambda=20$ & $\Delta\lambda=50$ \\
\midrule

\multirow{6}{*}{PSNR $\uparrow$}
& LU~\cite{Zimmermann2014-nl}     & 31.73 & 38.45 & 43.77 & 49.13 & 53.31 & 31.44 & 38.35 & 43.75 & 48.52 & 52.52 \\
& NNLU~\cite{Slawski2011-nt}  & \secondval{32.76} & \secondval{38.98} & \secondval{44.17} & 49.44 & 53.38 & 31.53 & 37.15 & 43.65 & 48.52 & 52.52 \\
& HyU~\cite{Chiang2023-vm}    & 31.88 & 38.56 & 43.66 & 49.31 & 53.12 & \secondval{32.50} & \secondval{39.63} & \secondval{44.64} & \secondval{48.74} & 52.16 \\
& RLU~\cite{Kumar2025-iy}    & 31.69 & 37.68 & 44.07 & \secondval{49.68} & \secondval{53.95} & 32.02 & 38.05 & 43.79 & 48.55 & \secondval{52.75} \\
& NMF-RI~\cite{Jimenez-Sanchez2020-mh} & 31.38 & 36.01 & 42.58 & 46.99 & 48.78 & 31.70 & 37.45 & 43.39 & 47.97 & 51.30 \\
& $\lambda$Split (ours)
         & \bestval{37.61} & \bestval{42.91} & \bestval{47.55} & \bestval{51.40} & \bestval{54.29}
         & \bestval{36.97} & \bestval{42.68} & \bestval{47.43} & \bestval{49.57} & \bestval{54.78} \\
\midrule

\multirow{6}{*}{MS3IM $\uparrow$}
& LU     & \secondval{0.923} & 0.976 & 0.992 & \bestval{0.998} & \bestval{1.000} & 0.917 & 0.976 & 0.992 & \bestval{0.998} & \bestval{0.999} \\
& NNLU  & 0.916 & \secondval{0.977} & \secondval{0.993} & \bestval{0.998} & \bestval{1.000} & 0.891 & 0.947 & 0.992 & \bestval{0.998} & \bestval{0.999} \\
& HyU    & 0.922 & 0.974 & 0.991 & \bestval{0.998} & \secondval{0.999} & \secondval{0.928} & \secondval{0.978} & \secondval{0.993} & \bestval{0.998} & \bestval{0.999} \\
& RLU    & 0.892 & 0.946 & 0.986 & \bestval{0.998} & \bestval{1.000} & 0.895 & 0.954 & 0.988 & \secondval{0.997} & \bestval{0.999} \\
& NMF-RI & 0.908 & 0.940 & 0.985 & \secondval{0.993} & \secondval{0.993} & 0.916 & 0.961 & 0.990 & \secondval{0.997} & \secondval{0.998} \\
& $\lambda$Split (ours)
         & \bestval{0.949} & \bestval{0.981} & \bestval{0.994} & \bestval{0.998} & \secondval{0.999}
         & \bestval{0.947} & \bestval{0.983} & \bestval{0.994} & \secondval{0.997} & \bestval{0.999} \\
\midrule

\multirow{6}{*}{LPIPS $\downarrow$}
& LU     & 0.436 & 0.223 & 0.097 & 0.041 & \secondval{0.013} & 0.484 & 0.230 & 0.097 & 0.047 & \bestval{0.017} \\
& NNLU  & 0.314 & 0.183 & 0.088 & 0.037 & \secondval{0.013} & \secondval{0.219} & 0.188 & 0.096 & 0.046 & \bestval{0.017} \\
& HyU    & 0.370 & 0.186 & 0.093 & 0.037 & \secondval{0.013} & 0.370 & \secondval{0.178} & \secondval{0.075} & \secondval{0.043} & \secondval{0.018} \\
& RLU    & \secondval{0.220} & \secondval{0.160} & \secondval{0.075} & \secondval{0.032} & \bestval{0.011} & 0.242 & \secondval{0.168} & 0.085 & \secondval{0.043} & \bestval{0.017} \\
& NMF-RI & 0.323 & 0.197 & 0.099 & 0.046 & 0.021 & 0.364 & 0.198 & 0.096 & 0.048 & \bestval{0.017} \\
& $\lambda$Split (ours)
         & \bestval{0.196} & \bestval{0.094} & \bestval{0.041} & \bestval{0.022} & 0.027
         & \bestval{0.195} & \bestval{0.084} & \bestval{0.040} & \bestval{0.024} & 0.024 \\
\midrule

\multirow{6}{*}{$\mu$MS3IM $\uparrow$}
& LU     & 0.448 & 0.493 & \secondval{0.499} & \secondval{0.502} & 0.988 & 0.439 & 0.492 & \secondval{0.499} & 0.501 & 0.983 \\
& NNLU  & 0.001 & 0.495 & \secondval{0.499} & \secondval{0.502} & 0.988 & \secondval{0.895} & 0.495 & \secondval{0.499} & 0.501 & 0.983 \\
& HyU    & 0.003 & 0.494 & \secondval{0.499} & \secondval{0.502} & 0.989 & 0.003 & 0.495 & 0.005 & \secondval{0.502} & 0.983 \\
& RLU    & \secondval{0.896} & \secondval{0.496} & \secondval{0.500} & \secondval{0.503} & \secondval{0.991} & 0.477 & \secondval{0.496} & 0.005 & \secondval{0.502} & \secondval{0.984} \\
& NMF-RI & 0.441 & 0.490 & \secondval{0.499} & \secondval{0.502} & 0.936 & 0.454 & 0.004 & \secondval{0.499} & 0.501 & 0.968 \\
& $\lambda$Split (ours)
         & \bestval{0.914} & \bestval{0.926} & \bestval{0.969} & \bestval{0.993} & \bestval{0.999}
         & \bestval{0.911} & \bestval{0.927} & \bestval{0.969} & \bestval{0.991} & \bestval{0.999} \\
\midrule

\multirow{6}{*}{Pearson $\uparrow$}
& LU     & \secondval{0.858} & 0.963 & 0.989 & \secondval{0.997} & \bestval{0.999} & 0.852 & 0.963 & 0.989 & \secondval{0.997} & \bestval{0.999} \\
& NNLU  & 0.820 & \secondval{0.965} & \secondval{0.990} & \secondval{0.997} & \bestval{0.999} & 0.725 & 0.912 & 0.988 & \secondval{0.997} & \bestval{0.999} \\
& HyU    & 0.845 & 0.961 & 0.988 & \secondval{0.997} & \bestval{0.999} & \secondval{0.873} & \secondval{0.970} & \secondval{0.991} & \secondval{0.997} & \bestval{0.999} \\
& RLU    & 0.731 & 0.910 & 0.984 & \secondval{0.997} & \bestval{0.999} & 0.746 & 0.928 & 0.986 & 0.996 & \bestval{0.999} \\
& NMF-RI & 0.818 & 0.886 & 0.981 & \secondval{0.993} & \secondval{0.993} & 0.830 & 0.942 & 0.987 & 0.996 & \secondval{0.998} \\
& $\lambda$Split (ours)
         & \bestval{0.933} & \bestval{0.982} & \bestval{0.995} & \bestval{0.998} & \bestval{0.999}
         & \bestval{0.933} & \bestval{0.983} & \bestval{0.994} & \bestval{0.998} & \bestval{0.999} \\
\midrule

\multirow{6}{*}{SNR$_u$ $\uparrow$}
& LU     & 17.13 & 40.01 & 69.29 & 106.67 & 147.93 & 14.53 & 35.36 & 63.55 & 95.07 & 126.88 \\
& NNLU  & 28.03 & 47.92 & 75.24 & 110.00 & 148.17 & \secondval{61.43} & 43.69 & 63.89 & 95.08 & 126.88 \\
& HyU    & 20.02 & 44.80 & 73.23 & 113.25 & 150.13 & 20.46 & \secondval{48.75} & \secondval{78.98} & \secondval{104.54} & \secondval{130.17} \\
& RLU    & \secondval{65.21} & \secondval{56.78} & \secondval{83.00} & \secondval{114.46} & \secondval{155.26} & 45.77 & 43.08 & 68.53 & 98.69 & 128.24 \\
& NMF-RI & 22.67 & 46.90 & 70.89 & 108.97 & 151.94 & 17.75 & 37.46 & 64.00 & 95.90 & 127.94 \\
& $\lambda$Split (ours)
         & \bestval{727.70} & \bestval{1048.60} & \bestval{1287.36} & \bestval{1356.12} & \bestval{1711.83}
         & \bestval{611.02} & \bestval{1119.60} & \bestval{1074.09} & \bestval{775.49} & \bestval{2599.14} \\
\bottomrule
\end{tabular}%
}
\end{minipage}%
}
\end{table*}

%% file: tables/table_4b_EGFP-MC.tex
\begin{table*}[t]
\centering
\caption{
\textbf{Quantitative results for different spectral overlaps. }
For these experiments, we use the Microtubules and CCPs structures from the \textit{BioSR} dataset and associate them with the EGFP fluorophore. 
We report the results for both 5- and 32-band input spectral data.
For each metric, the best result is shown in \textbf{bold}, while the second best is \uline{underlined}.
}
\label{tab:overlap_egfp_mc_merged}
\vspace{-16mm}
\rotatebox{90}{%
\begin{minipage}{\textheight}
\centering
\footnotesize
\renewcommand{\arraystretch}{0.95}
\setlength{\tabcolsep}{2.2pt}

\resizebox{0.80\textheight}{!}{%
\begin{tabular}{llccccc@{\hskip 6pt}ccccc}
\toprule
\multicolumn{12}{c}{\textbf{EGFP -- Microtubules/CCPs}} \\
\midrule
\textbf{Metric} & \textbf{Method}
& \multicolumn{5}{c}{\textbf{5 bands}} & \multicolumn{5}{c}{\textbf{32 bands}} \\
\cmidrule(lr){3-7} \cmidrule(lr){8-12}
&
& $\Delta\lambda=2$ & $\Delta\lambda=5$ & $\Delta\lambda=10$ & $\Delta\lambda=20$ & $\Delta\lambda=50$
& $\Delta\lambda=2$ & $\Delta\lambda=5$ & $\Delta\lambda=10$ & $\Delta\lambda=20$ & $\Delta\lambda=50$ \\
\midrule

\multirow{6}{*}{PSNR $\uparrow$}
& LU~\cite{Zimmermann2014-nl}    & 32.76 & 39.59 & 44.95 & 50.36 & 54.53 & 32.34 & 39.33 & 44.77 & 49.57 & \secondval{53.44} \\
& NNLU~\cite{Slawski2011-nt}  & \secondval{33.78} & \secondval{40.15} & \secondval{45.33} & 50.65 & \secondval{54.60} & 32.16 & 38.06 & 44.67 & 49.56 & \secondval{53.44} \\
& HyU~\cite{Chiang2023-vm}    & 32.98 & 39.76 & 44.86 & 50.54 & 54.12 & \secondval{33.34} & \secondval{40.49} & \secondval{45.53} & \secondval{49.64} & 52.75 \\
& RLU~\cite{Kumar2025-iy}    & 32.33 & 38.59 & 45.14 & \secondval{50.83} & \bestval{55.13} & 32.66 & 38.86 & 44.60 & 49.16 & \bestval{53.60} \\
& NMF-RI~\cite{Jimenez-Sanchez2020-mh} & 33.06 & 38.76 & 44.12 & 48.65 & 50.70 & 32.68 & 38.82 & 44.26 & 49.02 & 53.01 \\
& $\lambda$Split (ours)
         & \bestval{37.91} & \bestval{42.81} & \bestval{47.31} & \bestval{51.27} & 52.65
         & \bestval{37.07} & \bestval{43.19} & \bestval{47.37} & \bestval{51.00} & 52.81 \\
\midrule

\multirow{6}{*}{MS3IM $\uparrow$}
& LU     & \secondval{0.939} & \secondval{0.981} & \bestval{0.994} & \bestval{0.999} & \bestval{1.000} & 0.932 & 0.980 & \secondval{0.994} & \bestval{0.998} & \bestval{1.000} \\
& NNLU  & 0.928 & \secondval{0.981} & \bestval{0.994} & \bestval{0.999} & \bestval{1.000} & 0.900 & 0.953 & 0.993 & \bestval{0.998} & \bestval{1.000} \\
& HyU    & 0.938 & 0.979 & \secondval{0.993} & \bestval{0.999} & \bestval{1.000} & \secondval{0.939} & \secondval{0.981} & \secondval{0.994} & \bestval{0.998} & \secondval{0.999} \\
& RLU    & 0.902 & 0.951 & 0.988 & \secondval{0.998} & \bestval{1.000} & 0.905 & 0.958 & 0.989 & \secondval{0.997} & \bestval{1.000} \\
& NMF-RI & 0.934 & 0.967 & 0.989 & 0.996 & 0.996 & 0.936 & 0.970 & 0.990 & \secondval{0.997} & \secondval{0.999} \\
& $\lambda$Split (ours)
         & \bestval{0.954} & \bestval{0.983} & \bestval{0.994} & \secondval{0.998} & \secondval{0.999}
         & \bestval{0.952} & \bestval{0.984} & \bestval{0.995} & \bestval{0.998} & \secondval{0.999} \\
\midrule

\multirow{6}{*}{LPIPS $\downarrow$}
& LU     & 0.400 & 0.171 & 0.083 & 0.030 & \secondval{0.008} & 0.431 & 0.191 & 0.082 & 0.034 & \bestval{0.012} \\
& NNLU  & 0.242 & \secondval{0.119} & \secondval{0.055} & \secondval{0.023} & \secondval{0.008} & \bestval{0.189} & 0.148 & 0.069 & 0.032 & \bestval{0.012} \\
& HyU    & 0.370 & 0.159 & 0.072 & 0.028 & \secondval{0.008} & 0.344 & 0.147 & \secondval{0.066} & \secondval{0.030} & \secondval{0.013} \\
& RLU    & \bestval{0.181} & 0.126 & 0.056 & \bestval{0.021} & \bestval{0.007} & \secondval{0.205} & \secondval{0.137} & 0.067 & 0.031 & \bestval{0.012} \\
& NMF-RI & 0.257 & 0.134 & 0.065 & 0.031 & 0.014 & 0.318 & 0.150 & 0.071 & 0.034 & \bestval{0.012} \\
& $\lambda$Split (ours)
         & \secondval{0.209} & \bestval{0.109} & \bestval{0.049} & 0.024 & 0.030
         & 0.210 & \bestval{0.104} & \bestval{0.048} & \bestval{0.026} & 0.033 \\
\midrule

\multirow{6}{*}{$\mu$MS3IM $\uparrow$}
& LU     & 0.003 & 0.497 & 0.501 & 0.505 & 0.991 & 0.449 & 0.494 & 0.500 & 0.502 & \secondval{0.986} \\
& NNLU  & \secondval{0.489} & \secondval{0.499} & 0.501 & \secondval{0.980} & 0.991 & 0.001 & 0.496 & 0.500 & 0.502 & \secondval{0.986} \\
& HyU    & 0.474 & 0.497 & 0.501 & 0.977 & \secondval{0.992} & 0.469 & 0.497 & \secondval{0.501} & \secondval{0.503} & \secondval{0.986} \\
& RLU    & 0.426 & \secondval{0.499} & \secondval{0.502} & 0.979 & 0.496 & \secondval{0.480} & \secondval{0.497} & 0.500 & 0.502 & \secondval{0.986} \\
& NMF-RI & 0.477 & 0.497 & 0.501 & 0.958 & 0.964 & 0.460 & 0.495 & 0.500 & 0.006 & 0.983 \\
& $\lambda$Split (ours)
         & \bestval{0.920} & \bestval{0.932} & \bestval{0.971} & \bestval{0.996} & \bestval{0.998}
         & \bestval{0.916} & \bestval{0.933} & \bestval{0.972} & \bestval{0.994} & \bestval{0.999} \\
\midrule

\multirow{6}{*}{Pearson $\uparrow$}
& LU     & \secondval{0.872} & 0.968 & 0.990 & \bestval{0.998} & \bestval{0.999} & 0.862 & 0.967 & 0.990 & \secondval{0.997} & \bestval{0.999} \\
& NNLU  & 0.832 & \secondval{0.969} & \secondval{0.991} & \bestval{0.998} & \bestval{0.999} & 0.733 & 0.919 & 0.989 & \secondval{0.997} & \bestval{0.999} \\
& HyU    & 0.864 & 0.967 & 0.990 & \bestval{0.998} & \bestval{0.999} & \secondval{0.880} & \secondval{0.973} & \secondval{0.992} & \secondval{0.997} & \bestval{0.999} \\
& RLU    & 0.740 & 0.916 & 0.986 & \secondval{0.997} & \bestval{0.999} & 0.755 & 0.932 & 0.986 & 0.996 & \bestval{0.999} \\
& NMF-RI & 0.855 & 0.948 & 0.986 & 0.995 & \secondval{0.996} & 0.862 & 0.952 & 0.987 & 0.996 & \bestval{0.999} \\
& $\lambda$Split (ours)
         & \bestval{0.935} & \bestval{0.982} & \bestval{0.995} & \bestval{0.998} & \bestval{0.999}
         & \bestval{0.935} & \bestval{0.983} & \bestval{0.995} & \bestval{0.998} & \bestval{0.999} \\
\midrule

\multirow{6}{*}{SNR$_u$ $\uparrow$}
& LU     & 38.68 & 89.39 & 166.18 & 305.16 & 377.91 & 21.08 & 48.13 & 90.37 & 145.91 & 190.71 \\
& NNLU  & 114.91 & 133.05 & 175.20 & 305.04 & 377.89 & \secondval{86.79} & 61.46 & 90.91 & 145.92 & 190.71 \\
& HyU    & 46.41 & 103.64 & 180.91 & \secondval{327.08} & 378.66 & 28.97 & \secondval{67.99} & \secondval{116.35} & \secondval{163.07} & \secondval{192.73} \\
& RLU    & \secondval{177.47} & \secondval{140.60} & \secondval{211.90} & 324.53 & 374.01 & 56.95 & 54.30 & 94.91 & 146.09 & 179.04 \\
& NMF-RI & 81.04 & 111.52 & 174.90 & 308.57 & \secondval{379.94} & 24.68 & 51.03 & 93.59 & 147.59 & 190.77 \\
& $\lambda$Split (ours)
         & \bestval{806.40} & \bestval{1017.92} & \bestval{1527.50} & \bestval{1170.58} & \bestval{992.97}
         & \bestval{638.38} & \bestval{1027.20} & \bestval{1105.19} & \bestval{1663.74} & \bestval{1337.86} \\
\bottomrule
\end{tabular}%
}
\end{minipage}%
}
\end{table*}

%% file: tables/table_4d_mTurquoise-EC.tex
\begin{table}[t]
\centering
\caption{
\textbf{Quantitative results for different spectral overlaps. }
For these experiments, we use the ER and CCP structures from the \textit{BioSR} dataset and associate them with the mTurquoise fluorophore.
We report the results for both 5- and 32-band input spectral data.
For each metric, the best result is shown in \textbf{bold}, while the second best is \uline{underlined}.
}
\label{tab:ss_bins_mturq_ec_merged}
\vspace{-16mm}
\rotatebox{90}{%
\begin{minipage}{\textheight}
\centering
\footnotesize
\renewcommand{\arraystretch}{0.95}
\setlength{\tabcolsep}{2.2pt}

\resizebox{0.80\textheight}{!}{%
\begin{tabular}{llccccc@{\hskip 6pt}ccccc}
\toprule
\multicolumn{12}{c}{\textbf{mTurquoise -- ER/CCPs}} \\
\midrule
\textbf{Metric} & \textbf{Method}
& \multicolumn{5}{c}{\textbf{5 bands}} & \multicolumn{5}{c}{\textbf{32 bands}} \\
\cmidrule(lr){3-7} \cmidrule(lr){8-12}
&
& $\Delta\lambda=2$ & $\Delta\lambda=5$ & $\Delta\lambda=10$ & $\Delta\lambda=20$ & $\Delta\lambda=50$
& $\Delta\lambda=2$ & $\Delta\lambda=5$ & $\Delta\lambda=10$ & $\Delta\lambda=20$ & $\Delta\lambda=50$ \\
\midrule

\multirow{6}{*}{PSNR $\uparrow$}
& LU~\cite{Zimmermann2014-nl}     & 29.96 & 36.21 & 41.30 & 45.18 & 50.55 & 29.57 & 35.78 & 40.83 & 45.20 & 49.68 \\
& NNLU~\cite{Slawski2011-nt}  & \secondval{31.50} & \secondval{37.05} & \secondval{42.19} & 45.63 & 50.69 & 30.91 & 34.63 & 40.53 & 45.20 & 49.68 \\
& HyU~\cite{Chiang2023-vm}    & 29.85 & 36.41 & 41.53 & 45.01 & 50.34 & 30.54 & \secondval{37.45} & \secondval{42.32} & \secondval{45.87} & 49.52 \\
& RLU~\cite{Kumar2025-iy}    & 31.07 & 35.65 & 41.82 & \secondval{46.00} & \secondval{51.29} & \secondval{31.32} & 35.91 & 40.88 & 45.14 & \secondval{49.77} \\
& NMF-RI~\cite{Jimenez-Sanchez2020-mh} & 31.13 & 35.47 & 41.03 & 44.28 & 47.45 & 29.86 & 35.21 & 40.40 & 44.81 & 49.22 \\
& $\lambda$Split (ours)
         & \bestval{36.88} & \bestval{42.24} & \bestval{47.32} & \bestval{49.75} & \bestval{52.60}
         & \bestval{36.48} & \bestval{43.74} & \bestval{48.07} & \bestval{50.93} & \bestval{53.29} \\
\midrule

\multirow{6}{*}{MS3IM $\uparrow$}
& LU     & 0.904 & 0.968 & 0.988 & \secondval{0.995} & 0.999 & 0.891 & 0.967 & 0.989 & 0.996 & \secondval{0.999} \\
& NNLU  & 0.904 & \secondval{0.971} & \secondval{0.990} & \secondval{0.995} & 0.999 & 0.885 & 0.925 & 0.984 & 0.996 & \secondval{0.999} \\
& HyU    & 0.898 & 0.963 & 0.986 & 0.994 & 0.999 & \secondval{0.908} & \secondval{0.972} & \secondval{0.991} & \secondval{0.997} & \secondval{0.999} \\
& RLU    & 0.887 & 0.933 & 0.979 & 0.994 & 0.999 & 0.889 & 0.939 & 0.979 & 0.994 & \secondval{0.999} \\
& NMF-RI & \secondval{0.922} & 0.948 & 0.984 & 0.990 & \secondval{0.993} & 0.895 & 0.951 & 0.985 & 0.995 & \secondval{0.998} \\
& $\lambda$Split (ours)
         & \bestval{0.950} & \bestval{0.980} & \bestval{0.992} & \bestval{0.996} & \bestval{0.999}
         & \bestval{0.945} & \bestval{0.983} & \bestval{0.995} & \bestval{0.998} & \bestval{0.999} \\
\midrule

\multirow{6}{*}{$\mu$MS3IM $\uparrow$}
& LU     & 0.418 & 0.004 & 0.498 & 0.932 & 0.972 & 0.006 & 0.482 & 0.497 & 0.931 & 0.962 \\
& NNLU  & 0.465 & 0.492 & 0.440 & \secondval{0.944} & 0.972 & \secondval{0.890} & 0.427 & 0.497 & 0.931 & 0.962 \\
& HyU    & 0.423 & 0.489 & 0.433 & 0.934 & 0.975 & 0.442 & \secondval{0.492} & \secondval{0.926} & \secondval{0.937} & \secondval{0.963} \\
& RLU    & \secondval{0.892} & \secondval{0.913} & \secondval{0.924} & \secondval{0.937} & \secondval{0.980} & \secondval{0.891} & 0.007 & 0.921 & 0.931 & \secondval{0.964} \\
& NMF-RI & 0.010 & 0.004 & 0.923 & 0.927 & 0.933 & 0.006 & 0.485 & 0.497 & 0.930 & 0.956 \\
& $\lambda$Split (ours)
         & \bestval{0.911} & \bestval{0.925} & \bestval{0.953} & \bestval{0.983} & \bestval{0.997}
         & \bestval{0.909} & \bestval{0.927} & \bestval{0.976} & \bestval{0.994} & \bestval{0.997} \\
\midrule

\multirow{6}{*}{Pearson $\uparrow$}
& LU     & 0.794 & 0.943 & 0.981 & 0.992 & \secondval{0.998} & 0.777 & 0.939 & 0.980 & 0.993 & \secondval{0.998} \\
& NNLU  & 0.784 & \secondval{0.950} & \secondval{0.984} & \secondval{0.993} & \secondval{0.998} & 0.701 & 0.855 & 0.975 & 0.993 & \secondval{0.998} \\
& HyU    & 0.774 & 0.939 & 0.979 & 0.991 & \secondval{0.998} & \secondval{0.801} & \secondval{0.955} & \secondval{0.985} & \secondval{0.994} & \secondval{0.998} \\
& RLU    & 0.708 & 0.875 & 0.975 & \secondval{0.993} & \secondval{0.998} & 0.725 & 0.894 & 0.973 & 0.992 & \secondval{0.998} \\
& NMF-RI & \secondval{0.839} & 0.908 & 0.978 & 0.988 & \secondval{0.992} & 0.784 & 0.916 & 0.976 & 0.991 & 0.997 \\
& $\lambda$Split (ours)
         & \bestval{0.935} & \bestval{0.979} & \bestval{0.993} & \bestval{0.997} & \bestval{0.999}
         & \bestval{0.930} & \bestval{0.984} & \bestval{0.995} & \bestval{0.998} & \bestval{0.999} \\
\midrule

\multirow{6}{*}{LPIPS $\downarrow$}
& LU     & 0.534 & 0.298 & 0.147 & 0.082 & 0.028 & 0.547 & 0.315 & 0.171 & 0.089 & 0.039 \\
& NNLU  & 0.374 & 0.238 & 0.120 & 0.071 & \secondval{0.026} & \secondval{0.230} & 0.248 & 0.163 & 0.086 & 0.039 \\
& HyU    & 0.475 & 0.248 & 0.123 & 0.075 & 0.029 & 0.449 & \secondval{0.216} & \secondval{0.123} & \secondval{0.076} & 0.039 \\
& RLU    & \secondval{0.235} & \secondval{0.214} & \secondval{0.109} & \secondval{0.062} & \bestval{0.022} & 0.279 & 0.224 & 0.141 & 0.077 & \secondval{0.036} \\
& NMF-RI & 0.383 & 0.241 & 0.137 & 0.080 & 0.036 & 0.459 & 0.285 & 0.164 & 0.086 & 0.038 \\
& $\lambda$Split (ours)
         & \bestval{0.173} & \bestval{0.122} & \bestval{0.051} & \bestval{0.030} & \secondval{0.026}
         & \bestval{0.201} & \bestval{0.093} & \bestval{0.036} & \bestval{0.023} & \bestval{0.024} \\
\midrule

\multirow{6}{*}{SNR$_u$ $\uparrow$}
& LU     & 12.12 & 28.90 & 51.52 & 75.28 & 114.32 & 9.58 & 22.66 & 41.30 & 65.31 & 96.84 \\
& NNLU  & 22.80 & 37.17 & 62.54 & 80.67 & 114.71 & \secondval{58.85} & 34.72 & 42.68 & 65.32 & 96.84 \\
& HyU    & 13.94 & 34.03 & 60.23 & 79.59 & 114.71 & 14.56 & \secondval{35.00} & \secondval{57.16} & \secondval{75.30} & \secondval{99.31} \\
& RLU    & \secondval{62.90} & \secondval{44.40} & \secondval{65.03} & \secondval{84.18} & \secondval{120.36} & 37.21 & 33.43 & 47.26 & 71.12 & \secondval{99.55} \\
& NMF-RI & 16.91 & 32.87 & 54.50 & 77.56 & 117.50 & 11.30 & 24.53 & 42.00 & 66.19 & 97.37 \\
& $\lambda$Split (ours)
         & \bestval{428.91} & \bestval{735.90} & \bestval{1331.06} & \bestval{1641.54} & \bestval{1028.25}
         & \bestval{684.31} & \bestval{1015.44} & \bestval{1017.81} & \bestval{1278.92} & \bestval{1873.25} \\
\bottomrule
\end{tabular}%
}
\end{minipage}%
}
\end{table}

%% file: tables/table_4e_mTurquoise-MC.tex
\begin{table*}[t]
\centering
\caption{
\textbf{Quantitative results for different spectral overlaps. }
For these experiments, we use the Microtubules and CCPs structures from the \textit{BioSR} dataset and associate them with the mTurquoise fluorophore. 
We report the results for both 5- and 32-band input spectral data.
For each metric, the best result is shown in \textbf{bold}, while the second best is \uline{underlined}.
}
\label{tab:ss_bins_mturq_mc_merged}
\vspace{-16mm}
\rotatebox{90}{%
\begin{minipage}{\textheight}
\centering
\footnotesize
\renewcommand{\arraystretch}{0.95}
\setlength{\tabcolsep}{2.2pt}

\resizebox{0.80\textheight}{!}{%
\begin{tabular}{llccccc@{\hskip 6pt}ccccc}
\toprule
\multicolumn{12}{c}{\textbf{mTurquoise -- Microtubules/CCPs}} \\
\midrule
\textbf{Metric} & \textbf{Method}
& \multicolumn{5}{c}{\textbf{5 bands}} & \multicolumn{5}{c}{\textbf{32 bands}} \\
\cmidrule(lr){3-7} \cmidrule(lr){8-12}
&
& $\Delta\lambda=2$ & $\Delta\lambda=5$ & $\Delta\lambda=10$ & $\Delta\lambda=20$ & $\Delta\lambda=50$
& $\Delta\lambda=2$ & $\Delta\lambda=5$ & $\Delta\lambda=10$ & $\Delta\lambda=20$ & $\Delta\lambda=50$ \\
\midrule

\multirow{6}{*}{PSNR $\uparrow$}
& LU~\cite{Zimmermann2014-nl}     & 30.95 & 37.33 & 42.48 & 46.38 & 51.76 & 30.36 & 36.67 & 41.76 & 46.17 & \secondval{50.57} \\
& NNLU\cite{Slawski2011-nt}  & \secondval{32.46} & \secondval{38.22} & \secondval{43.39} & 46.80 & 51.89 & 31.52 & 35.41 & 41.47 & 46.17 & \secondval{50.57} \\
& HyU~\cite{Chiang2023-vm}    & 30.97 & 37.65 & 42.81 & 46.25 & 51.55 & 31.35 & \secondval{38.18} & \secondval{43.08} & \secondval{46.66} & 50.30 \\
& RLU~\cite{Kumar2025-iy}    & 31.70 & 36.52 & 42.88 & \secondval{47.08} & \secondval{52.42} & \secondval{31.96} & 36.67 & 41.63 & 45.60 & 50.41 \\
& NMF-RI~\cite{Jimenez-Sanchez2020-mh} & 31.67 & 37.11 & 41.68 & 45.66 & 49.21 & 30.79 & 36.64 & 41.14 & 45.87 & 50.31 \\
& $\lambda$Split (ours)
         & \bestval{37.64} & \bestval{43.17} & \bestval{46.52} & \bestval{49.26} & \bestval{53.27}
         & \bestval{36.79} & \bestval{42.73} & \bestval{47.46} & \bestval{50.71} & \bestval{53.28} \\
\midrule

\multirow{6}{*}{MS3IM $\uparrow$}
& LU     & 0.925 & 0.975 & \secondval{0.991} & \bestval{0.996} & \bestval{0.999} & 0.908 & 0.974 & 0.991 & \secondval{0.997} & \bestval{0.999} \\
& NNLU  & 0.918 & \secondval{0.976} & \bestval{0.992} & \bestval{0.996} & \bestval{0.999} & 0.894 & 0.933 & 0.987 & \secondval{0.997} & \bestval{0.999} \\
& HyU    & 0.921 & 0.972 & 0.989 & \secondval{0.995} & \bestval{0.999} & \secondval{0.922} & \secondval{0.977} & \secondval{0.992} & \secondval{0.997} & \bestval{0.999} \\
& RLU    & 0.896 & 0.940 & 0.982 & \secondval{0.995} & \bestval{0.999} & 0.899 & 0.946 & 0.981 & 0.993 & \bestval{0.999} \\
& NMF-RI & \secondval{0.930} & 0.963 & 0.981 & 0.993 & \secondval{0.995} & 0.914 & 0.971 & 0.984 & 0.996 & \bestval{0.999} \\
& $\lambda$Split (ours)
         & \bestval{0.956} & \bestval{0.982} & \bestval{0.992} & \bestval{0.996} & \bestval{0.999}
         & \bestval{0.950} & \bestval{0.983} & \bestval{0.995} & \bestval{0.998} & \bestval{0.999} \\
\midrule

\multirow{6}{*}{$\mu$MS3IM $\uparrow$}
& LU     & 0.436 & 0.492 & 0.500 & 0.942 & 0.979 & 0.409 & 0.484 & 0.498 & 0.938 & 0.968 \\
& NNLU  & 0.479 & \secondval{0.497} & \secondval{0.942} & \secondval{0.956} & 0.978 & \secondval{0.900} & \secondval{0.916} & 0.498 & 0.938 & 0.968 \\
& HyU    & 0.445 & 0.005 & 0.933 & 0.945 & 0.981 & 0.446 & 0.493 & \secondval{0.932} & \secondval{0.945} & \secondval{0.969} \\
& RLU    & \secondval{0.902} & 0.430 & 0.930 & 0.947 & \secondval{0.985} & 0.899 & 0.492 & 0.927 & 0.936 & 0.968 \\
& NMF-RI & 0.459 & 0.494 & 0.929 & 0.442 & 0.956 & 0.006 & 0.485 & 0.498 & 0.938 & 0.966 \\
& $\lambda$Split (ours)
         & \bestval{0.919} & \bestval{0.931} & \bestval{0.956} & \bestval{0.983} & \bestval{0.999}
         & \bestval{0.917} & \bestval{0.932} & \bestval{0.973} & \bestval{0.996} & \bestval{0.999} \\
\midrule

\multirow{6}{*}{LPIPS $\downarrow$}
& LU     & 0.497 & 0.244 & 0.107 & 0.056 & 0.019 & 0.526 & 0.284 & 0.129 & 0.066 & 0.028 \\
& NNLU  & 0.289 & \secondval{0.156} & \secondval{0.079} & 0.047 & \secondval{0.016} & \bestval{0.201} & 0.206 & 0.120 & 0.062 & 0.028 \\
& HyU    & 0.436 & 0.205 & 0.088 & 0.052 & 0.019 & 0.445 & 0.215 & \secondval{0.089} & \secondval{0.053} & 0.029 \\
& RLU    & \bestval{0.189} & 0.161 & 0.080 & \secondval{0.042} & \bestval{0.013} & 0.239 & \secondval{0.191} & \secondval{0.109} & 0.058 & \secondval{0.026} \\
& NMF-RI & 0.333 & 0.178 & 0.098 & 0.054 & 0.024 & 0.404 & 0.235 & 0.122 & 0.063 & 0.028 \\
& $\lambda$Split (ours)
         & \secondval{0.205} & \bestval{0.120} & \bestval{0.064} & \bestval{0.030} & \secondval{0.016}
         & \secondval{0.213} & \bestval{0.117} & \bestval{0.047} & \bestval{0.026} & \bestval{0.022} \\
\midrule

\multirow{6}{*}{Pearson $\uparrow$}
& LU     & 0.812 & 0.950 & 0.983 & 0.993 & \secondval{0.998} & 0.786 & 0.943 & 0.981 & 0.993 & \secondval{0.998} \\
& NNLU  & 0.798 & \secondval{0.956} & \secondval{0.986} & \secondval{0.994} & \secondval{0.998} & 0.708 & 0.863 & 0.978 & 0.993 & \secondval{0.998} \\
& HyU    & 0.800 & 0.949 & 0.983 & 0.993 & \secondval{0.998} & \secondval{0.812} & \secondval{0.957} & \secondval{0.986} & \secondval{0.994} & \secondval{0.998} \\
& RLU    & 0.716 & 0.883 & 0.977 & 0.993 & \bestval{0.999} & 0.734 & 0.901 & 0.974 & 0.991 & \secondval{0.998} \\
& NMF-RI & \secondval{0.834} & 0.934 & 0.974 & 0.991 & 0.995 & 0.801 & 0.941 & 0.975 & 0.992 & 0.997 \\
& $\lambda$Split (ours)
         & \bestval{0.942} & \bestval{0.981} & \bestval{0.993} & \bestval{0.997} & \bestval{0.999}
         & \bestval{0.927} & \bestval{0.982} & \bestval{0.995} & \bestval{0.998} & \bestval{0.999} \\
\midrule

\multirow{6}{*}{SNR$_u$ $\uparrow$}
& LU     & 25.40 & 58.26 & 105.80 & 158.97 & 269.47 & 12.87 & 28.39 & 51.33 & 84.45 & 128.62 \\
& NNLU  & 71.14 & 92.65 & \secondval{136.55} & 162.99 & 269.43 & \secondval{72.72} & \secondval{43.96} & 53.23 & 84.46 & 128.63 \\
& HyU    & 30.24 & 72.08 & 129.74 & 171.86 & 269.86 & 19.00 & \secondval{41.86} & \secondval{72.44} & \secondval{99.56} & \secondval{129.57} \\
& RLU    & \secondval{153.65} & \secondval{101.24} & \secondval{130.79} & \secondval{205.80} & \secondval{282.33} & 42.54 & 39.83 & 57.90 & 92.04 & 127.37 \\
& NMF-RI & 40.83 & 72.62 & 117.14 & 161.55 & 271.72 & 16.00 & 29.41 & 53.38 & 85.61 & 128.76 \\
& $\lambda$Split (ours)
         & \bestval{680.76} & \bestval{1071.93} & \bestval{1533.56} & \bestval{1446.52} & \bestval{1436.26}
         & \bestval{800.66} & \bestval{1190.44} & \bestval{1109.60} & \bestval{1204.03} & \bestval{1253.93} \\
\bottomrule
\end{tabular}%
}
\end{minipage}%
}
\end{table*}

%% file: tables/table_5_low_num_bands_extra_metrics.tex
\begin{table*}[t]
\centering
\caption{
{\bf Quantitative results on \textit{CellAtlas} for varying spectral dimensionality.}
We evaluate spectral unmixing performance when reducing the number of acquisition bands.
Two experimental settings are considered:
(\textit{left}) the number of bands is reduced while keeping the per-band SNR constant, isolating the effect of spectral dimensionality; 
(\textit{right}) the total photon budget is fixed, such that fewer bands lead to higher per-band SNR.
Find more details in \cref{sec:results_fewer_bands}.
Best results are shown in \textbf{bold}, second best \uline{underlined}.
Note that SNR$_u$ is not available for LUMoS, as the algorithm performs hard clustering and thus produces a flat, uniform background with a standard deviation of $0$, rendering our SNR computation ill-posed (see also \cref{eq:SNR_formula}).
}
\label{tab:results_low_num_bands_extra}
\vspace{2mm}
\renewcommand{\arraystretch}{1.0}
\setlength{\tabcolsep}{2.0pt}
\footnotesize

\begin{tabular*}{\linewidth}{@{\extracolsep{\fill}}llcccccccc}
\toprule
& & \multicolumn{4}{c}{\shortstack{\textbf{\# bands} \\ {Same per-band SNR}}} & \multicolumn{4}{c}{\shortstack{\textbf{\# bands} \\ {Same total
 photon budget}}} \\
\cmidrule(lr){3-6} \cmidrule(lr){7-10}
\textbf{Metric} & \textbf{Method} & 3 & 4 & 5 & 32 & 3 & 4 & 5 & 32 \\
\midrule

\multirow{7}{*}{$\mu$MS3IM $\uparrow$}
& LU~\cite{Zimmermann2014-nl}     & 0.516 & 0.548 & 0.597 & 0.696 & 0.564 & 0.726 & 0.597 & \secondval{0.770} \\
& NNLU~\cite{Slawski2011-nt}  & 0.728 & 0.691 & 0.620 & 0.696 & \bestval{0.773} & 0.750 & 0.620 & \secondval{0.770} \\
& HyU~\cite{Chiang2023-vm}    & 0.516 & 0.460 & 0.781 & 0.873 & 0.564 & \bestval{0.751} & 0.781 & 0.660 \\
& RLU~\cite{Kumar2025-iy}    & \secondval{0.733} & 0.693 & \secondval{0.813} & \secondval{0.907} & 0.631 & \secondval{0.752} & \secondval{0.813} & 0.713 \\
& NMF-RI~\cite{Jimenez-Sanchez2020-mh} & 0.724 & \secondval{0.700} & 0.544 & 0.703 & \secondval{0.757} & 0.747 & 0.544 & 0.596 \\
& LUMoS~\cite{McRae2019-cn}  & 0.688 & 0.617 & 0.647 & 0.642 & 0.721 & 0.644 & 0.647 & 0.682 \\
& $\lambda$Split (ours)
         & \bestval{0.752} & \bestval{0.873} & \bestval{0.910} & \bestval{0.988} & 0.737 & \bestval{0.893} & \bestval{0.910} & \bestval{0.930} \\
\midrule

\multirow{7}{*}{Pearson $\uparrow$}
& LU     & 0.733 & 0.879 & 0.912 & 0.987 & 0.761 & 0.898 & 0.912 & 0.877 \\
& NNLU  & \secondval{0.837} & 0.897 & 0.933 & 0.988 & 0.866 & 0.916 & 0.933 & 0.877 \\
& HyU    & 0.732 & 0.883 & 0.898 & 0.953 & 0.761 & 0.893 & 0.898 & 0.873 \\
& RLU    & \bestval{0.843} & 0.905 & \secondval{0.939} & \bestval{0.990} & \bestval{0.879} & \secondval{0.922} & \secondval{0.939} & \secondval{0.888} \\
& NMF-RI & 0.832 & \secondval{0.910} & 0.928 & \secondval{0.989} & 0.854 & 0.923 & 0.928 & 0.886 \\
& LUMoS  & 0.522 & 0.603 & 0.581 & 0.625 & 0.540 & 0.593 & 0.581 & 0.547 \\
& $\lambda$Split (ours)
         & \bestval{0.843} & \bestval{0.939} & \bestval{0.959} & \secondval{0.988} & \secondval{0.859} & \bestval{0.955} & \bestval{0.959} & \bestval{0.964} \\
\midrule

\multirow{7}{*}{SNR $\uparrow$}
& LU     & 25.64 & 21.65 & 30.34 & 78.53 & 41.43 & 28.59 & 30.34 & 13.76 \\
& NNLU  & 27.12 & 22.25 & 31.31 & 78.39 & 46.82 & 29.43 & 31.31 & 13.77 \\
& HyU    & 25.68 & 33.49 & 47.55 & 115.94 & 41.48 & 42.34 & 47.55 & 19.16 \\
& RLU    & 29.36 & 23.12 & 33.30 & 78.75 & 48.20 & 30.42 & 33.30 & 13.86 \\
& NMF-RI & 31.27 & 26.09 & 33.51 & 81.31 & 52.78 & 32.36 & 33.51 & 14.29 \\
& LUMoS  & -- & -- & -- & -- & -- & -- & -- & -- \\
& $\lambda$Split (ours)
         & \bestval{525.91} & \bestval{922.37} & \bestval{2508.35} & \bestval{1681.98} & \bestval{759.06} & \bestval{1640.65} & \bestval{2508.35} & \bestval{318.82} \\
\bottomrule
\end{tabular*}
\end{table*}

%% file: tables/table_6_autounmix_bands.tex
\begin{table*}[t]
\centering
\caption{
\textbf{Quantitative comparison of $\lambda$Split \vs supervised spectral unmixing baselines for low spectral dimensionality}.
The first four columns, from left to right, report: $(i)$~the number of bands in the input spectral data; $(ii)$~the considered baselines; $(iii)$~whether the method is self-supervised; $(iv)$~the number of parameters in the architecture used by each baseline.
The following columns report a subset of metrics for experiments in which: (\textit{left}) the number of bands is reduced while keeping the per-band SNR constant, isolating the effect of spectral dimensionality; (\textit{right}) the total photon budget is fixed, such that fewer bands lead to
higher per-band SNR.
}
\label{tab:autounmix_bands}
\renewcommand{\arraystretch}{0.95}
\setlength{\tabcolsep}{2.0pt}
\footnotesize
\vspace{2mm}

\begin{tabular*}{\linewidth}{@{\extracolsep{\fill}}c l c c ccc ccc}
\toprule
\multicolumn{4}{c}{} & \multicolumn{3}{c}{\textbf{Same per-band SNR}} & \multicolumn{3}{c}{\textbf{Same tot. photon budget}} \\
\cmidrule(lr){5-7} \cmidrule(lr){8-10}
\textbf{Bands} & \textbf{Method} & \textbf{SSL} & \textbf{Params} & PSNR$\uparrow$ & MS3IM$\uparrow$ & LPIPS$\downarrow$ & PSNR$\uparrow$ & MS3IM$\uparrow$ & LPIPS$\downarrow$ \\
\midrule

\multirow{3}{*}{3}
& AutoUnmix~\cite{Jiang2023-fm} & \textcolor{red}{\xmark} & 84M & {32.24} & {0.961} & {0.106} & {33.05} & {0.973} & {0.086} \\
& UNet~\cite{Ronneberger2015-ei} & \textcolor{red}{\xmark} & 31M & 32.54 & 0.965 & 0.128 & 32.72 & 0.968 & 0.139 \\
& $\lambda$Split (ours) & \textcolor{green}{\vmark} & 3M & 27.25 & 0.878 & 0.283 & 27.45 & 0.868 & 0.273 \\
\midrule

\multirow{3}{*}{4}
& AutoUnmix & \textcolor{red}{\xmark} & 84M & {34.01} & {0.980} & {0.088} & {34.50} & {0.982} & {0.081} \\
& UNet & \textcolor{red}{\xmark} & 31M & 33.76 & 0.977 & 0.139 & 33.98 & 0.977 & 0.146 \\
& $\lambda$Split (ours) & \textcolor{green}{\vmark} & 3M & 29.95 & 0.960 & 0.267 & 31.54 & 0.967 & 0.238 \\
\midrule

\multirow{3}{*}{5}
& AutoUnmix & \textcolor{red}{\xmark} & 84M & {34.16} & {0.983} & {0.076} & {34.16} & {0.983} & {0.076} \\
& UNet & \textcolor{red}{\xmark} & 31M & 33.81 & 0.980 & 0.135 & 33.81 & 0.980 & 0.135 \\
& $\lambda$Split (ours) & \textcolor{green}{\vmark} & 3M & 31.12 & 0.971 & 0.313 & 31.12 & 0.971 & 0.313 \\
\bottomrule
\end{tabular*}
\end{table*}

%% file: tables/table_7a_autounmix_overlap_EGFP-EC.tex
\begin{table*}[t]
\centering
\caption{
\textbf{Quantitative comparison of $\lambda$Split \vs supervised spectral unmixing baselines for different spectral overlaps.}
In these experiments, we consider the ER and CCPs structures from the \textit{BioSR} dataset and associate them with the EGFP fluorophore.
The datasets used here comprise 5-band spectral images. 
}
\label{tab:autounmix_overlap_EGFP-EC}
\renewcommand{\arraystretch}{0.95}
\setlength{\tabcolsep}{3.5pt}
\footnotesize
\vspace{2mm}

\begin{tabular*}{\linewidth}{@{\extracolsep{\fill}}l l c c ccccc}
\toprule
\multicolumn{9}{c}{\textbf{EGFP -- ER/CCPs -- 5 bands}} \\
\midrule
\textbf{Metric} & \textbf{Method} & \textbf{SSL} & \textbf{\#params}
& $\Delta\lambda=2$ & $\Delta\lambda=5$ & $\Delta\lambda=10$ & $\Delta\lambda=20$ & $\Delta\lambda=50$ \\
\midrule

\multirow{3}{*}{PSNR $\uparrow$}
& AutoUnmix~\cite{Jiang2023-fm}      & \textcolor{red}{\xmark} & 84M & {46.82} & {50.14} & {51.10} & 50.50 & 52.62 \\
& UNet~\cite{Ronneberger2015-ei} & \textcolor{red}{\xmark} & 31M & 51.77 & 49.91 & 51.88 & 50.14 & 53.38 \\
& $\lambda$Split (ours) & \textcolor{green}{\vmark} & 3M & 37.61 & 42.91 & 47.55 & {51.40} & {54.29} \\
\midrule

\multirow{3}{*}{MS3IM $\uparrow$}
& AutoUnmix      & \textcolor{red}{\xmark} & 84M & {0.998} & {0.999} & {0.999} & {0.999} & {0.999} \\
& UNet & \textcolor{red}{\xmark} & 31M & 0.999 & 0.998 & 0.999 & 0.998 & 0.999 \\
& $\lambda$Split (ours) & \textcolor{green}{\vmark} & 3M & 0.949 & 0.981 & 0.994 & 0.998 & {0.999} \\
\midrule

\multirow{3}{*}{LPIPS $\downarrow$}
& AutoUnmix      & \textcolor{red}{\xmark} & 84M & {0.031} & {0.016} & {0.012} & {0.011} & {0.008} \\
& UNet & \textcolor{red}{\xmark} & 31M & 0.013 & 0.020 & 0.018 & 0.024 & 0.010 \\
& $\lambda$Split (ours) & \textcolor{green}{\vmark} & 3M & 0.196 & 0.094 & 0.041 & 0.022 & 0.027 \\
\midrule

\multirow{3}{*}{$\mu$MS3IM $\uparrow$}
& AutoUnmix      & \textcolor{red}{\xmark} & 84M & {0.997} & {0.999} & {0.999} & {0.999} & {0.999} \\
& UNet & \textcolor{red}{\xmark} & 31M & 0.996 & 0.982 & 0.991 & 0.988 & 0.990 \\
& $\lambda$Split (ours) & \textcolor{green}{\vmark} & 3M & 0.914 & 0.926 & 0.969 & 0.993 & {0.999} \\
\midrule

\multirow{3}{*}{Pearson $\uparrow$}
& AutoUnmix      & \textcolor{red}{\xmark} & 84M & {0.995} & {0.998} & {0.998} & {0.998} & {0.999} \\
& UNet & \textcolor{red}{\xmark} & 31M & 0.999 & 0.998 & 0.999 & 0.998 & 0.999 \\
& $\lambda$Split (ours) & \textcolor{green}{\vmark} & 3M & 0.933 & 0.982 & 0.995 & {0.998} & {0.999} \\
\midrule

\multirow{3}{*}{SNR$_u$ $\uparrow$}
& AutoUnmix      & \textcolor{red}{\xmark} & 84M & {1573.79} & {inf} & {1951.91} & {inf} & {inf} \\
& UNet & \textcolor{red}{\xmark} & 31M & 317.94 & 182.18 & 203.52 & 206.69 & 309.55 \\
& $\lambda$Split (ours) & \textcolor{green}{\vmark} & 3M & 727.70 & 1048.60 & 1287.36 & 1356.12 & 1711.83 \\
\bottomrule
\end{tabular*}
\end{table*}

%% file: tables/table_7b_autounmix_overlap_EGFP-MC.tex
\begin{table*}[t]
\centering
\caption{
\textbf{Quantitative comparison of $\lambda$Split \vs supervised spectral unmixing baselines for different spectral overlaps.}
In these experiments, we consider the Microtubules and CCPs structures from the \textit{BioSR} dataset and associate them with the EGFP fluorophore.
The datasets used here comprise 5-band spectral images. 
}
\label{tab:autounmix_overlap_EGFP-MC}
\renewcommand{\arraystretch}{0.95}
\setlength{\tabcolsep}{3.5pt}
\footnotesize
\vspace{2mm}

\begin{tabular*}{\linewidth}{@{\extracolsep{\fill}}l l c c ccccc}
\toprule
\multicolumn{9}{c}{\textbf{EGFP -- Microtubules/CCPs -- 5 bands}} \\
\midrule
\textbf{Metric} & \textbf{Method} & \textbf{SSL} & \textbf{\#params}
& $\Delta\lambda=2$ & $\Delta\lambda=5$ & $\Delta\lambda=10$ & $\Delta\lambda=20$ & $\Delta\lambda=50$ \\
\midrule

\multirow{3}{*}{PSNR $\uparrow$}
& AutoUnmix~\cite{Jiang2023-fm}      & \textcolor{red}{\xmark} & 84M & {48.08} & {49.70} & {51.30} & {52.88} & {53.62} \\
& UNet~\cite{Ronneberger2015-ei} & \textcolor{red}{\xmark} & 31M & 49.49 & 52.56 & 51.34 & 50.62 & 52.53 \\
& $\lambda$Split (ours) & \textcolor{green}{\vmark} & 3M & 37.91 & 42.81 & 47.31 & 51.27 & 52.65 \\
\midrule

\multirow{3}{*}{MS3IM $\uparrow$}
& AutoUnmix      & \textcolor{red}{\xmark} & 84M & {0.998} & {0.999} & {0.999} & {1.000} & {1.000} \\
& UNet & \textcolor{red}{\xmark} & 31M & 0.998 & 0.999 & 0.999 & 0.999 & 0.999 \\
& $\lambda$Split (ours) & \textcolor{green}{\vmark} & 3M & 0.954 & 0.983 & 0.994 & 0.998 & 0.999 \\
\midrule

\multirow{3}{*}{LPIPS $\downarrow$}
& AutoUnmix      & \textcolor{red}{\xmark} & 84M & {0.018} & {0.012} & {0.009} & {0.005} & {0.004} \\
& UNet & \textcolor{red}{\xmark} & 31M & 0.033 & 0.018 & 0.020 & 0.026 & 0.021 \\
& $\lambda$Split (ours) & \textcolor{green}{\vmark} & 3M & 0.209 & 0.109 & 0.049 & 0.024 & 0.030 \\
\midrule

\multirow{3}{*}{$\mu$MS3IM $\uparrow$}
& AutoUnmix      & \textcolor{red}{\xmark} & 84M & {0.998} & {0.999} & {0.999} & {1.000} & {1.000} \\
& UNet & \textcolor{red}{\xmark} & 31M & 0.985 & 0.996 & 0.986 & 0.986 & 0.993 \\
& $\lambda$Split (ours) & \textcolor{green}{\vmark} & 3M & 0.920 & 0.932 & 0.971 & 0.996 & 0.998 \\
\midrule

\multirow{3}{*}{Pearson $\uparrow$}
& AutoUnmix      & \textcolor{red}{\xmark} & 84M & {0.996} & {0.998} & {0.998} & {0.999} & {0.999} \\
& UNet & \textcolor{red}{\xmark} & 31M & 0.997 & 0.999 & 0.998 & 0.998 & 0.999 \\
& $\lambda$Split (ours) & \textcolor{green}{\vmark} & 3M & 0.935 & 0.982 & 0.995 & 0.998 & {0.999} \\
\midrule

\multirow{3}{*}{SNR$_u$ $\uparrow$}
& AutoUnmix      & \textcolor{red}{\xmark} & 84M & {3202.17} & {inf} & {inf} & {inf} & {inf} \\
& UNet & \textcolor{red}{\xmark} & 31M & 198.68 & 431.39 & 169.56 & 240.46 & 419.66 \\
& $\lambda$Split (ours) & \textcolor{green}{\vmark} & 3M & 806.40 & 1017.92 & 1527.50 & 1170.58 & 992.97 \\
\bottomrule
\end{tabular*}
\end{table*}

%% file: tables/table_7c_autounmix_overlap_mTurquoise-EC.tex
\begin{table*}[t]
\centering
\caption{
\textbf{Quantitative comparison of $\lambda$Split \vs supervised spectral unmixing baselines for different spectral overlaps.}
In these experiments, we consider the ER and CCPs structures from the \textit{BioSR} dataset and associate them with the mTurquoise fluorophore.
The datasets used here comprise 5-band spectral images. 
}
\label{tab:autounmix_overlap_mTurquoise-EC}
\renewcommand{\arraystretch}{0.95}
\setlength{\tabcolsep}{3.5pt}
\footnotesize
\vspace{2mm}

\begin{tabular*}{\linewidth}{@{\extracolsep{\fill}}l l c c ccccc}
\toprule
\multicolumn{9}{c}{\textbf{mTurquoise -- ER/CCPs -- 5 bands}} \\
\midrule
\textbf{Metric} & \textbf{Method} & \textbf{SSL} & \textbf{\#params}
& $\Delta\lambda=2$ & $\Delta\lambda=5$ & $\Delta\lambda=10$ & $\Delta\lambda=20$ & $\Delta\lambda=50$ \\
\midrule

\multirow{3}{*}{PSNR $\uparrow$}
& AutoUnmix~\cite{Jiang2023-fm}      & \textcolor{red}{\xmark} & 84M & {43.18} & {48.58} & {49.39} & 49.67 & 52.27 \\
& UNet~\cite{Ronneberger2015-ei} & \textcolor{red}{\xmark} & 31M & 51.05 & 49.77 & 50.52 & 50.80 & 52.40 \\
& $\lambda$Split (ours) & \textcolor{green}{\vmark} & 3M & 36.88 & 42.24 & 47.32 & {49.75} & {52.60} \\
\midrule

\multirow{3}{*}{MS3IM $\uparrow$}
& AutoUnmix      & \textcolor{red}{\xmark} & 84M & {0.993} & {0.998} & {0.999} & {0.999} & {0.999} \\
& UNet & \textcolor{red}{\xmark} & 31M & 0.999 & 0.998 & 0.998 & 0.999 & 0.999 \\
& $\lambda$Split (ours) & \textcolor{green}{\vmark} & 3M & 0.950 & 0.980 & 0.992 & 0.996 & {0.999} \\
\midrule

\multirow{3}{*}{LPIPS $\downarrow$}
& AutoUnmix      & \textcolor{red}{\xmark} & 84M & {0.057} & {0.019} & {0.015} & {0.014} & {0.010} \\
& UNet & \textcolor{red}{\xmark} & 31M & 0.013 & 0.021 & 0.022 & 0.023 & 0.019 \\
& $\lambda$Split (ours) & \textcolor{green}{\vmark} & 3M & 0.173 & 0.122 & 0.051 & 0.030 & 0.026 \\
\midrule

\multirow{3}{*}{$\mu$MS3IM $\uparrow$}
& AutoUnmix      & \textcolor{red}{\xmark} & 84M & {0.993} & {0.998} & {0.999} & {0.999} & {0.999} \\
& UNet & \textcolor{red}{\xmark} & 31M & 0.994 & 0.985 & 0.982 & 0.990 & 0.989 \\
& $\lambda$Split (ours) & \textcolor{green}{\vmark} & 3M & 0.911 & 0.925 & 0.953 & 0.983 & 0.997 \\
\midrule

\multirow{3}{*}{Pearson $\uparrow$}
& AutoUnmix      & \textcolor{red}{\xmark} & 84M & {0.991} & {0.997} & {0.998} & {0.998} & {0.999} \\
& UNet & \textcolor{red}{\xmark} & 31M & 0.998 & 0.998 & 0.998 & 0.998 & 0.999 \\
& $\lambda$Split (ours) & \textcolor{green}{\vmark} & 3M & 0.935 & 0.979 & 0.993 & 0.997 & {0.999} \\
\midrule

\multirow{3}{*}{SNR$_u$ $\uparrow$}
& AutoUnmix      & \textcolor{red}{\xmark} & 84M & {inf} & {774.84} & {5229.45} & 1439.43 & {inf} \\
& UNet & \textcolor{red}{\xmark} & 31M & 338.27 & 208.80 & 193.97 & 276.15 & 240.87 \\
& $\lambda$Split (ours) & \textcolor{green}{\vmark} & 3M & 428.91 & 735.90 & 1331.06 & {1641.54} & 1028.25 \\
\bottomrule
\end{tabular*}
\end{table*}

%% file: tables/table_7d_autounmix_overlap_mTurquoise-MC.tex
\begin{table*}[t]
\centering
\caption{
\textbf{Quantitative comparison of $\lambda$Split \vs supervised spectral unmixing baselines for different spectral overlaps.}
In these experiments, we consider the Microtubules and CCPs structures from the \textit{BioSR} dataset and associate them with the mTurquoise fluorophore.
The simulated datasets used here comprise 5-band spectral images. 
}
\label{tab:autounmix_overlap_mTurquoise-MC}
\renewcommand{\arraystretch}{0.95}
\setlength{\tabcolsep}{3.5pt}
\footnotesize
\vspace{2mm}

\begin{tabular*}{\linewidth}{@{\extracolsep{\fill}}l l c c ccccc}
\toprule
\multicolumn{9}{c}{\textbf{mTurquoise -- Microtubules/CCPs -- 5 bands}} \\
\midrule
\textbf{Metric} & \textbf{Method} & \textbf{SSL} & \textbf{\#params}
& $\Delta\lambda=2$ & $\Delta\lambda=5$ & $\Delta\lambda=10$ & $\Delta\lambda=20$ & $\Delta\lambda=50$ \\
\midrule

\multirow{3}{*}{PSNR $\uparrow$}
& AutoUnmix~\cite{Jiang2023-fm}      & \textcolor{red}{\xmark} & 84M & {46.09} & {48.47} & {49.85} & {51.69} & 52.52 \\
& UNet~\cite{Ronneberger2015-ei} & \textcolor{red}{\xmark} & 31M & 48.55 & 51.08 & 51.73 & 55.76 & 51.89 \\
& $\lambda$Split (ours) & \textcolor{green}{\vmark} & 3M & 37.64 & 43.17 & 46.52 & 49.26 & {53.27} \\
\midrule

\multirow{3}{*}{MS3IM $\uparrow$}
& AutoUnmix      & \textcolor{red}{\xmark} & 84M & {0.997} & {0.998} & {0.999} & {0.999} & {1.000} \\
& UNet & \textcolor{red}{\xmark} & 31M & 0.998 & 0.999 & 0.999 & 1.000 & 0.999 \\
& $\lambda$Split (ours) & \textcolor{green}{\vmark} & 3M & 0.956 & 0.982 & 0.992 & 0.996 & 0.999 \\
\midrule

\multirow{3}{*}{LPIPS $\downarrow$}
& AutoUnmix      & \textcolor{red}{\xmark} & 84M & {0.026} & {0.021} & {0.011} & {0.009} & {0.005} \\
& UNet & \textcolor{red}{\xmark} & 31M & 0.027 & 0.032 & 0.021 & 0.010 & 0.024 \\
& $\lambda$Split (ours) & \textcolor{green}{\vmark} & 3M & 0.205 & 0.120 & 0.064 & 0.030 & 0.016 \\
\midrule

\multirow{3}{*}{$\mu$MS3IM $\uparrow$}
& AutoUnmix      & \textcolor{red}{\xmark} & 84M & {0.997} & {0.998} & {0.999} & {0.999} & {0.999} \\
& UNet & \textcolor{red}{\xmark} & 31M & 0.991 & 0.993 & 0.991 & 0.999 & 0.992 \\
& $\lambda$Split (ours) & \textcolor{green}{\vmark} & 3M & 0.919 & 0.931 & 0.956 & 0.983 & {0.999} \\
\midrule

\multirow{3}{*}{Pearson $\uparrow$}
& AutoUnmix      & \textcolor{red}{\xmark} & 84M & {0.994} & {0.997} & {0.998} & {0.999} & {0.999} \\
& UNet & \textcolor{red}{\xmark} & 31M & 0.997 & 0.998 & 0.998 & 0.999 & 0.999 \\
& $\lambda$Split (ours) & \textcolor{green}{\vmark} & 3M & 0.942 & 0.981 & 0.993 & 0.997 & {0.999} \\
\midrule

\multirow{3}{*}{SNR$_u$ $\uparrow$}
& AutoUnmix      & \textcolor{red}{\xmark} & 84M & {inf} & {3665.90} & 1398.69 & {1567.03} & {7436.73} \\
& UNet & \textcolor{red}{\xmark} & 31M & 366.82 & 319.27 & 234.11 & 537.91 & 267.64 \\
& $\lambda$Split (ours) & \textcolor{green}{\vmark} & 3M & 680.76 & 1071.93 & {1533.56} & 1446.52 & 1436.26 \\
\bottomrule
\end{tabular*}
\end{table*}